\documentclass[10pt,journal,compsoc]{IEEEtran}

\usepackage{graphicx}
\usepackage{amsmath}
\usepackage{amsfonts}
\usepackage{subcaption}
\usepackage{xcolor}

\def \figurepath {plots/}
\newcommand{\norm}[1]{\Vert #1 \Vert}
\newcommand{\fnorm}[1]{\norm{#1}_{\mathrm{F}}}
\newcommand{\myparagraph}[1]{\noindent\textbf{#1.}\,}

\usepackage[mathscr]{euscript}
\usepackage{enumitem}

\usepackage[noend]{algpseudocode}
\usepackage{algorithm,algorithmicx}
\algrenewcommand\alglinenumber[1]{\sf\tiny\color{medblue}{#1}\quad}
\algrenewcommand\algorithmicrequire{\textbf{Input:}}
\algrenewcommand\algorithmicensure{\textbf{Output:}}
\algdef{SE}[DOWHILE]{Do}{doWhile}{\algorithmicdo}[1]{\algorithmicwhile\ #1}%

\newcommand{\T}[2][]{#1\mathscr{\MakeUppercase{#2}}}

\definecolor{purple}{rgb}{0.3,0.0,.4}
\definecolor{medblue}{rgb}{0,0,.75}

\DeclareMathOperator*{\argmax}{arg\,max}
\DeclareMathOperator*{\argmin}{arg\,min}

\usepackage[english]{babel}
\newtheorem{theorem}{Theorem}[section]
\newtheorem{lemma}[theorem]{Lemma}
\usepackage{multirow}
\usepackage{url}

\ifCLASSOPTIONcompsoc
  \usepackage[nocompress]{cite}
\else
  \usepackage{cite}
\fi

\ifCLASSINFOpdf
\else
\fi

\begin{document}
\title{Efficient AutoML Pipeline Search \\ with Matrix and Tensor Factorization}

\author{Chengrun~Yang,
        Jicong~Fan,
        Ziyang~Wu,
        and~Madeleine~Udell
\IEEEcompsocitemizethanks{\IEEEcompsocthanksitem C. Yang, J. Fan, Z.Wu and M. Udell are with Cornell University, Ithaca,
NY, 14850.\protect\\
E-mail: $\lbrace$cy438, jf577, zw287, udell$\rbrace$@cornell.edu
}
\thanks{}}

\IEEEtitleabstractindextext{%
\textcolor{blue}{This is an extended version of \textit{AutoML Pipeline Selection: Efficiently Navigating the Combinatorial Space} (DOI: 10.1145/3394486.3403197) at the 26th ACM SIGKDD International Conference on Knowledge Discovery and Data Mining, 2020. }
\linebreak

\begin{abstract}
Data scientists seeking a good supervised learning model on a new dataset
have many choices to make: they must preprocess the data, select features,
possibly reduce the dimension, select an estimation algorithm,
and choose hyperparameters for each of these pipeline components.
With new pipeline components comes a combinatorial explosion in the number of choices!
In this work, we design a new AutoML system to address this challenge:
an automated system to design a supervised learning pipeline.
Our system uses matrix and tensor factorization as surrogate models to model the combinatorial pipeline search space.
Under these models, we develop greedy experiment design protocols to efficiently gather information about a new dataset.
Experiments on large corpora of real-world classification problems demonstrate the effectiveness of our approach.
\end{abstract}

}

\maketitle

\IEEEdisplaynontitleabstractindextext

\IEEEpeerreviewmaketitle

\ifCLASSOPTIONcompsoc
\IEEEraisesectionheading{\section{Introduction}}
\else
\section{Introduction}
\label{sec:introduction}
\fi

\IEEEPARstart{A}{machine} learning pipeline is a directed graph of learning components
including imputation, encoding, standardization, dimensionality reduction, and estimation,
that together define a function mapping input data to output predictions.
Each components may also include hyperparameters, such as the output dimension of PCA, or
the number of trees in a random forest.
Simple pipelines may consist of sequences of these components; more complex pipelines
may combine inputs to form pipelines with more complex topologies.
\begin{figure}[h]
	\centering
	\includegraphics[width=\linewidth]{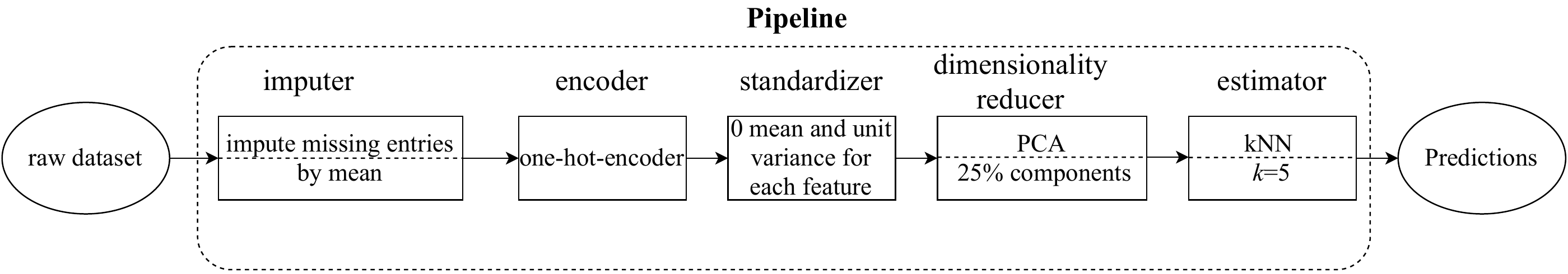}
	\caption{An example pipeline.}
	\label{fig:pipeline_example}
\end{figure}
An example pipeline is shown as Figure~\ref{fig:pipeline_example}.

The job of a data scientist facing a new supervised learning problem is to
choose the pipeline that yields a low out of sample error from among all possible pipelines.
This task is challenging.
First, no component dominates all others:
there is ``no free lunch'' \cite{wolpert1997no}.
Rather, each performs well on certain data distributions.
For example, the PCA dimensionality reducer works well on data points in $\mathbb{R}^d$ that roughly lie in a low rank subspace $\mathbb{R}^k$ with $k < d$;
the feature selector that keeps features with large variances works well on datasets if such features are more informative;
the Gaussian naive Bayes classifier works well on features with normally distributed values in each class.
However, it is difficult to check these distributional assumptions without running the component on the data: an expensive proposition!
The second is the dependence of these choices: for example, standardizing the data may help some estimators, and harm others.
Moreover, as the number of possible machine learning components grows,
the number of possibilities grows exponentially, defying enumeration.
Automating the selection of a pipeline is thus an important problem,
which has received attention both from academia and industry \cite{olson2019tpot, feurer2015efficient, drori2018alphad3m, liu2019admm}.

Human experts tackle this difficulty by choosing the right combination according to their domain knowledge.
However, finding the right combination takes substantial expertise, and
still requires several model fits to find the right combination of components and hyperparameters.
An automated pipeline construction system, like a human expert, first forms a \emph{surrogate model}
to predict which pipelines are likely to work well. 
Surrogate models are meta-models that map dataset and machine learning model properties
to quantities that characterize performance or informativeness.

A good surrogate model enables efficient search through the space of pipelines.
``All models are wrong, but some are useful \cite{box1976science}'':
a good surrogate model makes predictions that guide the search for pipelines
without the need for many model fits,
since it is expensive to evaluate the performance of a pipeline on a large dataset.
Auto-sklearn \cite{feurer2015efficient} uses meta-learning \cite{thrun2012learning, andrychowicz2016learning, lake2017building, vanschoren2018meta} to choose promising pipelines from those that performed best on neighboring datasets,
and uses Bayesian optimization to fine-tune hyperparameters.
TPOT \cite{olson2019tpot} uses genetic programming to search over pipeline topologies.
Alpine Meadow \cite{shang2019democratizing} casts pipeline structure search
as a multi-armed bandit problem and tunes model hyperparameters by Bayesian optimization.
Our surrogate models in this paper are low rank matrices, tensors and kernelized matrices.
This model makes explicit use of the combinatorial structure of the problem:
as a result, the number of pipeline evaluations required to fit the surrogate model
on a new dataset is modest, and independent of the number of pipeline components.

Our system learns a model for a new dataset by fitting a few pipelines on the dataset.
The problem of which pipelines to evaluate first, in order to predict the effectiveness of others,
is called the \emph{cold-start problem} in the literature on recommender systems.
This problem is also of great interest to the AutoML community.
Proximity in meta-features, ``simple, statistical or landmarking metrics to characterize datasets \cite{yang2019oboe}'',
are used by many AutoML systems \cite{pfahringer2000meta, feurer2014using, feurer2015efficient, fusi2018probabilistic}
to select models that work well on neighboring datasets, with the belief that models perform similarly on datasets with similar characteristics.
Probabilistic matrix factorization has been used to extract dataset latent representations from pipeline performance \cite{fusi2018probabilistic}.
Other dataset and pipeline embeddings have also been proposed: use pipeline performance
or even textual dataset or algorithm descriptions to build surrogate models \cite{wistuba2015learning, yang2019oboe, drori2019automl}.

The active learning subproblem is to gain the most information
to guide further model selection.
Some approaches choose a function class to capture
the dependence of model performance on hyperparameters; examples are Gaussian processes \cite{williams2006gaussian,snoek2012practical, bergstra2011algorithms,fusi2018probabilistic, sebastiani2000maximum,herbrich2003fast, mackay1992information, srinivas2009gaussian}, sparse Boolean functions \cite{hazan2018hyperparameter} and decision trees \cite{bartz2004tuning,hutter2011sequential}.
In this paper,  our \textsc{Oboe} \footnote{The eponymous musical instrument plays the initial note to tune an orchestra, which echoes our method that learn the entire space by using the knowledge from some initial evaluations.} and its extensions choose the set of multilinear models as its function class:
predicted performance is linear in each of the model and dataset embeddings.

In this work, we build pipeline embeddings by fitting factorization models
to the (sometimes incompletely observed) matrix or tensor of pipeline performance on a set of training datasets.
These models are easy to extend to a single new dataset by fitting a constant number of pipelines on the new dataset.
We describe a simple rule to select which pipelines to observe by solving a constrained version of the classical experiment design \cite{wald1943efficient, john1975d, pukelsheim1993optimal, boyd2004convex} problem, using a greedy heuristic \cite{madan2019combinatorial}.

We consider the following concrete challenge:
select several pipelines that perform the best within a given time limit for a new dataset, in the case that we already know or have time to collect pipeline performance on some existing datasets.
What we contribute as new ideas include greedy experiment design for \textsc{Oboe}, a sampling model for preprocessing that takes less time for meta-training, a tensor approach as a new pipeline search mechanism, and a kernelized method that further increases accuracy.
We name the system we use to search across pipelines with the same preprocessors \textsc{Oboe}, the system that builds on the tensor surrogate model to search across different pipeline components \textsc{TensorOboe}, and the kernelized version we have to speed up meta-training \textsc{KernelOboe}.
Together, these ideas yield a state-of-the-art system for AutoML pipeline selection.

This paper is organized as follows.
Section~\ref{sec:notation} introduces notation and terminology.
Section~\ref{sec:oboe} describes the main ideas we use in \textsc{Oboe}.
Section~\ref{sec:tensoroboe} describes \textsc{TensorOboe}.
Section~\ref{sec:KernelOboe} describes \textsc{KernelOboe}.
Section~\ref{sec:experiments} shows experimental results.

\subsection{Notation and Terminology}
\label{sec:notation}
\begin{figure}
	\begin{subfigure}[t]{0.48\linewidth}
		\includegraphics[width=0.8\linewidth]{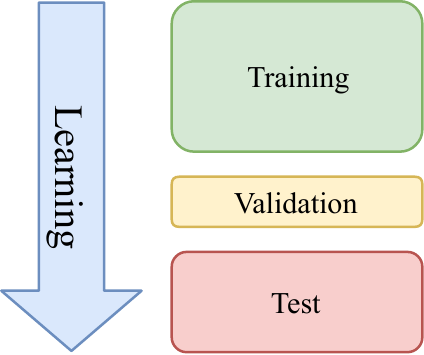}
		\caption{Learning}
		\label{fig:standardlearning}
	\end{subfigure}
	\begin{subfigure}[t]{0.48\linewidth}
		\includegraphics[width=0.8\linewidth]{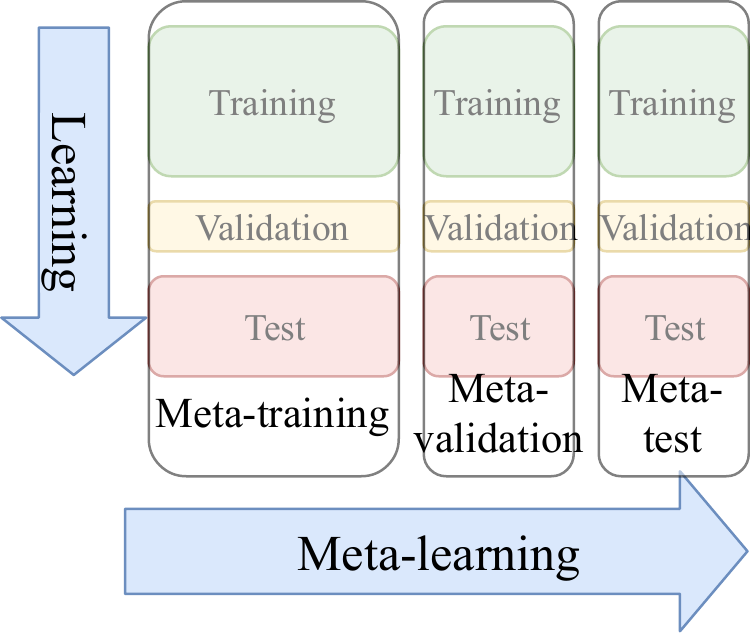}
		\caption{Meta-learning}
		\label{fig:metalearning}
	\end{subfigure}
	\caption{Standard learning vs meta-learning.}
	\label{fig:metalearningall}
\end{figure}

\myparagraph{Meta-learning}
Meta-learning, also called ``learning to learn'', uses results from past tasks to make predictions or decisions on a new task.
In our setting, we learn from a corpus of datasets called \emph{meta-training} datasets by fitting pipelines to these datasets in an offline stage; the new dataset, which requires a fast recommendation for a pipeline, is called the \emph{meta-test} dataset.
There can also exist \emph{meta-validation} datasets, which are the ones we use to evaluate our pipeline recommendation method. 
Each of the three phases in meta-learning --- meta-training, meta-validation and meta-test ---
is a standard learning process that includes training, validation and test, as shown in Figure~\ref{fig:metalearningall}.

\myparagraph{Model}
A \emph{model} $ \mathcal{A} $ is a specific combination of algorithm and hyperparameter settings,
e.g. $ k $-nearest neighbors with $ k=3 $.

\myparagraph{Pipeline component}
A pipeline component is a model or model type.
Examples include missing entry imputers, dimensionality reducers, supervised learners, and data visualizers.
We consider the following components in this paper:
\begin{itemize}
	\item \emph{Data imputer}: A preprocessor that fills in missing entries.
	\item \emph{Encoder}: A transformer that converts categorical features to numerical codes.
	Here, we consider encoding categoricals as integers or with a one-hot encoder.
	\item \emph{Standardizer}: A standardizer centers and rescales data.
	\item \emph{Dimensionality reducer}: A transformer that reduces the dimensionality of the dataset by either creating new features (like PCA) or subsampling features.
	\item \emph{Estimator}: The supervised learner. For the classification tasks in this paper, estimators are classifiers.
\end{itemize}

\myparagraph{Linear algebra} 
Our paper follows the notation of \cite{yang2019oboe} and \cite{kolda2009tensor}.
We denote \textit{vector}, \textit{matrix}, and \textit{tensor} variables
respectively by lowercase letters ($x$), capital letters ($X$) and Euler script letters ($\T{X}$).
The order of a tensor is the number of dimensions; matrices are order-two tensors.
Each dimension is called a mode.
Throughout this paper, all vectors are column vectors.
To denote a part of matrix or tensor, we use a colon to denote the dimension that is not fixed:
given a matrix $A \in \mathbb{R}^{m \times n}$, $A_{i, :}$ and $A_{:, j}$ denote the $i$th row and $j$th column of $A$, respectively.
A fiber is a one-dimensional section of a tensor $\T{X}$, defined by fixing every index but one;
for example, one fiber of the order-3 tensor $\T{X}$ is $\T{X}_{: jk}$.
Fibers of a tensor are analogous to rows and columns of a matrix.
A slice is an $(N-1)$-dimensional section of an order-N tensor $\T{X}$.
The mode-$n$ matricization of $\T{X}$, denoted as $\T{X}^{(n)}$, is a matrix whose columns are the mode-$n$ fibers of $\T{X}$.
For example, given an order-3 tensor $\T{X} \in \mathbb{R}^{I \times J \times K}$, $X^{(1)} \in \mathbb{R}^{I \times (J \times K)}$.
We denote the \emph{$n$-mode product} of a tensor $\T{X} \in \mathbb{R}^{I_1 \times I_2 \times \cdots I_N}$ with a matrix $U \in \mathbb{R}^{J \times I_n}$ by $\T{X} \times_n U \in \mathbb{R}^{I_1 \times \cdots I_{n-1} \times J \times I_{n+1} \times \cdots I_N}$; the $(i_1, i_2, \dots, i_{n-1}, j, i_{n+1}, \dots, i_N)$-th entry is $\Sigma_{i_n = 1}^{I_n} x_{i_1 i_2 \cdots i_{n-1} i_n i_{n+1} \cdots i_N} u_{j i_n}$.
Given two tensors with the same shape, we use $\odot$ to denote their entrywise product.
We define $[n] = \{1,\ldots,n\}$ for $n \in \mathbb{Z}$.
Given an ordered set $\mathcal{S} = \{ s_1, \ldots, s_k \}$ where $s_1 <  \ldots < s_k \in [n]$,
we write $A_\mathcal{:S} = [A_{:,s_1}, A_{:,s_2}, \ldots, A_{:,s_k} ]$; given an ordinary set $S$, we use $A_{:S}$ to denote $A_{:\mathcal{S'}}$, in which $\mathcal{S'}$ is the ordered version of set $S$.
The Frobenius norm of a tensor a tensor $\T{X} \in \mathbb{R}^{I_1 \times I_2 \times \cdots I_N}$ is $\| \T{X} \|_\text{F} = \sqrt{\sum_{i_1 \in [I_1], i_2 \in [I_2], \cdots, i_N \in [I_N]} x_{i_1 i_2 \cdots i_N}^2}$.

\myparagraph{Pipeline performance}
The performance of a machine learning pipeline is usually characterized by cross-validation error.
Given a dataset $\mathcal{D}$ and a pipeline $\mathcal{P}$,
we denote the error of $\mathcal{P}$ on $\mathcal{D}$ as $\mathcal{P}(\mathcal{D})$.
It is common practice to evaluate this error by cross-validating $\mathcal{P}$ on $\mathcal{D}$
with a certain number of folds (often 3, 5 or 10) and a fixed dataset partition.
We use $\mathcal{P}(\mathcal{D})$ to denote the cross-validation error we observe with a certain number of folds and a certain partition.

\myparagraph{Error tensor and error matrix}
Pipeline errors on training datasets form an \emph{error tensor}, which we denote as $\T{E}$.
In our experiments, $\T{E}$ is an order-6 tensor,
with 6 modes corresponding to datasets, imputers, encoders, standardizers, dimensionality reducers and estimators, respectively.
The $(i_1, i_2, \ldots, i_6)$-th entry of $\T{E}$ is the error of the pipeline
formed by composing the $i_2$-th imputer, $i_3$-th encoder, $i_4$-th standardizer, $i_5$-th dimensionality reducer,
and $i_6$-th estimator and evaluating this pipeline on the $i_1$-th dataset.
If a pipeline-dataset combination has been evaluated, we say the corresponding entry in the error tensor $\T{E}$ is observed.
The first unfolding of the error tensor, $\T{E}^{(1)}$, is called
the \emph{error matrix} $E$,
whose $ij$th entry $E_{ij} = \mathcal{P}_j(\mathcal{D}_i)$ is the error of pipeline $j$ on dataset $i$.

\myparagraph{Time target and time budget}
The time target refers to the anticipated time spent running models to infer
latent features of each fixed dimension and can be exceeded.
However, the runtime does not usually deviate much from the target
since our model runtime prediction works well.
The time budget refers to the total time limit for \textsc{Oboe} and is never exceeded.

\myparagraph{Ensemble}
An ensemble \cite{dietterich2000ensemble, breiman1996bagging, schapire2003boosting, wolpert1992stacked} combines a finite set of individual machine learning models into a single prediction model.
For simplicity, the combination method we use is majority voting for classification.
We define the \emph{candidate learner} to be individual machine learning pipelines that we select from to create the ensemble,
and \emph{base learner} to be pipelines that are included in the ensemble.
An ensemble of pipelines is itself a pipeline, but not a simple linear pipeline.
By creating ensembles of linear pipelines, our system can perform better than any linear pipeline.

\section{Oboe}
\label{sec:oboe}
\textsc{Oboe} selects among a set of machine learning estimators, and build pipelines with the same set of preprocessors. 
This is because the estimator is usually the most crucial part of a pipeline and is what practitioners usually spend a large amount of resources to choose.
However, the method described here is not limited to choosing among estimators; it is able to choose among machine learning models in general. 

\subsection{Overview}
\label{sec:oboe-overview}
\begin{figure*}
	\centering
	\includegraphics[width=.88\linewidth]{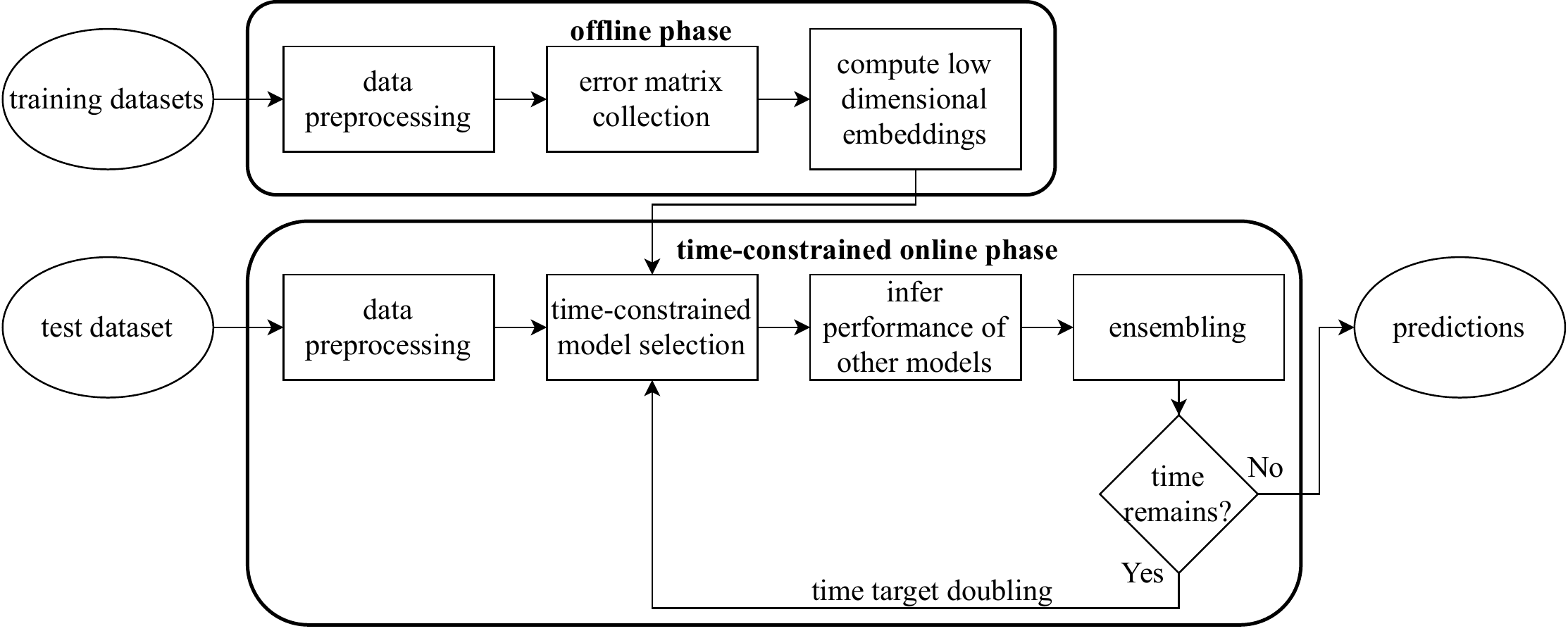}
	\caption{Diagram of data processing flow in the \textsc{Oboe} system.}
	\label{fig:oboe_diagram}
\end{figure*}

Shown in Figure~\ref{fig:oboe_diagram}, the meta-learning system for estimator selection, \textsc{Oboe}, has two phases.
In the \emph{offline} phase, we compute the performance of estimators on meta-training datasets and compute a surrogate model.
In the \emph{online} phase, we run a small number of estimators on the new meta-test dataset to help infer the performance of the rest.
The offline phase is executed only once and explores the space of estimator performance on meta-training datasets.
Time taken in this phase does not affect the runtime of \textsc{Oboe} on a new dataset;
the runtime experienced by user is that of the online phase.

One advantage of \textsc{Oboe} is that the vast majority of the time in the
online phase is spent training standard machine learning models,
while very little time is required to decide which models to sample.
Training these standard machine learning models requires running
algorithms on datasets with thousands of data points and features,
while the meta-learning task --- deciding which models to sample ---
requires only solving a small least-squares problem.

\subsubsection{Offline Stage}
The $(i,j)$th entry of error matrix $E \in \mathbb{R}^{m \times n}$ in \textsc{Oboe}, denoted as $E_{ij}$, records the performance of the $j$th estimator on the $i$th meta-training dataset.
We collect the error matrix using the \emph{balanced error rate} metric,
the average of false positive and false negative rates across different classes.
At the same time we record runtime of estimators on datasets. 
This is used to fit runtime predictors that will be addressed in Section~\ref{sec:runtime_prediction}.
Pseudocode for the offline phase is shown as Algorithm~\ref{alg:offline}.

\begin{algorithm}
	\caption{Offline Stage}
	\begin{algorithmic}[1]
		\Require{meta-training datasets $\{\mathcal{D}_i\}_{i=1}^m$, estimators $\{\mathcal{A}_j\}_{j=1}^n$, algorithm performance metric $\mathcal{M}$}
		
		\Ensure{error matrix $E$, runtime matrix $ T $, fitted runtime predictors $\{f_j\}_{j=1}^n$}
		\For{$i=1, 2, \ldots, m$}
		\State $ n^{\mathcal{D}_i}, p^{\mathcal{D}_i} \gets$ number of data points and features in $ \mathcal{D}_i $
		\For{$j=1,2,\ldots,n$}
		\State $E_{ij} \gets$ error of model $\mathcal{A}_j$ on dataset $\mathcal{D}_i$ according to metric $\mathcal{M}$
		\State $ T_{ij} \gets$ observed runtime for model $\mathcal{A}_j$ on dataset $\mathcal{D}_i$
		\EndFor
		\EndFor
		\For{$j=1, 2, \ldots, n$}
		\State fit $f_j = $ \texttt{fit\_runtime}$(n, p, T_j)$ (Section~\ref{sec:runtime_prediction})
		\EndFor
	\end{algorithmic}
	\label{alg:offline}
\end{algorithm}

\subsubsection{Online Stage} 
We repeatly double the time target of each round until we use up the total time budget. 
Each round is a subroutine of the entire online stage and is shown as Algorithm~\ref{alg:fit_one_round}.

\begin{itemize}[leftmargin=*]
	\item \textbf{Time-constrained model selection}
	Our active learning procedure selects a fast and informative collection of estimators to run on the meta-test dataset.
	\textsc{Oboe} uses the results of these fits to estimate the performance of all other models as accurately as possible.
	The procedure is as follows.
	First predict estimator runtime on the meta-test dataset using fitted runtime predictors.
	Then use experiment design to select a subset $\mathcal{S} $ of entries of $e$,
	the performance vector of the test dataset, to observe.
	The observed entries are used to compute $ \hat{x} $, an estimate of the latent meta-features of the test dataset,
	which in turn is used to predict every entry of $ e $.
	We build an ensemble out of pipelines predicted to perform well within the time target $ \tilde{\tau} $.
	This subroutine $ \tilde{A}= $\texttt{ensemble\_selection}$(\mathcal{S}, e_\mathcal{S}, z_\mathcal{S})$ takes as input the set of base learners $ \mathcal{S} $ with their cross-validation errors $ e_\mathcal{S} $ and predicted labels $ z_\mathcal{S} = \{z_s | s \in \mathcal{S}\}$,
	and outputs ensemble learner $ \tilde{A}$.
	The hyperparameters used by models in the ensemble can be tuned further, but in	our experiments we did not observe substantial improvements from further hyperparameter tuning.
	
	\begin{algorithm}
		\caption{Online phase, single round}
		\begin{algorithmic}[1]
			
			\Require{model latent meta-features $ \{y_j\}_{j=1}^n $, fitted runtime predictors $\{f_j\}_{j=1}^n$, training fold of the meta-test dataset $\mathcal{D}_\text{tr}$, number of best models $ N $ to select from the estimated performance vector, time target for this round $\tilde{\tau}$}
			\Ensure{ensemble learner $ \tilde{A}$}
			\Function{fit\_one\_round}{}
			\For{$j=1, 2, \ldots, n$}
			\State $\hat{t}_j \gets f_j(n^{\mathcal{D}_\text{tr}}, p^{\mathcal{D}_\text{tr}})$
			\EndFor
			\State $\mathcal{S} = \texttt{min\_variance\_ED}(\hat{t}, \{y_j\}_{j=1}^n, \tilde{\tau})$
			\For{$k=1, 2, \ldots, |\mathcal{S}|$}
			\State $e_{\mathcal{S}_k} \gets$ cross-validation error of model $ \mathcal{A}_{\mathcal{S}_k} $ on $ \mathcal{D}_\text{tr} $
			\EndFor
			\State $ \hat{x} \gets (\begin{bmatrix} y_{\mathcal{S}_1} & y_{\mathcal{S}_2} & \cdots & y_{\mathcal{S}_{|\mathcal{S}|}} \end{bmatrix}^\top)^\dagger e_S$
			\State $ \hat{e} \gets \begin{bmatrix} y_1 & y_2 & \cdots & y_n \end{bmatrix}^\top \hat{x}$
			\State $ \mathcal{T} \gets $ the $ N $ models with lowest predicted errors in $ \hat{e} $
			\For{$k=1, 2, \ldots, |\mathcal{T}|$}
			\State $e_{\mathcal{T}_k}, z_{\mathcal{T}_k} \gets$ cross-validation error of model $ \mathcal{A}_{\mathcal{T}_k} $ on $ \mathcal{D}_\text{tr} $
			\EndFor
			\State $\tilde{A} \gets $\texttt{ensemble\_selection}$(\mathcal{T}, e_\mathcal{T}, z_\mathcal{T})$
			\EndFunction
		\end{algorithmic}
		\label{alg:fit_one_round}
	\end{algorithm}
	
	\item \textbf{Time target doubling}\label{s-doubling}
	To select rank $ k $, \textsc{Oboe} starts with a small initial rank along with a small time target,
	and then doubles the time target for \texttt{fit\_one\_round} until the elapsed time reaches half of the total budget.
	The rank $ k $ increments by 1 if the validation error of the ensemble learner
	decreases after doubling the time target, and otherwise does not change.
	Since the matrices returned by PCA with rank $k$ are submatrices of those returned by PCA with rank $l$ for $l > k$, we can compute the factors as submatrices of the $ \textit{m} $-by-$ \textit{n} $ matrices returned by PCA with full rank $\min(m,n)$ \cite{golub2012matrix}.
	The pseudocode is shown as Algorithm~\ref{alg:onlinefitdoubling}.
\end{itemize}

\begin{algorithm}
	\caption{Online Stage}
	\begin{algorithmic}[1]
		\Require{error matrix $E$, runtime matrix $T$, meta-test dataset $\mathcal{D}$, total time budget $\tau$, fitted runtime predictors $\{f_j\}_{j=1}^n$}, initial time target $\tilde{\tau}_0$, initial approximate rank $ k_0 $
		
		\Ensure{ensemble learner $ \tilde{A} $}
		\Function{ensemble\_selection}{}
		\State $x_i, y_j \gets \argmin \sum_{i=1}^m \sum_{j=1}^n (E_{ij} - x_i^\top y_j)^2$, $x_i \in \mathbb{R}^{\min(m,n)}$ for $i \in [M]$ , $y_j \in \mathbb{R}^{\min(m,n)}$ for $j \in [N]$
		\State $ \mathcal{D}_\text{tr}, \mathcal{D}_\text{val}, \mathcal{D}_\text{te} \gets$ training, validation and test folds of $ \mathcal{D} $
		\State $ \tilde{\tau} \gets \tilde{\tau}_0$
		\State $ k \gets k_0$
		\While{$\tilde{\tau} \leq \tau/2$}
		\State $\{\tilde{y}_j\}_{j=1}^n \gets$ \textit{k}-dimensional subvectors of $ \{y_j\}_{j=1}^n $
		\State $ \tilde{A} \gets$ \texttt{fit\_one\_round}$(\{\tilde{y}_j\}_{j=1}^n, \{f_j\}_{j=1}^n, \mathcal{D}_\text{tr}, \tilde{\tau})$
		\State $ e_{\tilde{A}}' \gets \tilde{A}(\mathcal{D}_\text{val})$
		\If {$ e_{\tilde{A}}' < e_{\tilde{A}} $ }
		\State $ k \gets k+1 $
		\EndIf
		\State $ \tilde{\tau} \gets 2\tilde{\tau}$
		\State $ e_{\tilde{A}} \gets e_{\tilde{A}}' $
		
		\EndWhile
		\State \Return{$\tilde{A}$}
		\EndFunction
	\end{algorithmic}
	\label{alg:onlinefitdoubling}
\end{algorithm}

\subsection{Pipeline Performance Prediction}
It can be difficult to determine \textit{a priori} which meta-features to use
so that algorithms perform similarly well on datasets with similar meta-features.
Also, the computation of some landmarking meta-features can be expensive.
% (see Appendix~\ref{metafeaturetime}, Figure~\ref{fig:metafeature_calculation}).
To infer model performance on a dataset without any
expensive meta-feature calculations,
we use collaborative filtering to infer latent meta-features for datasets.

\begin{figure}
	\centering
	\includegraphics[width=\linewidth]{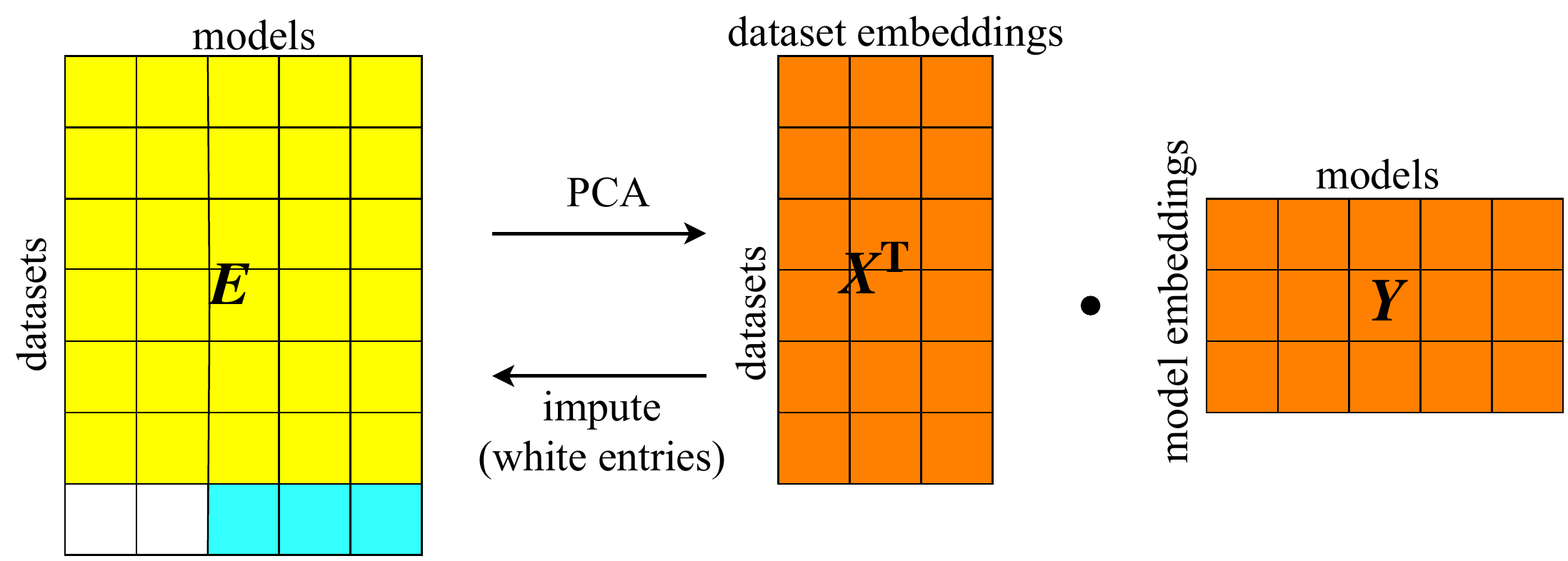}
	\caption{Model performance prediction by the error matrix $ E $ (yellow blocks only).
		Perform PCA on the error matrix (offline) to compute dataset ($X$) and
		model ($Y$) embeddings (orange blocks).
		Given a new dataset (row with white and blue blocks),
		pick a subset of models to observe (blue blocks).
		Use $ Y $ together with the observed models
		to impute the performance of the unobserved models on the new dataset (white blocks).}
	\label{fig:imputation}
\end{figure}

\begin{figure}
	\centering
	\includegraphics[width=.5\linewidth]{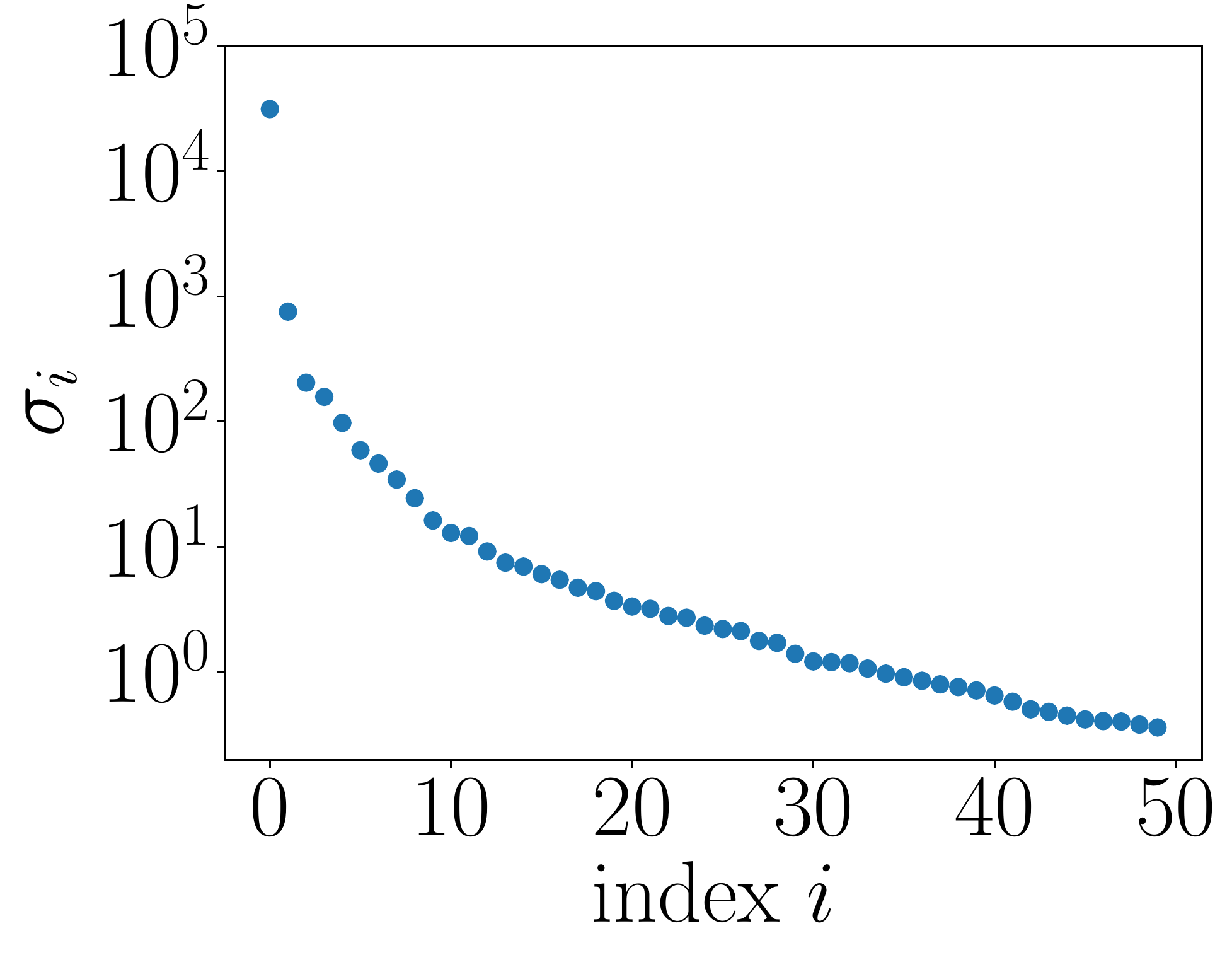}
	\caption{Singular value decay of an error matrix. The entries are calculated by 5-fold cross validation of machine pipelines (listed in Appendix~\ref{supp:configurations}, Table~\ref{table:pipeline_components}) with fixed components except estimators, on meta-training OpenML datasets (list in Appendix~\ref{supp:dataset_indices}).}
	\label{fig:singularValues}
\end{figure}

As shown in Figure~\ref{fig:imputation},
we construct an empirical error matrix $E \in \mathbb{R}^{m \times n}$,
where every entry $E_{ij}$ records the cross-validated error of model $j$ on dataset $i$.
Empirically, $ E $ has approximately low rank: Figure~\ref{fig:singularValues} shows
the singular values $\sigma_i(E)$ decay rapidly as a function of the index $i$.
This observation serves as foundation of our algorithm.
The value $E_{ij}$ provides a noisy but unbiased estimate
of the true performance of a model on the dataset:
$\mathbb{E} E_{ij} = \mathcal{A}_j(\mathcal{D}_i)$.

To denoise this estimate,
we approximate $E_{ij} \approx x_i^\top y_j$ where $x_i$ and $y_j$ minimize
$ \sum_{i=1}^m \sum_{j=1}^n (E_{ij} - x_i^\top y_j)^2$ with $x_i, y_j \in \mathbb{R}^k$ for $i \in [M]$ and $j \in [N]$; the solution is given by PCA.
Thus $x_i$ and $ y_j $ are the latent meta-features of dataset $i$ and model $j$, respectively.
The rank $k$ controls model fidelity: small $k$s give coarse approximations, while
large $k$s may overfit.
We use a doubling scheme to choose $k$ within time budget; see Section~\ref{s-doubling} for details.

Given a new meta-test dataset, we choose a subset $\mathcal{S} \subseteq [N]$ of models and observe performance $e_j$ of model $j$ for each $j \in \mathcal{S}$.
A good choice of $S$ balances information gain against time needed to run the models;
we discuss how to choose $S$ in Section~\ref{sec:ED}.
We then infer latent meta-features for the new dataset by solving the least squares problem: 
$\text{minimize} \sum_{j \in \mathcal{S}} (e_j - \hat{x}^\top y_j)^2$ with $\hat{x} \in \mathbb{R}^k$.
For all unobserved models,
we predict their performance as $\hat e_j = \hat{x}^\top y_j$ for $j \notin \mathcal{S}$.

\subsection{Runtime Prediction}
\label{sec:runtime_prediction}
Estimating model runtime allows us to trade off between running
slow, informative models and fast, less informative models.
We use a simple method to estimate runtimes, using polynomial regression on
$n^\mathcal{D}$ and $p^\mathcal{D}$, the numbers of data points and features in $\mathcal{D}$,
and their logarithms, since the theoretical complexities of machine learning algorithms we use
are $O \big((n^\mathcal{D})^3, (p^\mathcal{D})^3, (\log(n^\mathcal{D}))^3 \big)$.
Hence we fit an independent polynomial regression model for each model:
\[
f_j = \mbox{argmin}_{f_j \in \mathcal{F}}
\sum_{i=1}^M \left(
f_j(n^{\mathcal{D}_i}, p^{\mathcal{D}_i}, \log(n^{\mathcal{D}_i})) - t_j^{\mathcal{D}_i}
\right)^2,
\,
j \in [n]
\]
where $t_j^\mathcal{D}$ is the runtime of machine learning model $j$ on dataset $\mathcal{D}$,
and $\mathcal{F}$ is the set of all polynomials of order no more than 3.
We denote this procedure by $f_j=$ \texttt{fit\_runtime}$(n, p, t)$.

We observe that this model predicts runtime within a factor of two
for half of the machine learning models on more than 75\% meta-training OpenML datasets,
and within a factor of four for nearly all models,
as shown in Section~\ref{sec:runtime_prediction_performance}.

\subsection{Fast and Accurate Resource-Constrained Active Learning}
\label{sec:ED}
The methodology we describe here applies to not only machine learning estimators but also other models and pipelines in general.

Given a new dataset, our first problem is to select a subset of pipelines to fit, so that we may estimate the performance of other pipelines.
We use ideas from linear experiment design, which picks a subset of low-cost statistical trials to minimize
the variance of the resulting estimator, to make this selection.

Concretely, we estimate the embedding $x$ of the new dataset by linear regression.
Given the linear model as Equation~\ref{eq:tucker_matricized},
given known performance $e_S$ of a subset $S \subseteq [n]$ of pipelines on the new dataset, we have
\begin{equation}
e_S = (Y_{:S})^\top x + \epsilon
\label{eq:linear_regression}
\end{equation}
in which $Y$ collects the latent embeddings of pipeline performance, and $\epsilon$ is the error in this linear model.
For example, the error may be due to misspecification of the low Tucker factorization model for the error tensor.
We estimate $x$ by linear regression and denote the result as $\hat{x}$.
Then we estimate the performance of pipelines in $[n] \backslash S$ by the corresponding entries in $\hat{e} = Y^\top \hat{x}$.

Now we consider which $S$ to choose to accurately estimate $x$.
Suppose the error $\epsilon \sim \mathcal{N} (0, \sigma^2 I)$.
Using the linear regression model, Equation~\ref{eq:linear_regression},
we want to minimize the expected $\ell_2$ error
$E_\epsilon \|\hat{x} - x\|^2 = E_\epsilon \|\hat{x} - E_\epsilon \hat{x}\|^2 + \|E_\epsilon \hat{x} - x\|^2.$
Here, the second term is 0 since linear regression is unbiased,
and the first term is the covariance $\sigma^2 (Y Y^\top)^{-1}$ of the estimated embedding $\hat{x}$,
which is straightforward to compute.

We will first show how to constrain the \emph{number} of sampled pipelines.
Imagine we have enough time to run at most $m$ pipelines (and all pipelines run equally slowly).
Given pipeline embeddings $\{y_j\}_{j=1}^n$ (which we call \emph{design vectors} or \emph{designs}),
in which each $y_j \in \mathbb{R}^k$, we minimize a scalarization of the covariance
to obtain the (number-constrained) $D$-optimal experiment design problem
\begin{equation}
\begin{array}{ll}
\text{maximize} & \log \det \Big(\sum_{j \in S} y_j y_j^\top \Big) \\
\text{subject to} & |S| \leq m\\
& S \subseteq [n].
\end{array}
\label{eq:ED_number}
\end{equation}
Here, $\sum_{j \in S} y_j y_j^\top$, the inverse of (scaled) covariance matrix, is called the Fisher information matrix.

Obtaining an exact solution for a mixed-integer nonlinear optimization problem like Problem~\ref{eq:ED_number} is prohibitively expensive.
People have thus developed a variety of approaches to solve Problem~\ref{eq:ED_number} to certain accuracy, including convexification, greedy, local search, etc.
We will show the convexification approach in Section~\ref{sec:ED_number_cvx}, the greedy approach in Section~\ref{sec:ED_number_greedy},
and then develop a \emph{time}-constrained version that we use in practice in Section~\ref{sec:ED_time}.

\subsubsection{Convexification method for size-constrained experiment design}
\label{sec:ED_number_cvx}
Convexification is commonly used to solve a mixed-integer problem that can be easily transformed to a convex optimization by relaxation \cite{boyd2004convex, pukelsheim1993optimal, yang2019oboe}.

Define an indicator vector $v \in \{0,1\}^n$,
where entry $v_j$ indicates whether to fit model $j$.
Problem~\ref{eq:ED_number} is thus equivalent to the following formulation with variable $v \in \mathbb{R}^n$:
\begin{equation}
\begin{array}{ll}
\text{minimize} & \log \det \Big(\sum_{j=1}^n v_j y_j y_j^\top \Big)^{-1} \\
\text{subject to} & \sum\limits_{j=1}^n v_j \leq m \\
& v_j \in \{0, 1\}, \forall j \in [n]
\end{array}
\label{eq:ED_number_v}
\end{equation}
Now relax to allow $v \in [0,1]^n$ to allow for non-Boolean values and get the convex version for Problem~\ref{eq:ED_number_v}, and solve by a convex solver.
To binarize the solution, we may either choose the set of $m$ entries with largest values as $S$, or truncate at a certain threshold.

\subsubsection{Greedy method for size-constrained experiment design}
\label{sec:ED_number_greedy}
In practice, there may be a large number of models in our model selection problem, commonly at least tens of thousands. 
The convexification method is too slow in that case.
Also, the convexification method does not have suboptimality guarantee. 
Moreover, we can find better solutions with the greedy heuristic we present next.

Greedy methods form another popular approach to combinatorial optimization problems like Problem~\ref{eq:ED_number}.
Importantly, the objective function of Problem~\ref{eq:ED_number}, $f(S) = \log \det \Big(\sum_{j \in S} y_j y_j^\top \Big) $,
is submodular.
(Recall a set function $g: 2^V \rightarrow \mathbb{R}$ defined on a subset of $V$
is submodular if for every $A \subseteq B \subseteq V$ and every element $s \in V \backslash B$,
we have $g(A \cup \{s\}) - g(A) \geq g(B \cup \{s\}) - g(B)$.
This characterizes a ``diminishing return'' property.)
Given a size constraint, the submodular function maximization problem
\begin{equation}
\begin{array}{ll}
\text{maximize} & g(S) \\
\text{subject to} & S \subseteq V\\
& |S| \leq m
\end{array}
\label{eq:submodular_maximization}
\end{equation}
can be solved with a $1 - \frac{1}{e}$ approximation ratio \cite{nemhauser1978analysis} by the greedy approach:
in every step, add the single element that maximizes the increase in function value \cite{sharma2015greedy, krause2008near, krause2014submodular}.
In $D$-optimal experiment design, we can compute this increase efficiently using Lemma~\ref{lemma:det}.
\begin{lemma}[Matrix Determinant Lemma \cite{harville1998matrix, madan2019combinatorial}]
	\label{lemma:det}
	For any invertible matrix $A \in \mathbb{R}^{k \times k}$ and $a, b \in \mathbb{R}^k$,
	$$\det(A + a b^\top) = \det(A) (1 + b^\top A^{-1} a)$$
\end{lemma}
At the $t$-th step in our setting, with an already constructed Fisher information matrix $X_t = \sum_{j \in S} y_j y_j^\top$, we have
\begin{equation}
\argmax_{j \in [n] \backslash S} \det(X_t + y_j y_j^\top) = \argmax_{j \in [n] \backslash S} y_j^\top X_t^{-1} y_j.
\label{eq:argmax}
\end{equation}
Here, $y_j^\top X_t^{-1} y_j$ can be seen as the payoff for adding pipeline $j$.
From the $t$-th to the $(t+1)$-th step, with the selected design vector at the $t$-th step as $y_i$, we can update $X_t$ to $X_{t+1}=(X_t + y_t y_t^\top)$ by Lemma~\ref{lemma:sherman-morrison}:
\begin{lemma}[Sherman-Morrison formula \cite{sherman1950adjustment, hager1989updating}]
\label{lemma:sherman-morrison}
For any invertible matrix $A \in \mathbb{R}^{k \times k}$ and $a, b \in \mathbb{R}^k$, 
$$(A + a b^\top)^{-1} = A^{-1} - \frac{A^{-1} a b^\top A^{-1}}{1 + b^\top A^{-1} a}$$
\end{lemma}

Pseudocode for the greedy algorithm for Problem~\ref{eq:ED_number} is shown as Algorithm~\ref{alg:ED_number_greedy_without_repetition}, with per-iteration time complexity $O(k^3 + n k^2)$: it takes $O(k^3)$ (for a naive matrix multiplication algorithm) to update $X_t^{-1}$ and $O(n k^2)$ to choose the best pipeline to add.

\begin{algorithm}
	\caption{Greedy algorithm for size-constrained $D$-design}
	\label{alg:ED_number_greedy_without_repetition}
	\begin{algorithmic}[1]
		\Require{design vectors $\{y_j\}_{j=1}^n$, in which $y_j \in \mathbb{R}^k$; maximum number of selected pipelines $m$; initial set of designs $S_0 \subseteq [n]$, s.t. $X_0 = \sum_{j \in S_0} y_j y_j^\top$ is non-singular}
		\Ensure{The selected set of designs $S\subseteq [n]$}
		\Function{Greedy\_ED\_Number}{}
		\Do
		\State $S \gets S_0$
		\State $i \gets \text{argmax}_{j \in [n] \backslash S} y_j^\top X_t^{-1} y_j$
		
		\State $S \gets S \cup \{i\}$
		
		\State $X_{t+1} \gets X_t + y_i y_i^\top$
		\State $X_{t+1}^{-1} \gets \texttt{Sherman\_Morrison}(X_t, y_i)$
		\doWhile{$|S| \leq m$}
		\State \Return{$S$}
		\EndFunction
	\end{algorithmic}
\end{algorithm}

There remains the problem of how to select an initial set of designs $S$ to start from, such that $X_0 = \sum_{j \in S} y_j y_j^\top = Y_S Y_S^\top$ is non-singular.
This is equivalent to the problem of finding a subset of vectors in $\{y_j\}_{j=1}^n$ that can span $\mathbb{R}^k$.
We select this sized-$k$ subset $S_0$ to be the first $k$ pivot columns from QR factorization with column pivoting \cite{golub2012matrix, gu1996efficient} on $Y$, with time complexity $O((n+k)k^2)$.

\subsubsection{Greedy method for time-constrained experiment design}
\label{sec:ED_time}
We here move on to the realistic case that we face in AutoML pipeline selection: which pipelines should we select to gain an accurate estimate of the entire pipeline space?
In this setting, each pipeline is associated with a different cost.
We characterize the cost as running time, and form the time-constrained version of experiment design as

\begin{equation}
\begin{array}{ll}
\text{maximize} & \log \det \Big(\sum_{j \in S} y_j y_j^\top \Big) \\
\text{subject to} & \sum_{j \in S} \hat{t}_j \leq \tau\\
& S \subseteq [n],
\end{array}
\label{eq:ED_time}
\end{equation}

\noindent in which $\{\hat{t}_i\}_{i=1}^n$ are the estimated pipeline running times, $\tau$ is the runtime limit.
The payoff of adding design $j$ in the $t$-th step can thus be formulated as $\frac{y_i^\top X^{-1} y_i}{\hat{t}_i}$.
We have as Algorithm~\ref{alg:ED_time_greedy_without_repetition} the greedy method to solve Problem~\ref{eq:ED_time}.
\begin{algorithm}
	\caption{Greedy algorithm for time-constrained $D$-design}
	\label{alg:ED_time_greedy_without_repetition}
	\begin{algorithmic}[1]
		\Require{design vectors $\{y_j\}_{j=1}^n$, in which $y_j \in \mathbb{R}^k$; estimated running time of all pipelines $\{\hat{t}_i\}_{i=1}^n$; maximum running time $\tau$; initial set of designs $S_0 \subseteq [n]$, s.t. $X_0 = \sum_{j \in S_0} y_j y_j^\top$ is non-singular}
		\Ensure{The selected set of designs $S\subseteq [n]$}
		\Function{Greedy\_ED\_Time}{}
		\Do
		\State $S \gets S_0$
		\State $i \gets \text{argmax}_{j \in [n] \backslash S} \frac{y_j^\top X_t^{-1} y_j}{\hat{t}_j}$
		
		\State $S \gets S \cup \{i\}$
		
		\State $X_{t+1} \gets X_t + y_i y_i^\top$
		\doWhile{$\sum_{i \in S} \hat{t}_i \leq \tau$}
		\State \Return{$S$}
		\EndFunction
	\end{algorithmic}
\end{algorithm}

The initialization problem is solved similarly by the QR method.
Given runtime limit $\tau$, we select among columns with corresponding pipelines predicted to finish within $\frac{\tau}{2k}$.
Pseudocode for this initialization algorithm is shown as Algorithm~\ref{alg:ED_time_qr_initialization}.
\begin{algorithm}
	\caption{Initialization of the greedy algorithm for time-constrained $D$-design, by QR factorization with column pivoting}
	\label{alg:ED_time_qr_initialization}
	\begin{algorithmic}[1]
		\Require{design vectors $\{y_j\}_{j=1}^n$, in which $y_j \in \mathbb{R}^k$; (predicted) running time of all pipelines $\{\hat{t}_i\}_{i=1}^n$; maximum running time $\tau$}
		\Ensure{A subset of designs $S_0\subseteq [n]$ for Algorithm~\ref{alg:ED_time_greedy_without_repetition} initialization}
		\Function{QR\_Initialization}{}
		\State $S_\text{valid} \gets \{i \in [n]: \hat{t}_i \leq \frac{\tau}{2k}\}$
		\State $S_0 \gets \emptyset$, $\hat{t}_\text{sum} \gets 0$
		\If{$|S_\text{valid}| < k$}
		\Comment{Case 1}
		\Do
		\State $i \gets \text{argmin}_{j \in [n] \backslash S} \hat{t}_j$
		\State $S_0 \gets S_0 \cup \{i\}$
		\State $\hat{t}_\text{sum} \gets \hat{t}_\text{sum} + \hat{t}_j$
		\doWhile{$\hat{t}_\text{sum} \leq \tau$}
		\Else
		\Comment{Case 2}
		\State $S_0 \gets \text{QR\_with\_column\_pivoting}(Y_{S_\text{valid}})[:k]$
		\EndIf
		\State \Return{$S_0$}
		\EndFunction
		
	\end{algorithmic}
\end{algorithm}

A corner case of Algorithm~\ref{alg:ED_time_qr_initialization}, shown as Case 1, is that there are not enough pipelines predicted to be able to finish within time limit.
This corresponds to the case that the runtime limit is relatively small compared to the time of fitting pipelines on current dataset.
In this case we greedily select the fast pipelines and do not run Algorithm~\ref{alg:ED_time_greedy_without_repetition} afterwards.

As a side note, the assumption that performance of different pipelines are predicted with equal
variance is not quite realistic, especially when some components have much more pipelines than others.
If the variance is known (but unequal), we obtain a weighted least squares problem.
In the error matrix $E$, we can estimate the variance of prediction error of each pipeline $j \in [n]$
by the sample variance of $E_{:j} - X^\top Y_{:j}$ and select the promising pipelines with the goal of minimizing the rescaled covariance.
Practically, however, this rescaled method does not systematically improve on the standard least squares approach in our experiments (shown in Appendix~\ref{supp:WLS}),
so we retrench to the simpler approach.

\section{TensorOboe}
\label{sec:tensoroboe}
\textsc{Oboe} focuses on the selection of machine learning estimators. 
In practice, preprocessors also have a large influence on model quality; the selection of which incurs high cost as well.
In this section, to model and utilize the combinatorial structure among pipeline components, we move on to the AutoML system for selecting among pipelines with different imputers, dimensionality reducers, estimators, etc. 
The name \textsc{TensorOboe} comes from the tensor surrogate model we use.
As we will demonstrate by calculation in Section~\ref{sec:tensor-completion} and by experiments in Section~\ref{sec:completion_experiments}, the tensor model is better at modeling the combinatorial structure of pipeline components than matrices from tensor matricization. 
 
\subsection{Overview}
\label{sec:tensoroboe-system}
\textsc{TensorOboe} has two phases that are similar to those of \textsc{Oboe}.
In the offline phase, we compute the performance of pipelines on meta-training datasets
and compute a surrogate model.
In the online phase, we run a small number of pipelines on the new meta-test dataset to
specialize the surrogate model and identify promising pipelines.

\myparagraph{Offline Stage}
We collect a partially observed error tensor using the approach described in Section~\ref{sec:tensor-collection}
to limit the total runtime of the offline phase.
We complete and factorize the error tensor $\T{E}$ using the EM-Tucker algorithm, shown as Algorithm~\ref{alg:EM-Tucker},
with dataset and estimator ranks empirically chosen to be the ones that give low reconstruction error,
described in Section~\ref{sec:completion_experiments}.

\myparagraph{Online Stage}
Online, given a new dataset $\mathcal{D}$ with $n^\mathcal{D}$ data points and $p^\mathcal{D}$ features,
we first predict the running time of each pipeline by a simple model:
order-3 polynomial regression on $n^\mathcal{D}$ and $p^\mathcal{D}$ and their logarithms.
This simple model works well because the time to fit the estimator dominates the time to fit the pipeline,
and the theoretical complexities of estimators we use have no higher order terms \cite{hutter2014algorithm, yang2019oboe}.

\begin{figure}
	\centering
	\includegraphics[width=\linewidth]{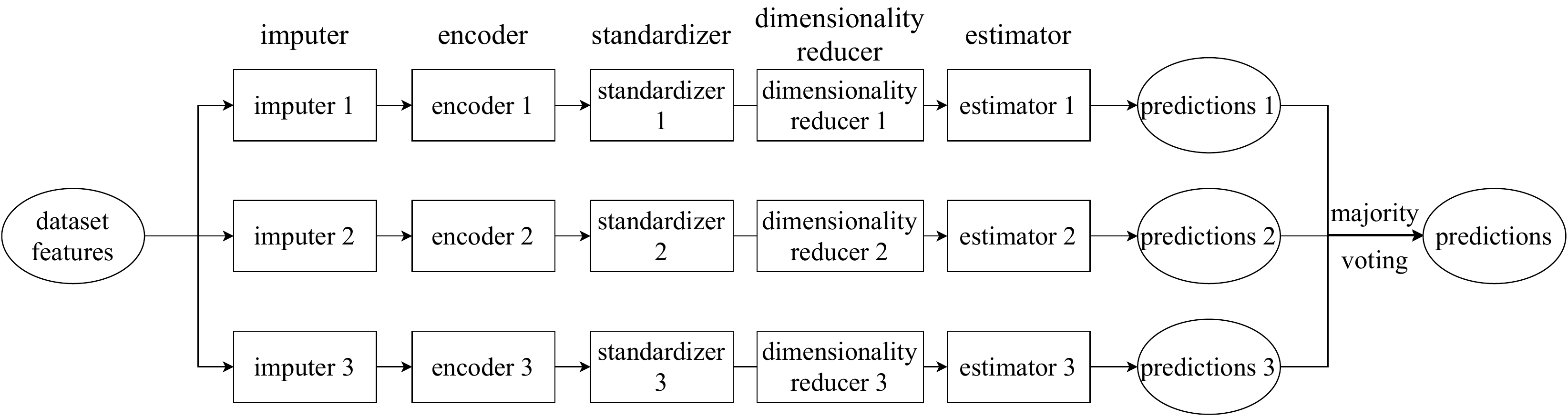}
	\caption{A pipeline ensemble with 3 base learners.}
	\label{fig:ensemble_example}
\end{figure}

The initial dataset and estimator ranks are set to the number of principal components
that capture 97\% of the energy in the respective tensor matricizations.
We double the runtime budget at each iteration and increment the estimator rank if the performance improves.
In each iteration, we select informative pipelines by formulating the time-constrained experiment design problem as Problem~\ref{eq:ED_time} and solve by the greedy approach as Algorithm~\ref{alg:ED_time_greedy_without_repetition}.
Then we build ensembles
whose base learners are the 5 pipelines with the best cross-validation error.
An ensemble can improve on the performance of the best base pipeline.
An example is shown as Figure~\ref{fig:ensemble_example}.

\subsection{Tensor Collection for Meta-Training}
\label{sec:tensor-collection}
In the meta-training phase of meta-learning, meta-training data is generally assumed to be already available or cheap to collect.
Given the large number of possible pipeline combinations, though, collecting meta-training data can be prohibitively expensive.
As an example, even if it takes one minute on average to evaluate each pipeline on each dataset,
evaluating $20,000$ pipelines on $200$ meta-training datasets would take more than $7$ years of CPU time.
This motivates us to use tensor completion to limit the time spent collecting meta-training data efficiently
while preserving accuracy of our surrogate model.

We collect pipeline performance in a biased way:
using 3-fold cross-validation, we only evaluate pipelines that complete within 120 seconds.
This rule gives a missing ratio of $3.3\%$.
Notice that the entries are not missing uniformly at random:
for example, some datasets are large and expensive to evaluate;
our training data systematically lacks data from these large datasets.
Nevertheless, we will show how to infer these entries
using tensor completion in Section~\ref{subsec:tensor_factorization_and_low_rank},
and demonstrate in Section~\ref{sec:completion_experiments} that the method performs well despite bias.

\subsection{Tensor Factorization and Rank}
\label{subsec:tensor_factorization_and_low_rank}
The meta-training phase constructs the error tensor $\T{E}$.
In the meta-test phase, we see a new dataset, corresponding to a new slice of $\T{E}$.
To learn about the slice efficiently, use a low rank tensor factorization
to predict all the entries in this slice from a subset of informative entries

Unlike matrices, there are many incompatible notions of tensor rank and low rank tensor decompositions,
including CANDECOMP/PARAFAC (CP) \cite{carroll1970analysis, harshman1970foundations}, Tucker \cite{tucker1966some},
and tensor-train \cite{oseledets2011tensor}.
Each emphasizes a different aspect of the tensor low rank property.

\begin{figure}
	\centering
	\includegraphics[width=8cm]{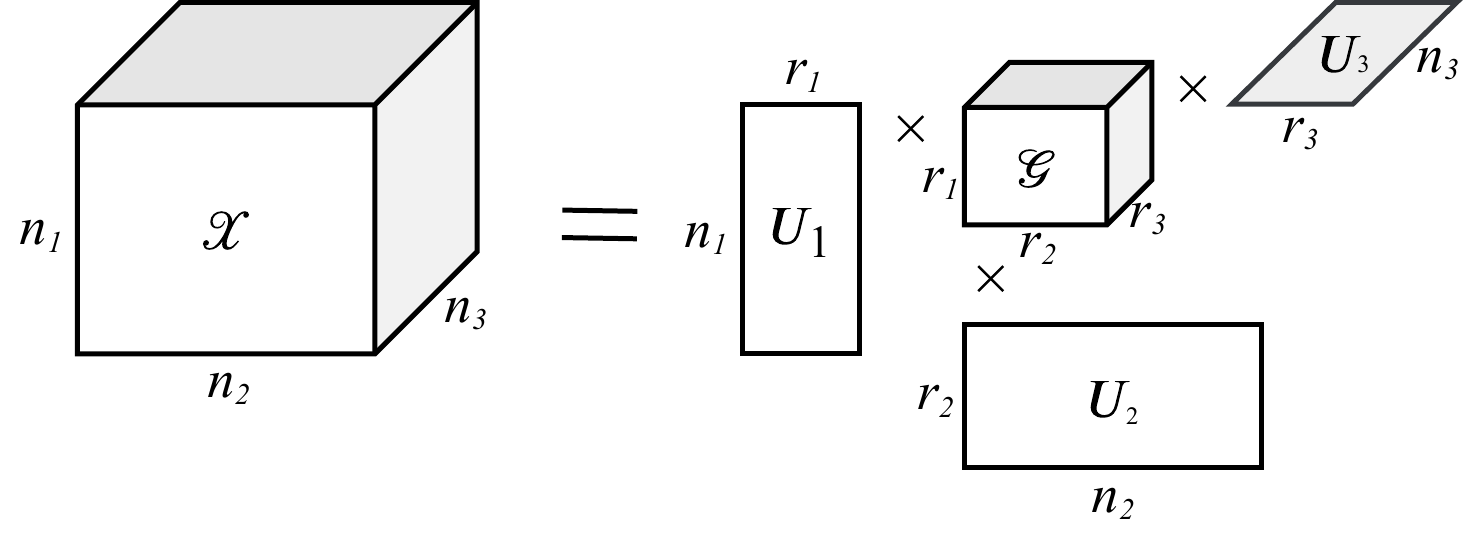}
	\caption{Tucker decomposition on an order-3 tensor.}
	\label{fig:tucker}
\end{figure}
In this paper, we use Tucker decomposition; an example on an order-3 tensor is shown as Figure~\ref{fig:tucker}.
As a form of higher-order PCA,
Tucker decomposes a tensor into the product of a \emph{core tensor} and several \emph{factor matrices},
one for each mode  \cite{kolda2009tensor}.
In our setting of order-6 tensors, Tucker decomposition of $\T{E}$ is
\begin{equation}
\T{E} \approx \hat{\T{E}} = \T{G}\times_1 U_1 \times \cdots \times_6 U_6
\label{eq:tucker}
\end{equation}
with core tensor $\T{G} \in \mathbb{R}^{r_1 \times r_2 \times \cdots \times r_6}$ and factor matrices $U_i \in \mathbb{R}^{n_i \times r_i}$, $i \in \{1, 2, \dots, 6\}$ with orthonormal columns.
Each factor matrix corresponds to the respective pipeline component; $\hat{\T{E}}$ is linear in the factor matrices.
Each factor matrix can thus be viewed as embedding the corresponding pipeline component,
with pipeline embeddings as columns of $Y = (\T{G} \times_2 U_2 \times \cdots \times_6 U_6)_1 \in \mathbb{R}^{r_1 \times (\Pi_{i=2}^6 n_i)}$, the mode-1 matricization of the product.
We can use this observation to approximately factor the error matrix $E$, using Equation~\ref{eq:tucker}, as
\begin{equation}
X^\top Y \approx E \in \mathbb{R}^{n_1 \times (\Pi_{i=2}^6 n_i)}
\label{eq:tucker_matricized}
\end{equation}
in which $X \in \mathbb{R}^{r_1 \times n_1}$ and $Y \in \mathbb{R}^{r_1 \times (\Pi_{i=2}^6 n_i)}$ are dataset and pipeline embeddings, respectively.

We evaluate the factorization performance by relative error, defined as $\fnorm{\boldsymbol{\Omega} \odot (\T{X} - \T{\hat{X}})}^2 / \fnorm{\boldsymbol{\Omega} \odot \T{X}}^2$:
$\T{X}$ is the true tensor, $\T{\hat{X}}$ is the predicted tensor, and $\boldsymbol{\Omega}$ is a binary tensor that has the same shape as $\T{X}$ and $\T{\hat{X}}$.
$\boldsymbol{\Omega}_{i_1, i_2, ..., i_6} = 1$ if the $(i_1, i_2, ..., i_6)$-th entry is in the set of entries we care about, and 0 otherwise.

\begin{figure}
	\centering
	\begin{subfigure}[c]{.38\linewidth}
		\includegraphics[width=\linewidth]{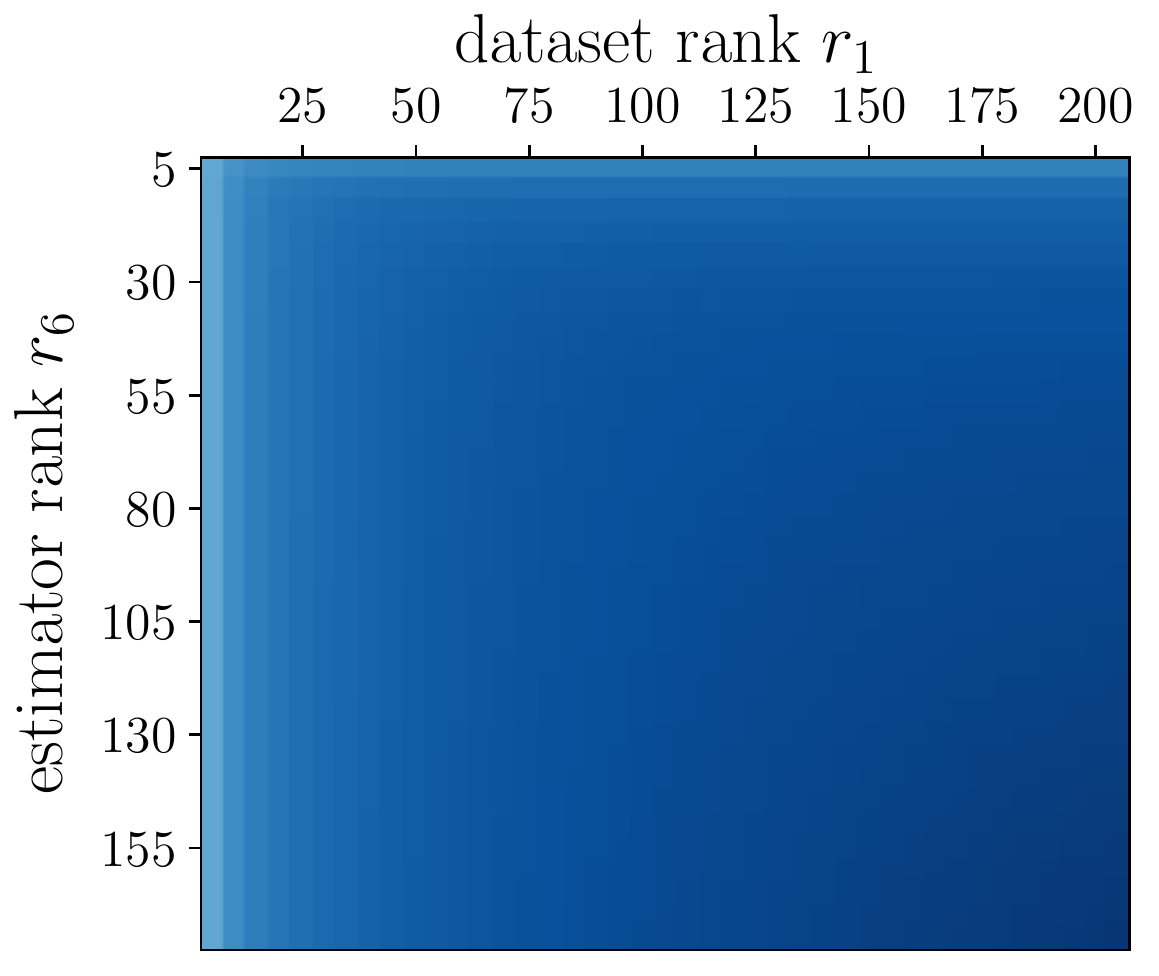}
		\caption{Training}
		\label{fig:relative_error_decay_training}
	\end{subfigure}%
	\hspace{.01\linewidth}
	\begin{subfigure}[c]{.38\linewidth}
		\includegraphics[width=\linewidth]{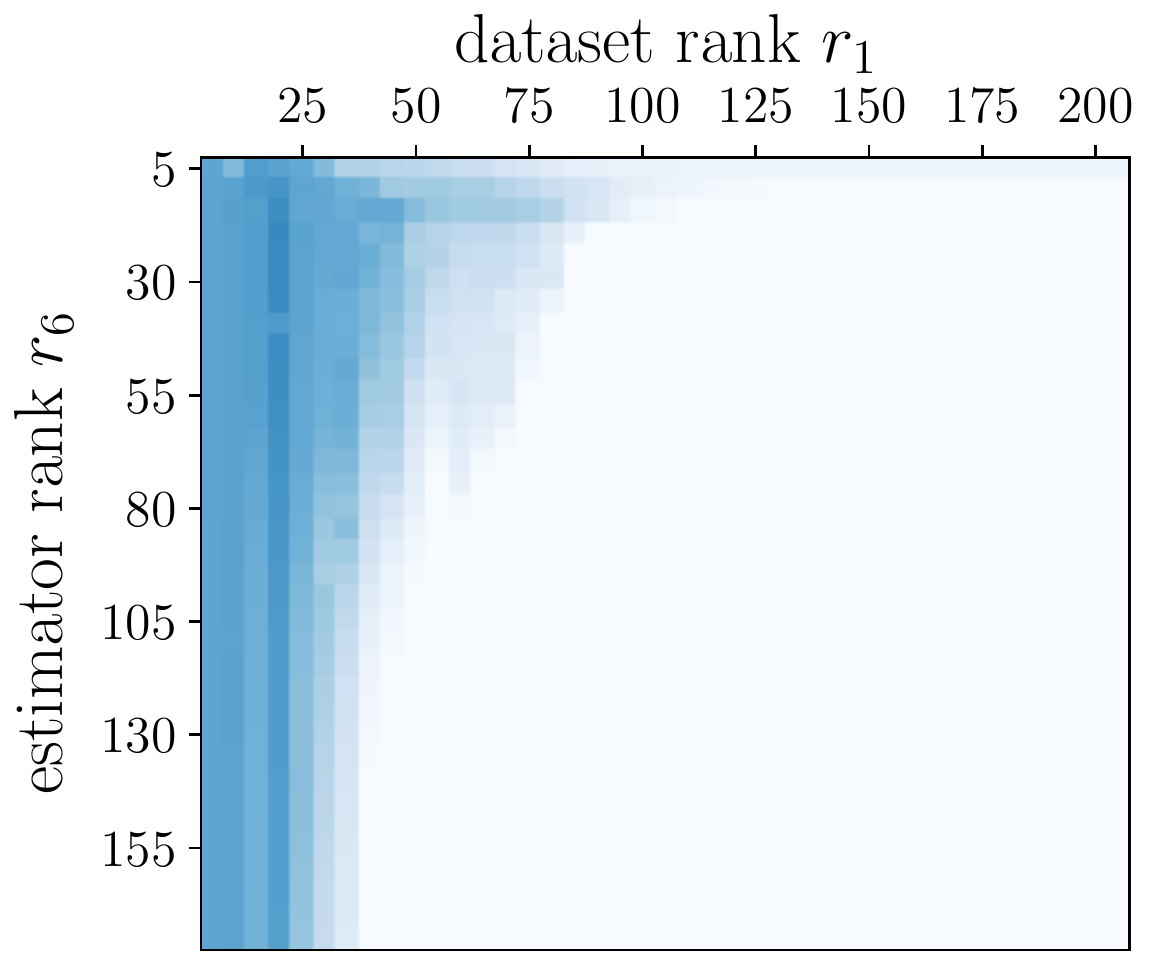}
		\caption{Test}
		\label{fig:relative_error_decay_test}
	\end{subfigure}%
	\hspace{.03\linewidth}
	\begin{subfigure}[c]{.1\linewidth}
	\includegraphics[width=\linewidth]{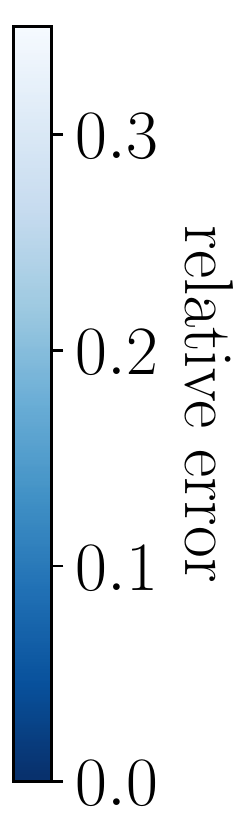}
\end{subfigure}%
	\caption{Relative error heatmaps when varying ranks in dataset and estimator dimensions.
		Here, training entries are the ones with runtime less than 90 seconds; the test entries are the ones with runtime between 90 and 120 seconds.}
	\label{fig:relative_error_decay}
\end{figure}

Figure~\ref{fig:relative_error_decay} shows the low Tucker rank factorization fits the error tensor well: the training relative errors are small at high ranks, while the test relative errors are small at low ranks.

\subsection{Tensor Completion}
\label{sec:tensor-completion}
To infer missing entries in the error tensor we collected, namely the entries that take more than the time threshold to evaluate, we use the expectation-maximization (EM) \cite{dempster1977maximum, song2019tensor} approach together with Tucker decomposition in each step, which we call EM-Tucker and present as Algorithm~\ref{alg:EM-Tucker}.

\begin{algorithm}
	\caption{EM-Tucker algorithm for tensor completion}
	\label{alg:EM-Tucker}
	\begin{algorithmic}[1]
		
		\Require{order-$n$ error tensor $\T{E}$ with missing entries, Tucker ranks $[r_1, \dots, r_n]$}
		\Ensure{imputed error tensor $\T{E}$}
		\State $\T{E}_\text{obs} \leftarrow \T{E}$
		\State $\boldsymbol{\Omega} \leftarrow $ observed entries in $\T{E}_\text{obs} $
		\Do
		\State $\T{G}$, $\{U_i\}_{i=1}^n$ $\leftarrow$ Tucker($\T{E}$, ranks=$[r_1, \dots, r_n]$)
		
		\State $\T{E}_\text{pred} \leftarrow \T{G}\times_1 U_1 \times \cdots \times_n U_n$
		
		\State $\T{E} \leftarrow \boldsymbol{\Omega} \odot \T{E}_\text{obs} + (1-\boldsymbol{\Omega}) \odot \T{E}_\text{pred}$
		
		\doWhile{not converged}
	\end{algorithmic}
\end{algorithm}

In Algorithm~\ref{alg:EM-Tucker}, $\boldsymbol{\Omega}$ is similarly a binary tensor that indicates whether each entry of the error tensor $\T{E}$ is observed or not.
$\boldsymbol{\Omega}$ has the same shape as the original error tensor, with the corresponding entry $\boldsymbol{\Omega}_{i_1, i_2, ..., i_n} = 1$ if the $(i_1, i_2, ..., i_n)$-th entry of the error tensor is observed, and 0 otherwise.
In our experiments, the algorithm is regarded as converged when the decrease of relative error is less than $0.01\%$, or the number of iterations reaches 1000.

Why bother with tensor completion?
To recover the missing entries of a tensor, we can also perform matrix completion after matricization or perform matrix completion on every slice separately.
Tensors are more constrained and so provide better fits to sparse and noisy data.
Consider a tensor $\T{X} \in \mathbb{R}^{I_1 \times I_2 \times \cdots I_N}$ with Tucker ranks $[r_1,r_2,\ldots,r_n]$, where $I_1=I_2=\ldots=I_n=I$ and $r_1=r_2=\ldots= r_n=r$.
The number of degrees of freedom of $\T{X}$, which is the minimum number of entries required to recover $\T{X}$, is  $r^n+n(rI-r^2) =: m_0$.
If we unfold $\T{X}$ to $X\in\mathbb{R}^{I\times I^{n-1}}$, the number of degrees of freedom of $X$ is $(I+I^{n-1}-r)r =: m_1$.
If we treat every slice of $\T{X}$ separately,  the number of degrees of freedom is $I^{n-2}(2rI-r^2) =: m_2$.
Therefore, when $r<I$, we have $m_0<m_1<m_2$, which means we need fewer parameters to determine $\T{X}$, compared to the matricization and union of slices.
Thus, tensor completion may outperform matrix completion on $\T{X}$ with the same number of observed entries.

\section{Kernel Oboe}\label{sec:KernelOboe}
In the offline stage of \textsc{TensorOboe}, we used the EM-Tucker algorithm to complete the error tensor we collected.
It relies on the assumption that the error tensor has low Tucker rank, which means the embeddings of each pipeline component lie in a low dimensional space.
This assumption has been numerically verified in Figure~\ref{fig:tucker}.
In this section, we turn to the kernel approach that assumes that pipeline embeddings lie in a union of subspaces. 
We show an efficient kernel method for meta-training that is able to achieve a lower error for tensor completion, and in most cases it gives a better performance for pipeline selection.

\textsc{KernelOboe} differs from \textsc{TensorOboe} in the tensor completion approach in the offline stage. 
All the other steps, including the entire online stage, remain the same.

\subsection{High Rank Matrix Completion for OBOE}
Low-rank matrix completion methods are not effective in recovering the missing entries of high-rank matrices even when the data have some low-dimensional latent structures. 
As pointed out by Fan et al. \cite{fan2019polynomial},  data points drawn from each of the following three models can form a high-rank or even full-rank matrix: a union of low-dimensional subspaces, a low-dimensional nonlinear manifolds, and a union of low-dimensional manifolds. 
Fan et al. \cite{fan2019polynomial} showed that, if the columns of a matrix $M\in\mathbb{R}^{m\times n}$ are generated by $k$ different polynomials of order at most $\alpha$ on a $d$-dimensional latent variable, the rank of $M$ can be as high as $\min\lbrace k\binom{d+\alpha}{\alpha},m,n\rbrace$. 
Thus, when either $k$ or $\alpha$ is large, $M$ can be full-rank. Performing a  $q$-order polynomial feature map $\phi$ on each column of $M$, they showed that 
\begin{align}
\textup{rank}(\phi(M))\leq\min\left\lbrace k\binom{d+\alpha q}{\alpha q},\binom{m+q}{q},n\right\rbrace.
\end{align}
Thus, when $n$ is sufficiently large and $d\ll m$, $\phi(M)$ is low-rank, compared to its side lengths. 
Consequently, the missing entries of $M$ can be recovered by minimizing the rank (or tractable rank regularizers) of $\phi(M)$. 
The algorithms proposed in \cite{fan2019polynomial}  require performing SVD on an $n\times n$ matrix in every iteration and hence are not scalable to very large matrices.

In \cite{fan2019online},  the authors proposed an efficient high rank matrix completion method called kernelized factorization matrix completion (KFMC). 
We propose to apply KFMC to the error matrix and solve
\begin{align}\label{Eq.KFMC}
\mathop{\textup{minimize}}_{\hat{E},D,Z}\ \  &\dfrac{1}{2} \fnorm{\phi(\hat{E})-\phi(D)Z}^2+\dfrac{\alpha}{2}\fnorm{\phi(D)}^2+\dfrac{\beta}{2}\fnorm{Z}^2\\ \nonumber
\textup{subject to}\ \ &\hat{E}_{ij}=E_{ij},\ (i,j)\in\Omega,
\end{align}
where $D\in\mathbb{R}^{m\times r}$, $Z\in\mathbb{R}^{r\times n}$,  $r$ denotes the approximate rank of $\phi(E)$, and $\Omega$ consists of the locations of observed entries. 
Let $\phi$ be the feature map implicitly determined by the Gaussian RBF kernel $k(x,y)=\exp\left(-\Vert x-y\Vert^2/(2\sigma^2)\right)$, then $\Vert\phi(D)\Vert_F^2\equiv r$, which means we can remove $\dfrac{\alpha}{2}\Vert\phi(D)\Vert_F^2$ from \eqref{Eq.KFMC}  without changing the solution. 
When $n$ is too large (e.g. $n\geq 10000$), we may use the online algorithm (Algorithm 2) of \cite{fan2019online} to solve \eqref{Eq.KFMC}. To further improve efficiency, we extend the online algorithm to a mini-batch algorithm, which is shown in Algorithm \ref{alg.mbkfmc}. 
In the algorithm, $\bar{\Omega}$ consists of the locations of the unknown entries of $E$; $K_{AB}$ denotes the kernel matrix computed from matrices $A$ and $B$.

\begin{algorithm}[h]
\caption{Mini-batch KFMC for Meta-training}
\label{alg.mbkfmc}
\begin{algorithmic}[1]
\Require
$E_{\Omega}$, $r$, $\sigma$, $\beta$, $\eta$, $n_{\text{batch}}$, $n_{\text{iter}}$, $n_{\text{pass}}$
\State Initialize: $D\sim\mathcal{N}(0,1)$, $\widehat{\Delta}_{{D}}={0}$
\State Split the columns of $E$ into $E_1,E_2,\ldots,E_{n_{\text{batch}}}$
\For{$u=1$ to $n_{\text{pass}}$}
\For{$j=1$ to $n_{\text{batch}}$}
\State $l=0$, $\widehat{{\Delta}}_{E_j}={0}$, ${C}=({K}_{{D}{D}}+\beta{I}_r)^{-1}$
\Repeat
\State $l\leftarrow l+1$ and ${Z}_j={C}{K}_{E_j{D}}^\top$
\State Compute ${\Delta}_{E_j}$ using (26) of \cite{fan2019online}
\State $\widehat{{\Delta}}_{E_j}\leftarrow\eta\widehat{{\Delta}}_{E_j}+{\Delta}_{E_j}$
\State $[{E}_j]_{\bar{\Omega}_j}\leftarrow[{E}_j]_{\bar{\Omega}_j}-[\widehat{{\Delta}}_{E_j}]_{\bar{\Omega}_j}$
\Until {converged or $l=n_{\text{iter}}$}
\State Compute ${\Delta}_{{D}}$ using (31) of \cite{fan2019online}
\State $\widehat{\Delta}_{{D}}\leftarrow\eta\widehat{{\Delta}}_{{D}}+{\Delta}_{{D}}$ and ${D}\leftarrow{D}-\widehat{{\Delta}}_{D}$
\EndFor
\EndFor
\Ensure $E$, $D$, $Z$
\end{algorithmic}
\end{algorithm}

Notice the fact that the columns of $E$ consist of the classification errors given by a few classifiers on the datasets and each classifier may correspond to a low-dimensional subspace of $\mathbb{R}^m$ or a low-dimensional nonlinear manifold embedded in $\mathbb{R}^m$. Therefore, we may model $E$ more accurately by a high-rank matrix with low-dimensional latent structure. 
Compared to low-rank matrix completion, KFMC is thus expected to provide higher prediction accuracy of pipeline performance.

\subsection{Performance Prediction on New Dataset by KFMC}
The $D$ and $Z$ given by Algorithm \ref{alg.mbkfmc} can be used to recover the unknown entries of a new column of $E$ or a new row of $E$, which correspond to predicting the performance of a new pipeline on the existing datasets or the performance of existing pipelines on a new dataset. 

To predict the performance of a new pipeline, we can just use the out-of-sample-extension of KFMC, namely the Algorithm 3 of \cite{fan2019online}. 
To predict the performance of the pipelines on a new dataset, we may need to solve the following problem
\begin{align}\label{Eq.KFMC_ose0}
\mathop{\textup{minimize}}_{\hat{E}',D'}\ \  &\dfrac{1}{2}\fnorm{\phi(\hat{E}')-\phi(D')Z}^2\\ \nonumber
\textup{subject to}\ \ &\hat{E}'_{ij}=E'_{ij},\ (i,j)\in\Omega',
\end{align}
where $E'=\left[\begin{matrix}
E\\
e_{\text{new}}
\end{matrix}\right]$, 
$D'=\left[\begin{matrix}
D\\
d_{\text{new}}
\end{matrix}\right]$, $\Omega'=\Omega\cup\omega_{\text{new}}$, and $\omega_{\text{new}}$ denotes the locations of known entries of $e_{\text{new}}$. 
Obviously, solving \eqref{Eq.KFMC_ose0} is time-consuming, even when we fix $E$ and $D$.

We can write $\phi(x)=c
\begin{bmatrix}
x\\
\tilde{\phi}(x)
\end{bmatrix}$, where $\tilde{\phi}$ consists of the features of $\phi$ with $x$ itself removed and $c$ is a constant. 
Then we have
$$\phi(E')=\phi(D')Z\ \ \Longrightarrow\ \ E'=D'Z.$$
Therefore we propose to solve
\begin{align}\label{Eq.KFMC_ose1}
\mathop{\textup{minimize}}_{\hat{e}_{\text{new}},d_{\text{new}}}\ \  &\dfrac{1}{2}\norm{ \hat{e}_{\text{new}}-d_{\text{new}}Z}^2+\dfrac{\alpha}{2} \norm{d_{\text{new}}}^2\\ \nonumber
\textup{subject to}\ \ &\hat{e}_{j}=e_{j},\ j\in\omega.
\end{align}

Compared to Problem~\ref{Eq.KFMC_ose0}, Problem~\ref{Eq.KFMC_ose1} drops the higher order terms from objective function and has the advantage of a closed-form solution
$[\hat{e}_{\text{new}}]_{\bar{\omega}}=d_{\text{new}}Z_{:,\bar{\omega}}$, where $d_{\text{new}}=e_{\omega}Z_{:,\omega}^\top(Z_{:,\omega}Z_{:,\omega}^\top+\alpha I_r)$.

\subsection{Experimental Design with KFMC}
Since the prediction model used in \eqref{Eq.KFMC_ose1} is similar to \eqref{eq:linear_regression}, the experimental design of KFMC can be easily adapted from Section \ref{sec:ED}.

\section{Experiments}
\label{sec:experiments}
All the code is in the GitHub repository at \url{https://github.com/udellgroup/oboe}.
We use Intel\textsuperscript{\textregistered} Xeon\textsuperscript{\textregistered} E7-4850 v4 2.10GHz CPUs for evaluation.
Offline, we collect cross-validated pipeline performance on meta-training datasets: 215 OpenML  \cite{OpenML2013, OpenMLPython2019} classification datasets with number of data points between 150 and 10,000, chosen alphabetically, and listed in Appendix~\ref{supp:dataset_indices}.
In \textsc{TensorOboe} and \textsc{KernelOboe}, pipelines are combinations of the machine learning components shown in Appendix~\ref{supp:search_space}, Table~\ref{table:pipeline_components}, which lists 4 data imputers, 2 encoders, 2 standardizers, 8 dimensionality reducers and 183 estimators.
Here we have 23,424 possible pipeline combinations in total, but we are demonstrating the ability of our methods to quickly explore the combinatorial space.
Because of ensembling, our methods are able to handle a much larger number of combinations.
In \textsc{Oboe}, we only vary estimators among the above 183, after preprocessing all datasets by imputing with mean (for numerical features) or mode (categorical features), one-hot encoding categorical features and then standardizing all features to have zero mean and unit variance.
Thus the search space of \textsc{Oboe} is a slice of those in \textsc{TensorOboe} and \textsc{KernelOboe}.

\subsection{Comparison of Time-Constrained AutoML Systems}
\label{sec:framework_comparison}
In this section, we demonstrate the performance of $\textsc{Oboe}$ and $\textsc{TensorOboe}$ as AutoML systems.
A naive approach for pipeline selection is to choose the one that on average performs the best among all meta-training datasets, which we call the baseline pipeline.
Given the pipeline selection problem, it is common for human practitioners to try out the best pipeline at the very beginning.
On our meta-training datasets, the baseline pipeline is: impute missing entries with the mode, encode categorical features as integers, standardize each feature, remove features with 0 variance, and classify by gradient boosting with learning rate 0.25 and maximum depth 3.
The baseline pipeline has an average ranking of 1568 among all 23,424 pipelines across all 215 meta-training datasets.

\begin{figure}
	\centering
	\includegraphics[width=\linewidth]{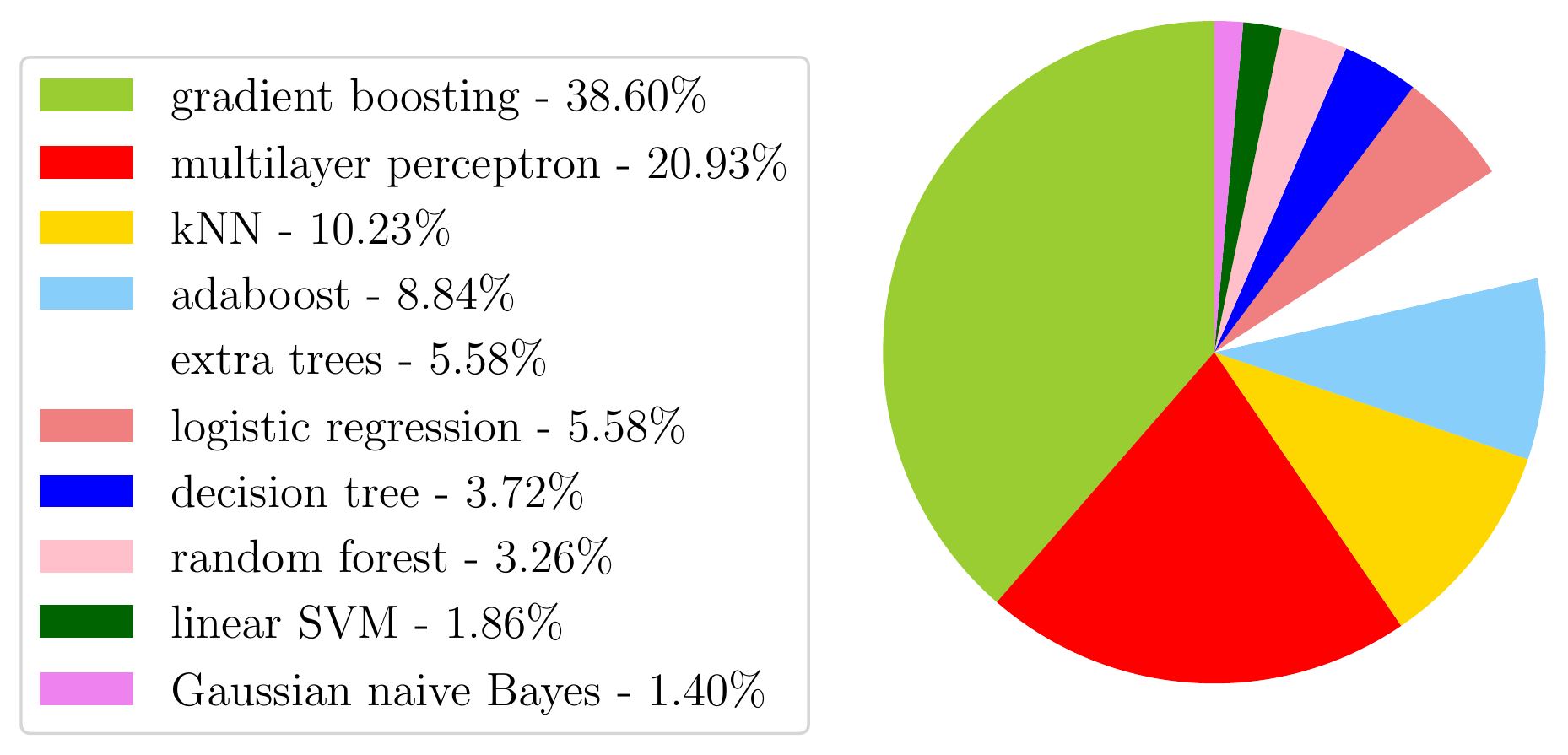}
	\caption{Which estimators work best?
		Distribution of estimator types in best pipelines on meta-training datasets.}
	\label{fig:best_estimators}
\end{figure}

Human practitioners may also reduce the number of trials by choosing certain pipeline components to be the type that performs the best on average.
Figure~\ref{fig:best_estimators}, however, shows that although some estimator types (gradient boosting and multilayer perceptron) are commonly seen among the best pipelines, no estimator type uniformly dominates the rest.

\begin{figure}
	\centering
	\begin{subfigure}[t]{.85\linewidth}
		\includegraphics[width=\linewidth]{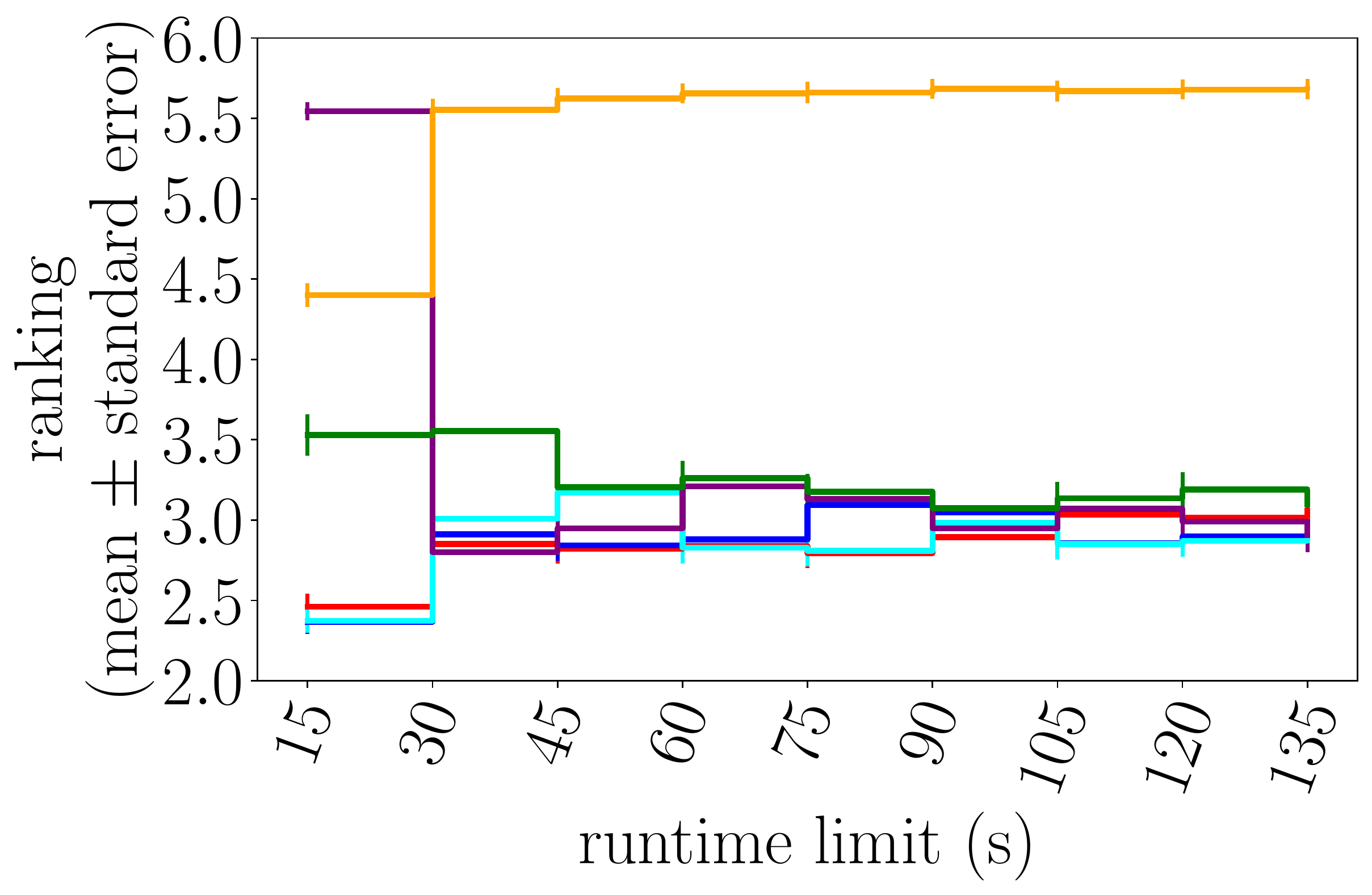}
		\caption{OpenML (meta-LOOCV)}
		\label{fig:ranking_OpenML}
	\end{subfigure}%

	\begin{subfigure}[t]{.85\linewidth}
		\includegraphics[width=\linewidth]{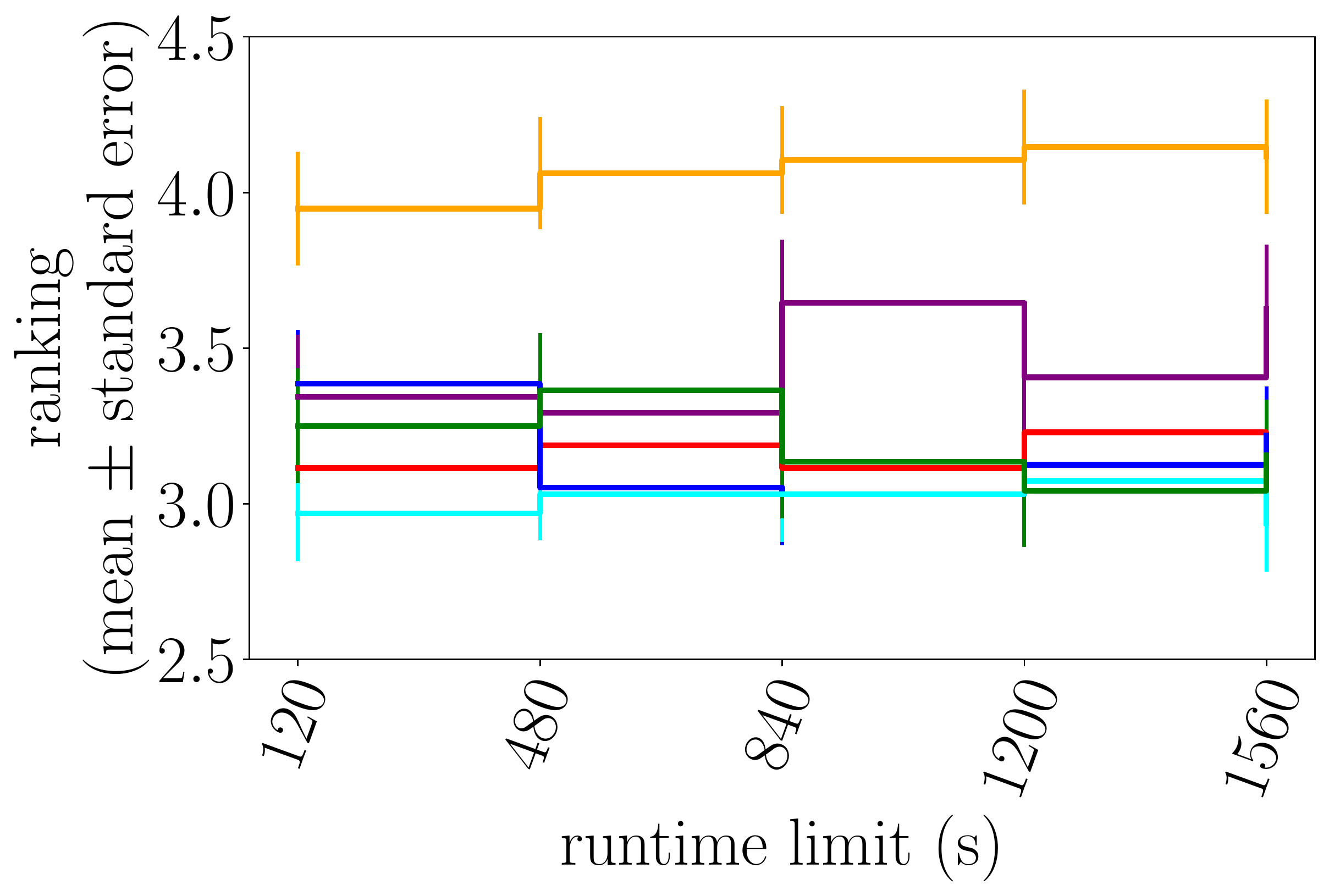}
		\caption{UCI (meta-test)}
		\label{fig:ranking_UCI}
	\end{subfigure}%

	\begin{subfigure}[t]{.65\linewidth}
	\includegraphics[width=\linewidth]{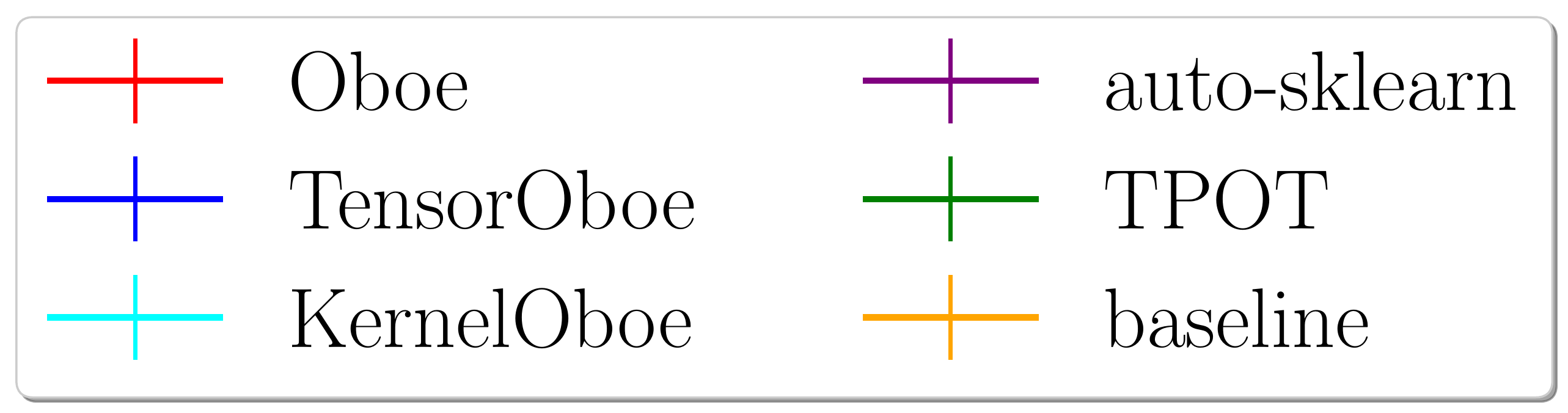}
\end{subfigure}%

	\caption{
		System rankings of AutoML systems for pipeline search in a time-constrained setting, vs the baseline pipeline.
		We meta-train on OpenML classification datasets and meta-test on UCI classification datasets \cite{UCI2019}.
		Until the first time the system can produce a pipeline,
		we classify every data point with the most common class label.
		Lower ranks are better.}
	\label{fig:comparison_of_frameworks}
\end{figure}

In the first experiment, we compare \textsc{Oboe}, \textsc{TensorOboe} and \textsc{KernelOboe} that work on the original error matrix and error tensor with auto-sklearn \cite{feurer2015efficient}, TPOT \cite{olson2019tpot}, and the baseline pipeline.
To ensure fair comparisons, we use a single CPU core for each AutoML system.
We allow each to choose from the same primitives.
The comparison plot is shown as Figure~\ref{fig:comparison_of_frameworks}.
We can see that:
\begin{enumerate}[label=\arabic*, wide, labelwidth=!, labelindent=0pt]
	\item All AutoML frameworks are able to construct pipelines that outperform the baseline on average once the method returns a pipeline (for auto-sklearn, this takes 30 seconds).
	\item The \textsc{Oboe} variants on average outperform the competing methods and produces meaningful pipeline configurations fastest.
	\item With much longer running time on meta-test datasets, shown in Figure~\ref{fig:ranking_UCI}, our $\textsc{Oboe}$ and $\textsc{TensorOboe}$ still outperforms in most cases.
	This shows that they are able to approximate the hyperparameter landscape accurately in general, which will be discussed to greater detail in Section~\ref{sec:hyperparameter_landscape}.
\end{enumerate}

\begin{figure}
	\centering
	\begin{subfigure}[t]{.85\linewidth}
		\includegraphics[width=\linewidth]{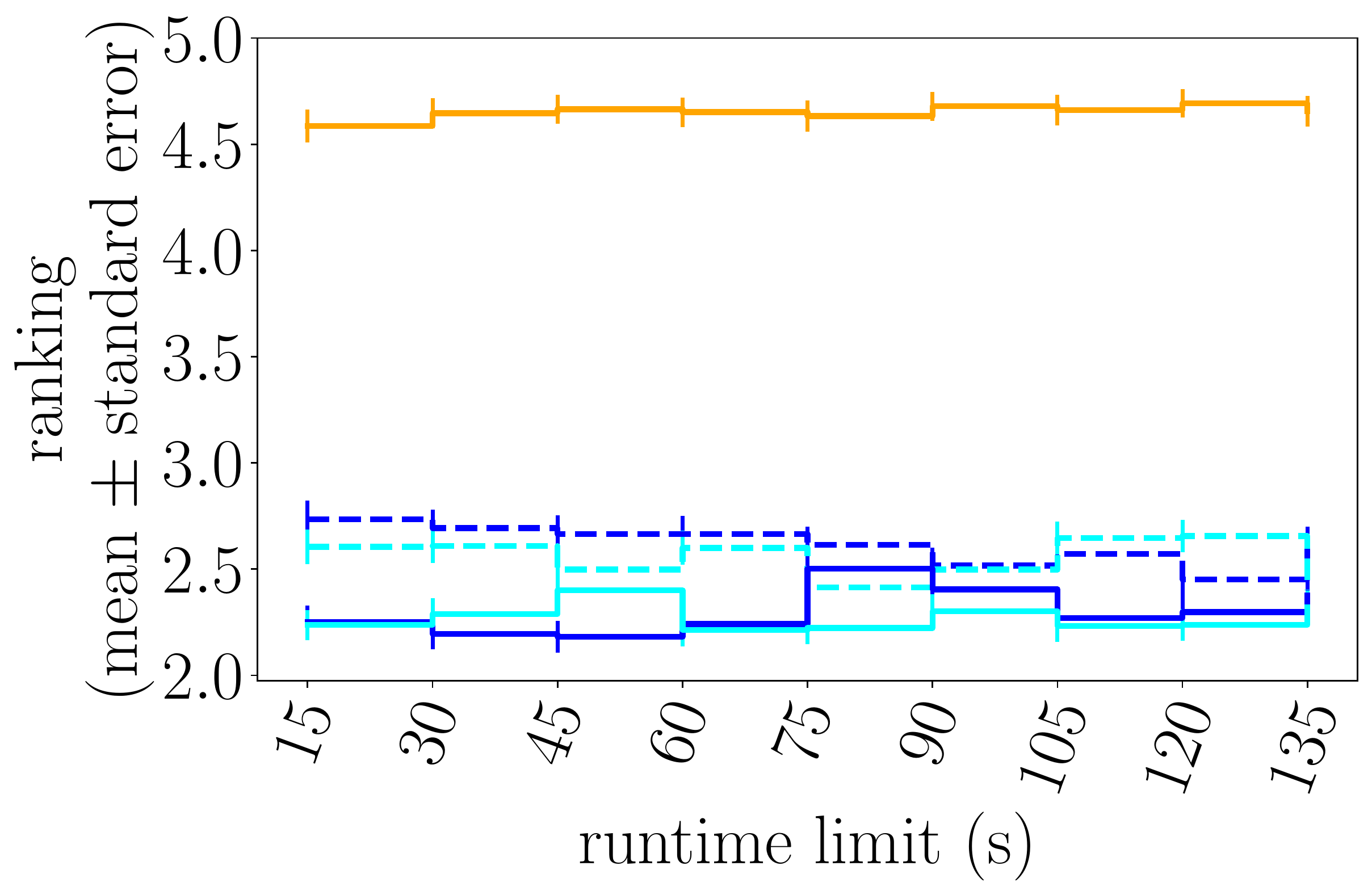}
		\caption{OpenML (meta-LOOCV)}
		\label{fig:ranking_OpenML_subsampled}
	\end{subfigure}%
	
	\begin{subfigure}[t]{.85\linewidth}
		\includegraphics[width=\linewidth]{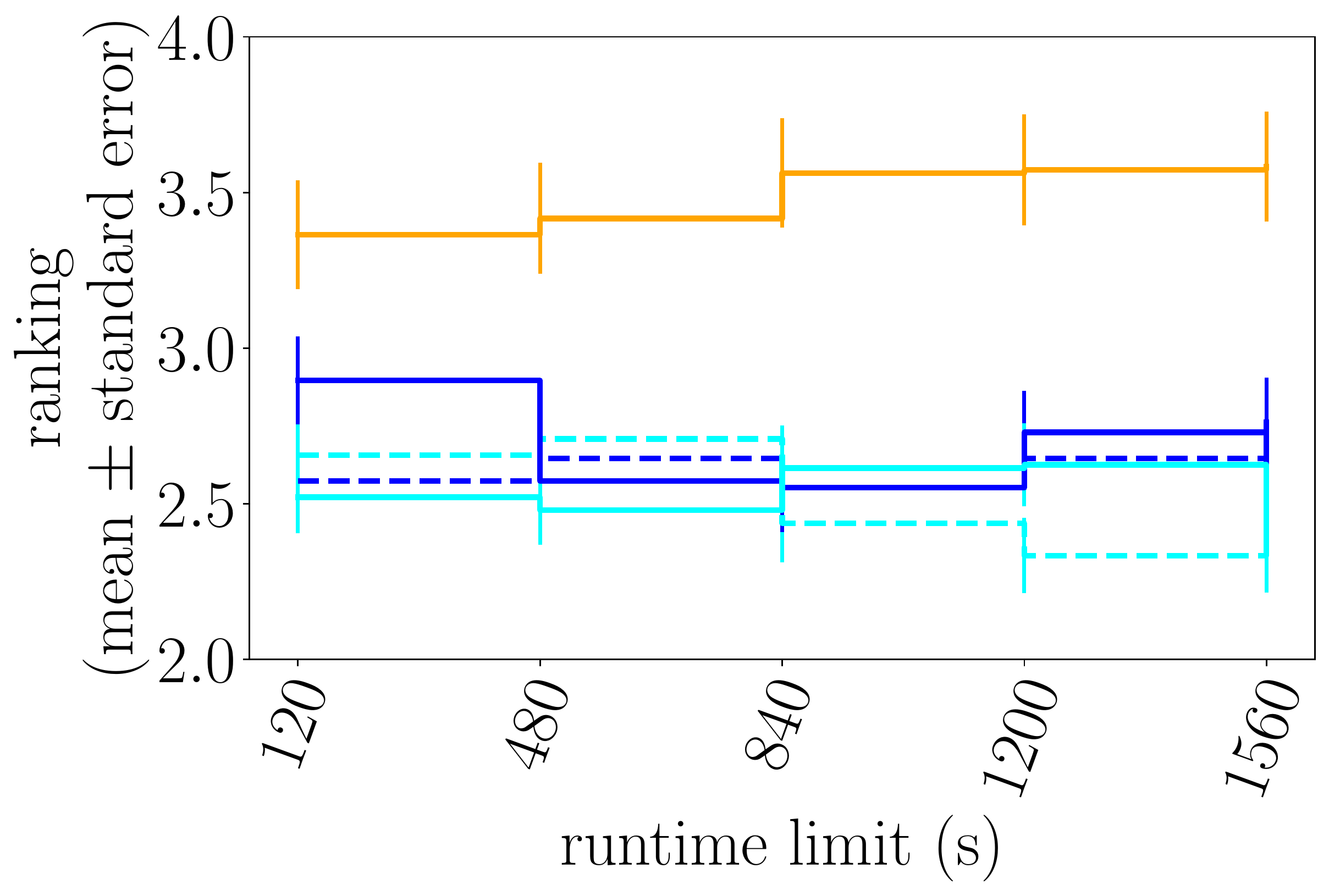}
		\caption{UCI (meta-test)}
		\label{fig:ranking_UCI_subsampled}
	\end{subfigure}%
	
	\begin{subfigure}[t]{.85\linewidth}
		\includegraphics[width=\linewidth]{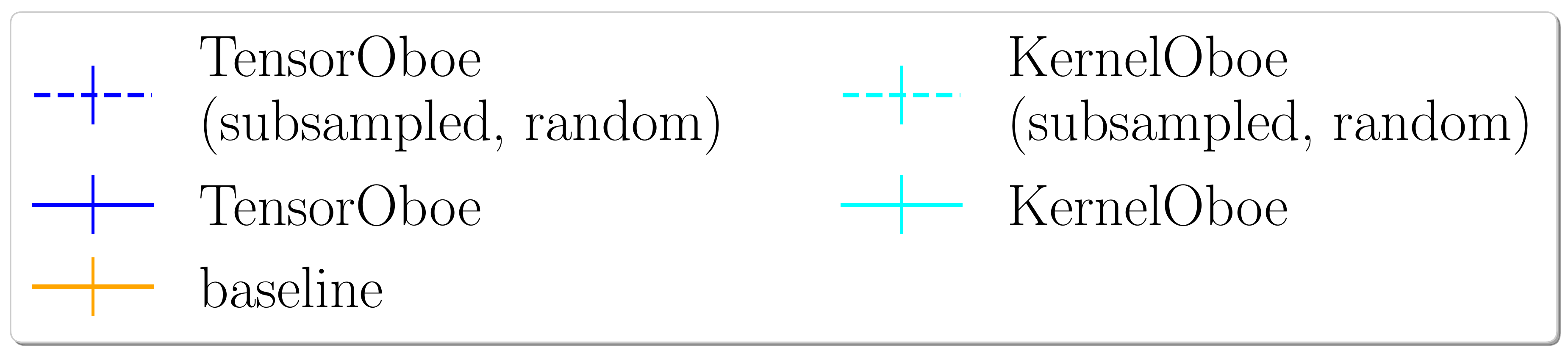}
	\end{subfigure}%
	
	\caption{
		System rankings of \textsc{TensorOboe} and \textsc{KernelOboe} for pipeline search with randomly subsampled error tensors, vs the baseline pipeline.
		The error tensor used by the dashed lines marked as ``subsampled, random'' lacks 90\% of the pipeline performance that were known in the original tensor.
Dataset and estimator ranks are set to be 40 to complete the error tensor in \textsc{TensorOboe}.}
	\label{fig:comparison_of_frameworks_subsampled}
\end{figure}

In the second experiment, we compare the performance of \textsc{TensorOboe} and \textsc{KernelOboe} that take randomly subsampled error tensors as input, and show the result in Figure~\ref{fig:comparison_of_frameworks_subsampled}. 
Offline, given the collected error tensor $\T{E}$ with missing ratio $3.3\%$, we randomly subsample $10\%$ of its known entries and regard the rest as missing, resulting in a missing ratio of $90.3\%$.
We denote this tensor as $\T{E}_\Omega$. 
We then use Algorithm~\ref{alg:EM-Tucker} and Algorithm~\ref{alg.mbkfmc}, respectively, to complete $\T{E}_\Omega$, and use the respective results as the error tensor for \textsc{TensorOboe} and \textsc{KernelOboe}.
We can see that:
\begin{enumerate}[label=\arabic*, wide, labelwidth=!, labelindent=0pt]
\item Results from the subsampled error tensor are in general worse than those from the original error tensor at smaller running times, indicating the abundance of meta-training data does help in the scarcity of computational power. 
\item Results from the subsampled error tensor have higher rankings (corresponding to smaller ranking values) at longer runtime thresholds, indicating that the meta-testing process gains more knowledge of the pipeline performance space with more time.
This opens up the horizon of a meta-training process that collects pipeline performance data on a larger space with more model types for each of the pipeline components.
\end{enumerate}

\begin{figure}
	\centering
	\begin{subfigure}[t]{.85\linewidth}
		\includegraphics[width=\linewidth]{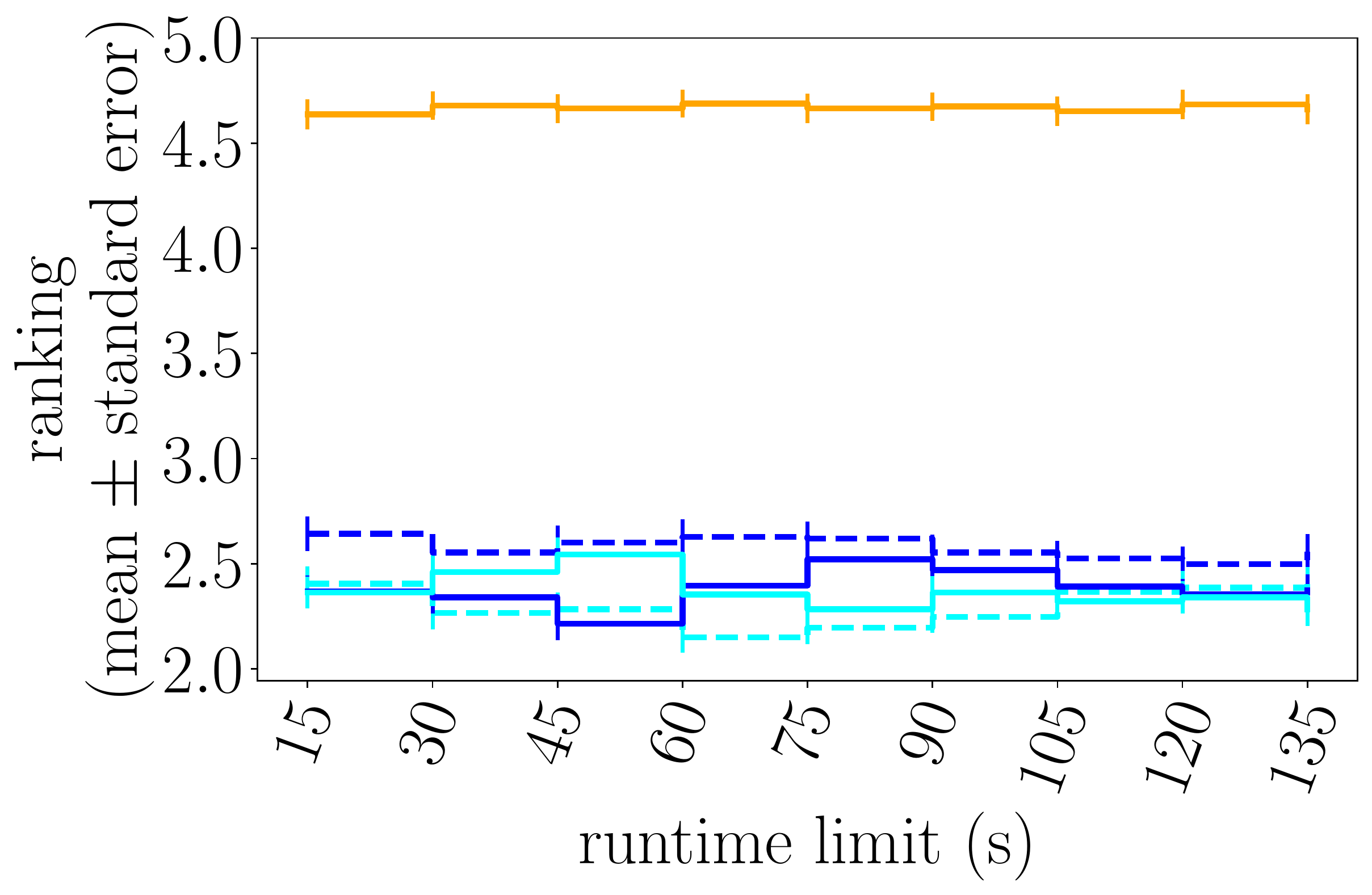}
		\caption{OpenML (meta-LOOCV)}
		\label{fig:ranking_OpenML_time}
	\end{subfigure}%
	
	\begin{subfigure}[t]{.85\linewidth}
		\includegraphics[width=\linewidth]{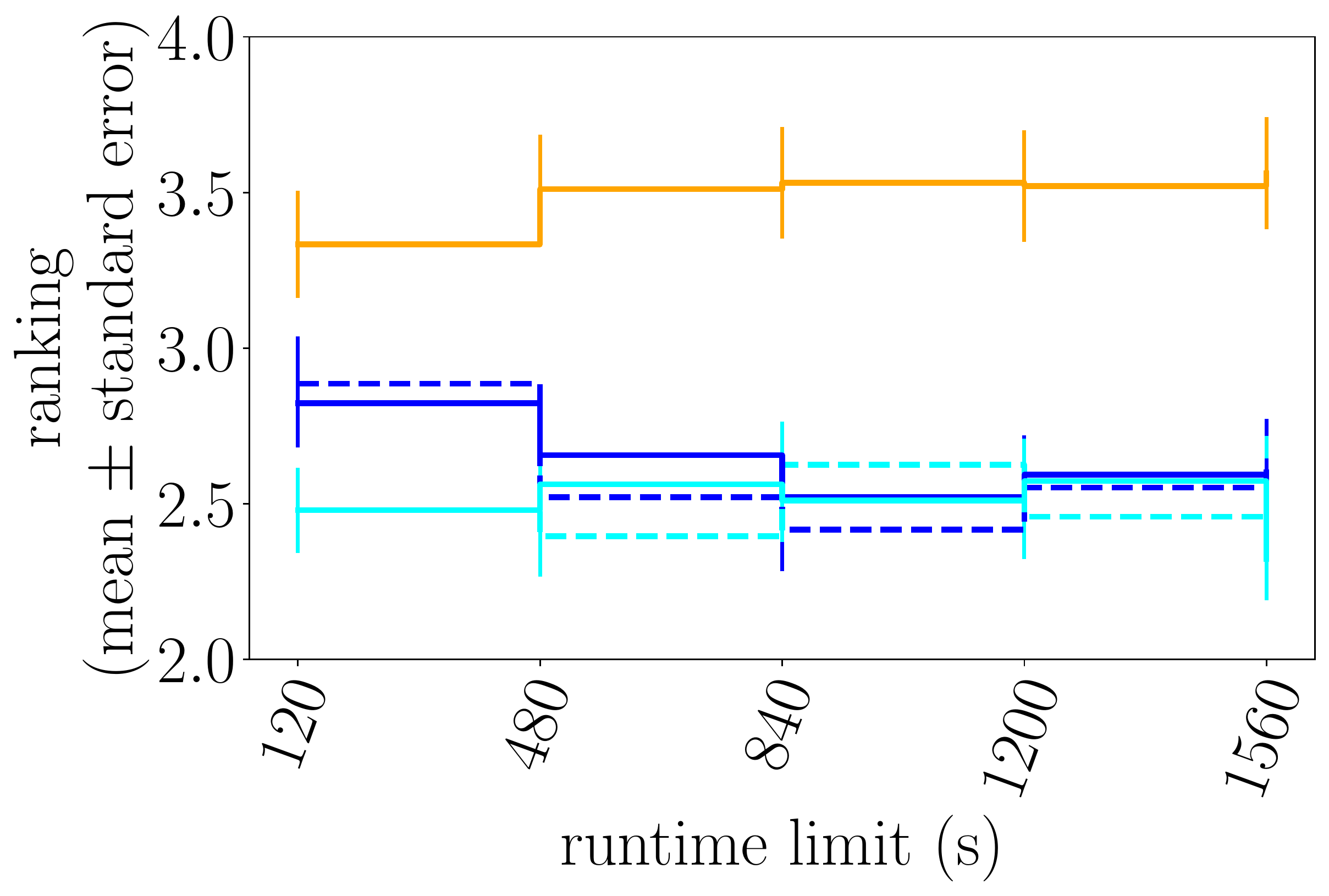}
		\caption{UCI (meta-test)}
		\label{fig:ranking_UCI_time}
	\end{subfigure}%
	
	\begin{subfigure}[t]{.85\linewidth}
		\includegraphics[width=\linewidth]{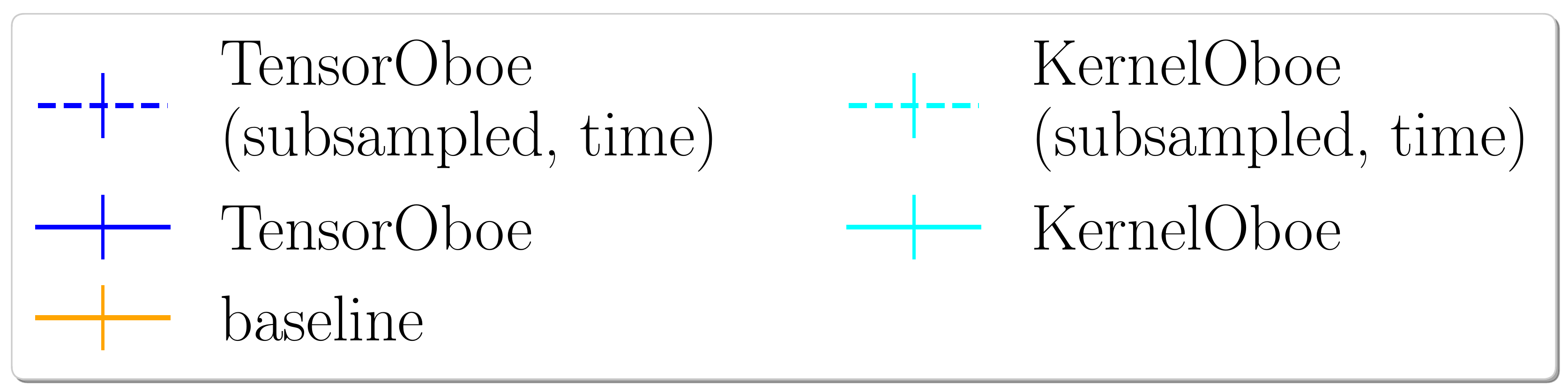}
	\end{subfigure}%
	
	\caption{
		System rankings of \textsc{TensorOboe} and \textsc{KernelOboe} for pipeline search with time-thresholded error tensors, vs the baseline pipeline.
		The error tensor used by the dashed lines marked as ``subsampled, time'' lacks all pipeline performance that take more than 20 seconds to evaluate.
		Dataset and estimator ranks are set to be 25 to complete the error tensor in \textsc{TensorOboe}.}
	\label{fig:comparison_of_frameworks_time}
\end{figure}

In the third experiment, we compare the performance of \textsc{TensorOboe} and \textsc{KernelOboe} with the error tensor that only contain pipeline execution results that take less than 20 seconds.
This error tensor has $9.2\%$ entries missing, compared to the original $3.3\%$ missing ratio with the time threshold of 120 seconds. 
Same as the evaluation above, we compare the performance of \textsc{TensorOboe} and \textsc{KernelOboe} as AutoML systems and show the results in Figure~\ref{fig:comparison_of_frameworks_time}. 
We can see similar trends as above.

\subsection{Comparison of Completion Methods}
\label{sec:completion_experiments}
In this section, we compare the performance of EM algorithm on matrix and tensor completion, and KFMC on our error tensor.
We explore two completion scenarios: entries missing at random, and entries with a longer running time missing.

\subsubsection{Uniformly Sample the Error Tensor}
\label{sec:sample_randomly_experiments}
Given meta-training data $\{\mathcal{D}, \mathcal{P}, \mathcal{P}(\mathcal{D})\}$ on a subset of dataset-pipeline combinations, a good surrogate model should be able to accurately predict the performance of new dataset-pipeline combinations.

We explore the setting commonly seen in matrix completion literature: entries are missing at random with a given missing ratio.
Practitioners may thus subsample pipeline-dataset combinations in the meta-training phase to reduce the number of evaluations.

\begin{figure}
	\centering
	\includegraphics[width=.85\linewidth]{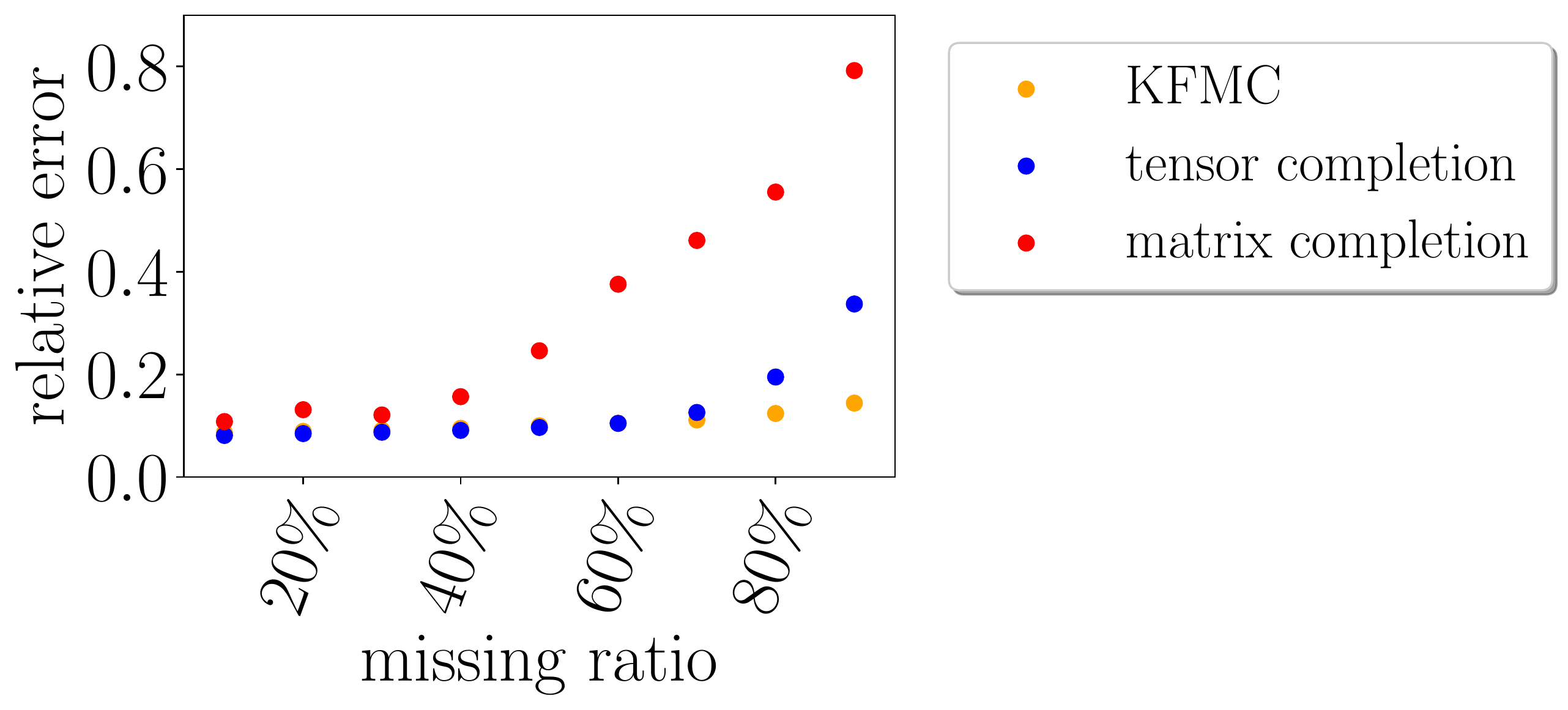}
	\caption{Prediction performance on pipeline-dataset combinations that are missing at random.
		Dataset and estimator ranks for tensor and matrix completions are set to be 40. The $r$ in KFMC is set to be 215.
	}
	\label{fig:missing_at_random}
\end{figure}

Figure~\ref{fig:missing_at_random} shows the relative error on entries that are taken out at random.
To ensure a fair comparison, we set the dataset and estimator ranks to be equal in the tensor model, which is required for the matrix model, since column rank equals row rank for a matrix.
We can see that:
\begin{enumerate}[label=\arabic*, wide, labelwidth=!, labelindent=0pt]
	\item The tensor model outperforms matrix in all cases, demonstrating that the additional combinatorial structure provided by the tensor model helps us recover the combinatorial relationships among different pipeline components.
	\item KFMC achieves lower error on all missing ratio settings and has no drastic increase in relative error in almost the entire range of missing ratios, making it possible for drastic subsampling to save computing power in meta-training.
\end{enumerate}

\subsubsection{Sample the Error Tensor by Running Time}
\label{sec:sample_by_time_experiments}
\begin{figure}
	\centering
	\includegraphics[width=.5\linewidth]{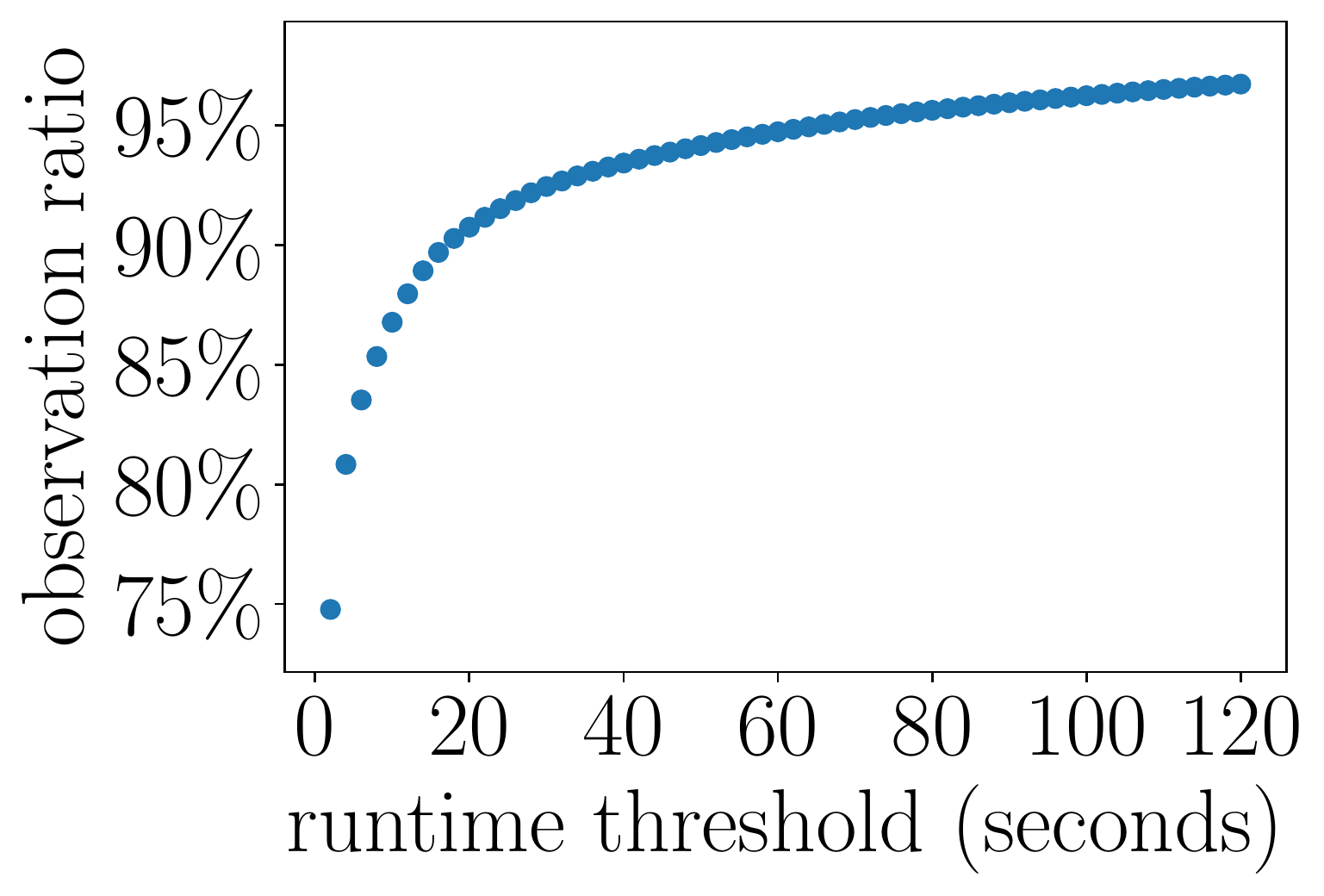}
	\caption{CDF of pipeline runtime on meta-training datasets.}
	\label{fig:missing_ratios}
\end{figure}

In this scenario, we explore the effect of only collecting fastest pipeline-dataset combinations in meta-learning. 
Figure~\ref{fig:missing_ratios} shows that most pipelines run quickly on most datasets: for example, over 90\% finish in less than 20 seconds and over 95\% finish in less than 80 seconds.

\begin{figure}
	\centering
	\begin{subfigure}[t]{.8\linewidth}
		\includegraphics[width=\linewidth]{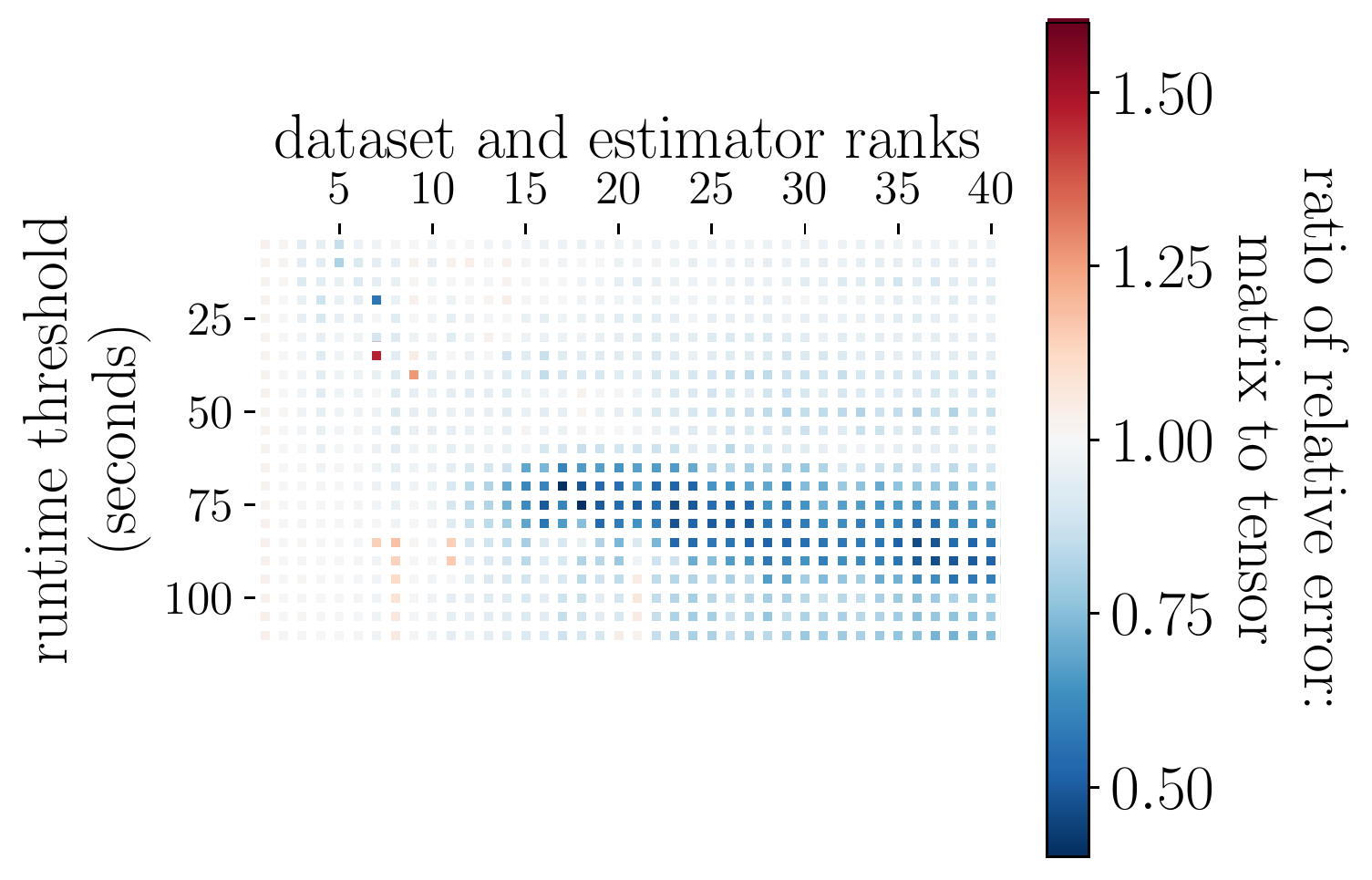}
		\caption{Runtime generalization by tensor vs matrix models.
			Blue denotes that the tensor model is better in achieving a smaller error for pipeline evaluations with longer running time.
			Red denotes the matrix model is better.}
		\label{fig:tensor_vs_matrix_heatmap}
	\end{subfigure}%
	
	\begin{subfigure}[t]{.7\linewidth}
		\includegraphics[width=\linewidth]{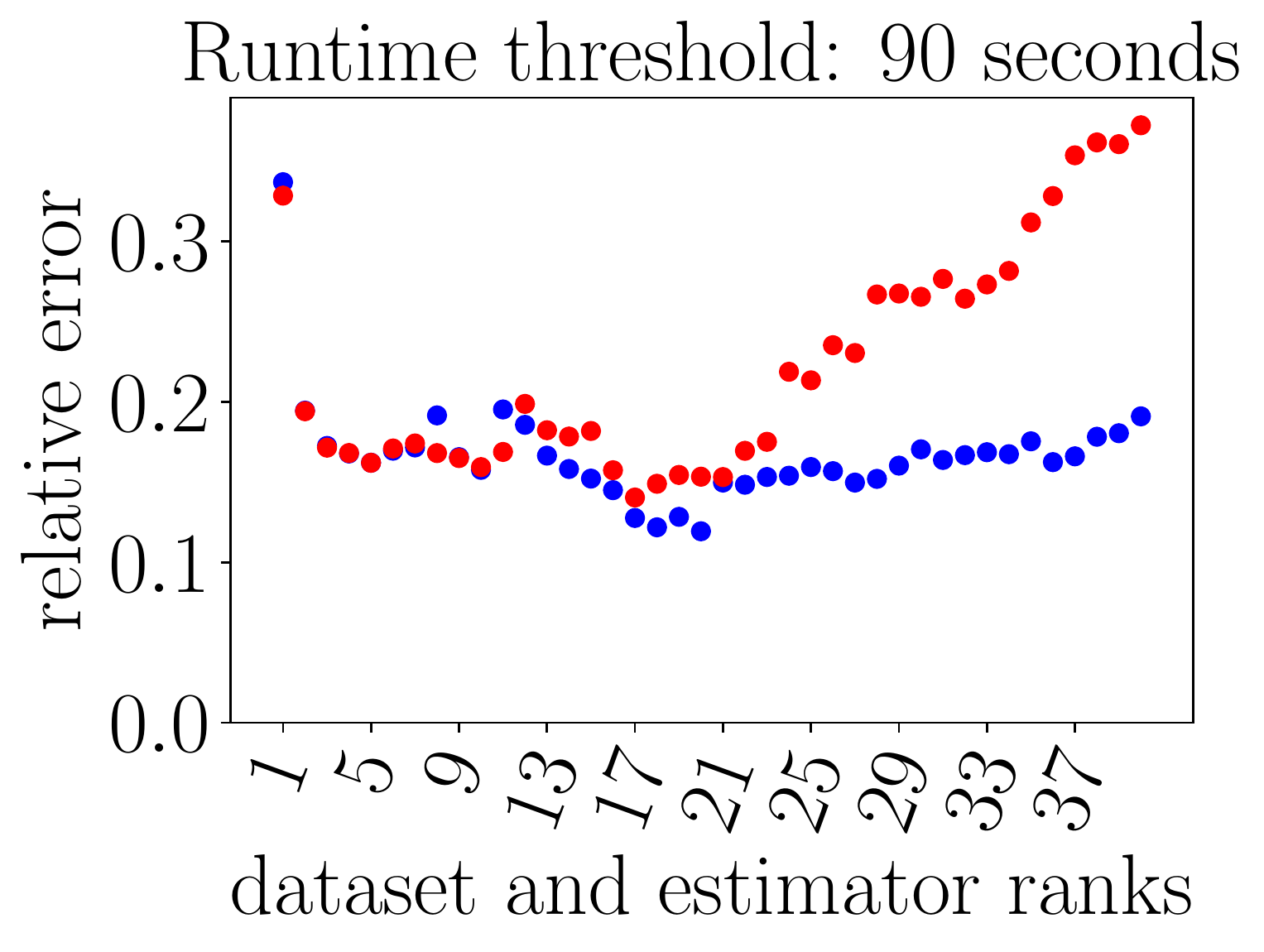}
		\caption{An example of tensor completion and matrix completion errors with runtime threshold to be 90 seconds for pipeline performance collection.
			Error tensor missing ratio 3.3\%.}
		\label{fig:tensor_vs_matrix_example}
	\end{subfigure}%
	
	\begin{subfigure}[t]{.8\linewidth}
		\includegraphics[width=\linewidth]{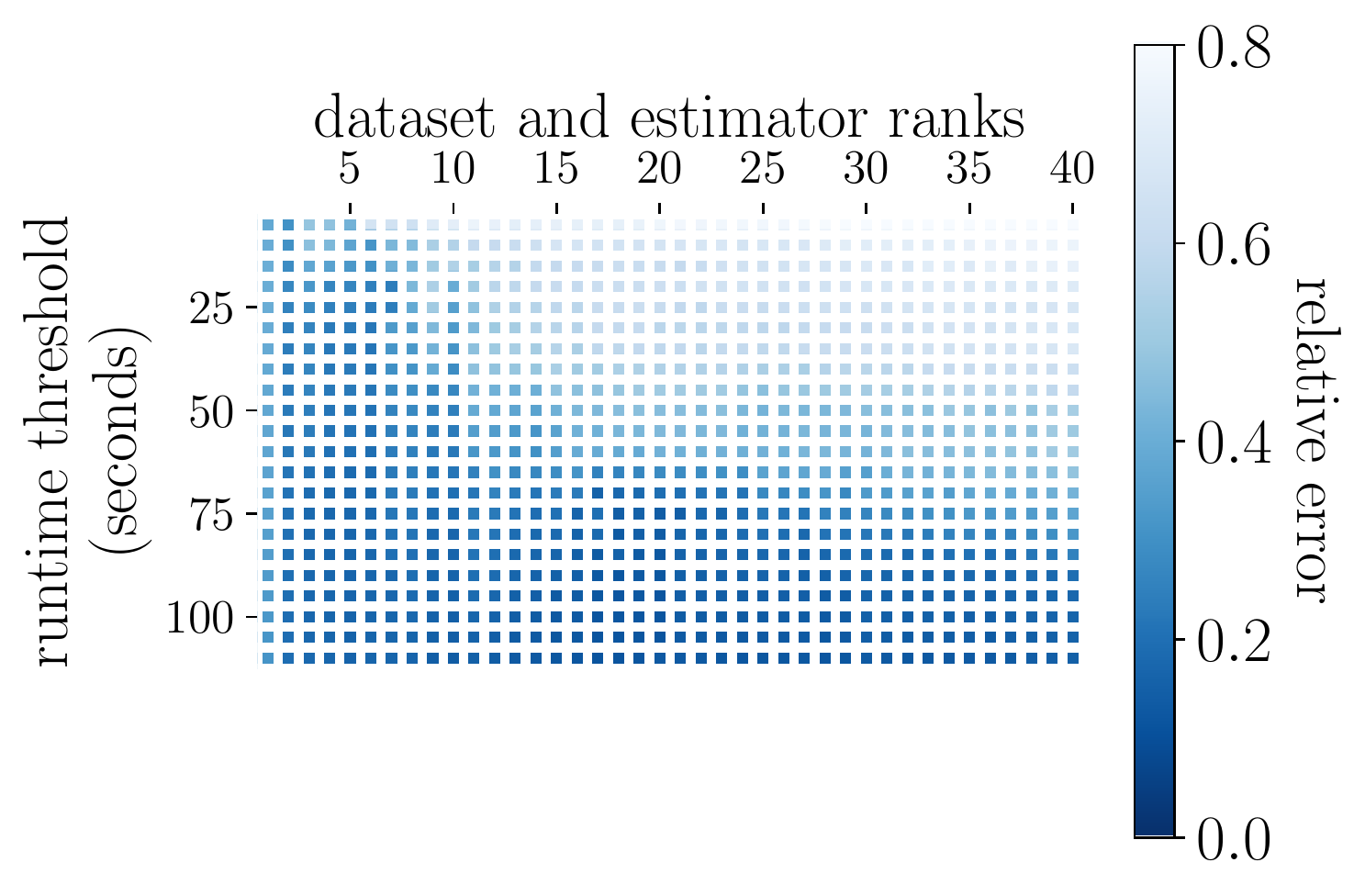}
		\caption{Runtime generalization error by tensor model.
			Darker the color, smaller the error.}
		\label{fig:tensor_heatmap}
	\end{subfigure}%
	
	\begin{subfigure}[t]{.8\linewidth}
		\includegraphics[width=\linewidth]{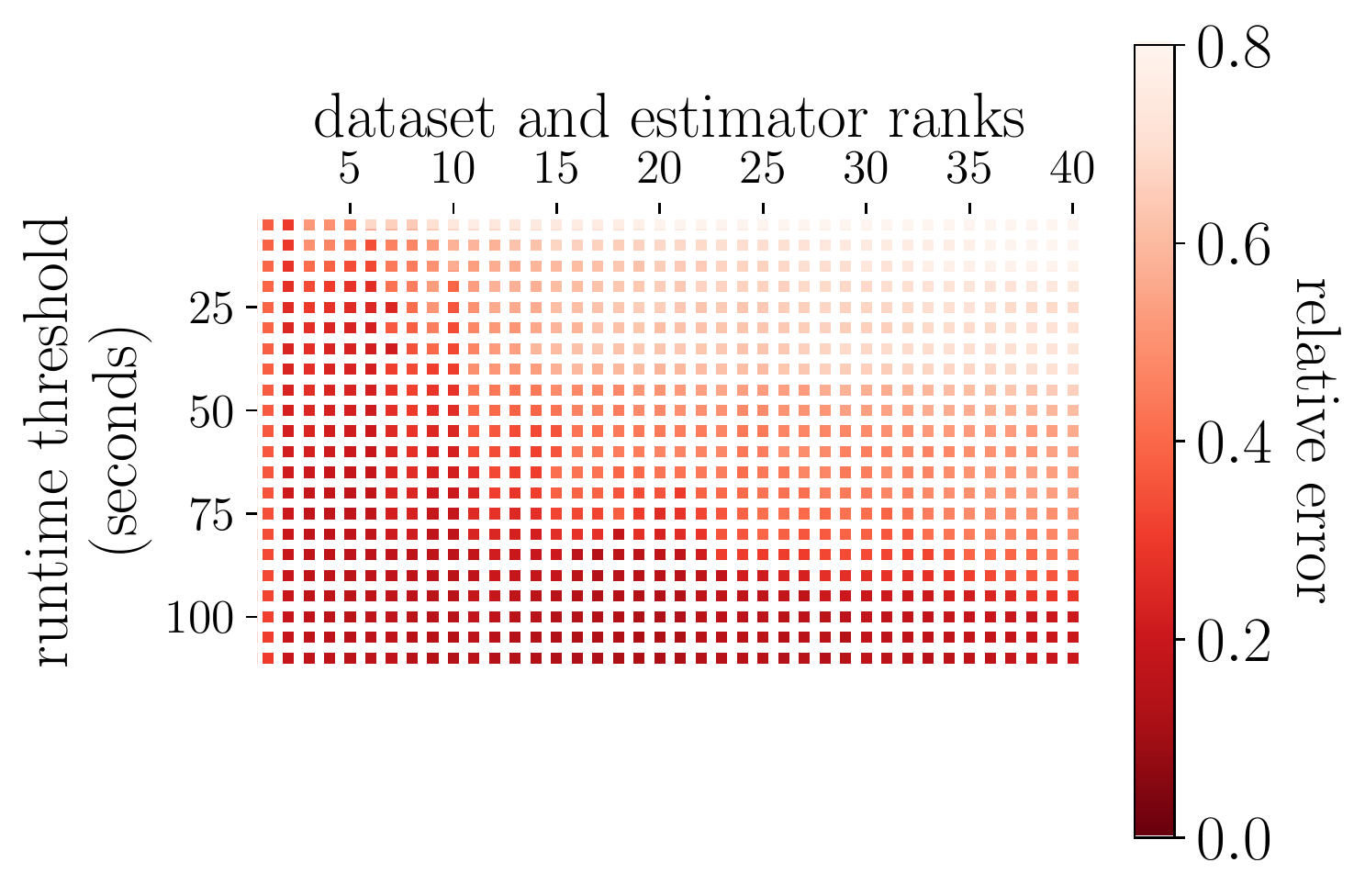}
		\caption{Runtime generalization error by matrix model.
			Darker the color, smaller the error.}
		\label{fig:matrix_heatmap}
	\end{subfigure}%
	
	\vspace{.5em}
	\begin{subfigure}[t]{.7\linewidth}
		\includegraphics[width=\linewidth]{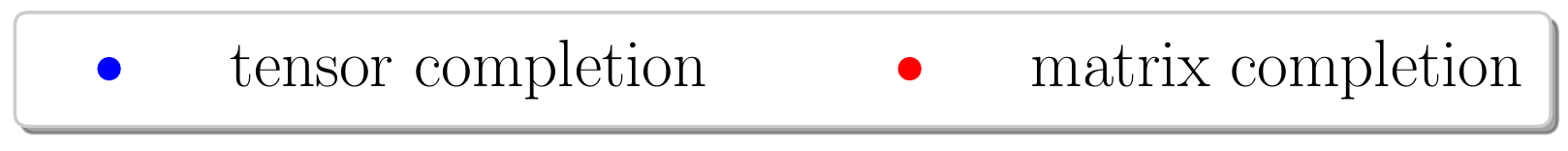}
		\label{fig:legend_tensor_vs_matrix}
	\end{subfigure}%
	\vspace{-2em}
	
	\caption{
		Tensor completion vs matrix completion for inferring pipeline performance.}
	\label{fig:tensor_vs_matrix_all}
\end{figure}

Figure~\ref{fig:tensor_vs_matrix_all} compares relative errors of predictions by tensor and matrix surrogate models.
For each runtime threshold, we treat pipeline-dataset combinations with running time less than the threshold as training data, and those with running time longer than threshold and less than 120 seconds as test data.
We compute relative errors on test data, hence the name ``runtime generalization''.
In addition to the observation in Section~\ref{sec:sample_randomly_experiments} on the advantage of tensor completion versus matrix completion, we can see from Figure~\ref{fig:tensor_vs_matrix_example}, the U-shaped curve of relative error when increasing dataset and estimator ranks for both matrix and tensor model, that the low rank models change from underfitting to overfitting as the ranks increase.
Thus we select both dataset and estimator ranks to be 20, the rank in the middle, in \textsc{TensorOboe}.

\begin{figure}
	\centering
	\includegraphics[width=.85\linewidth]{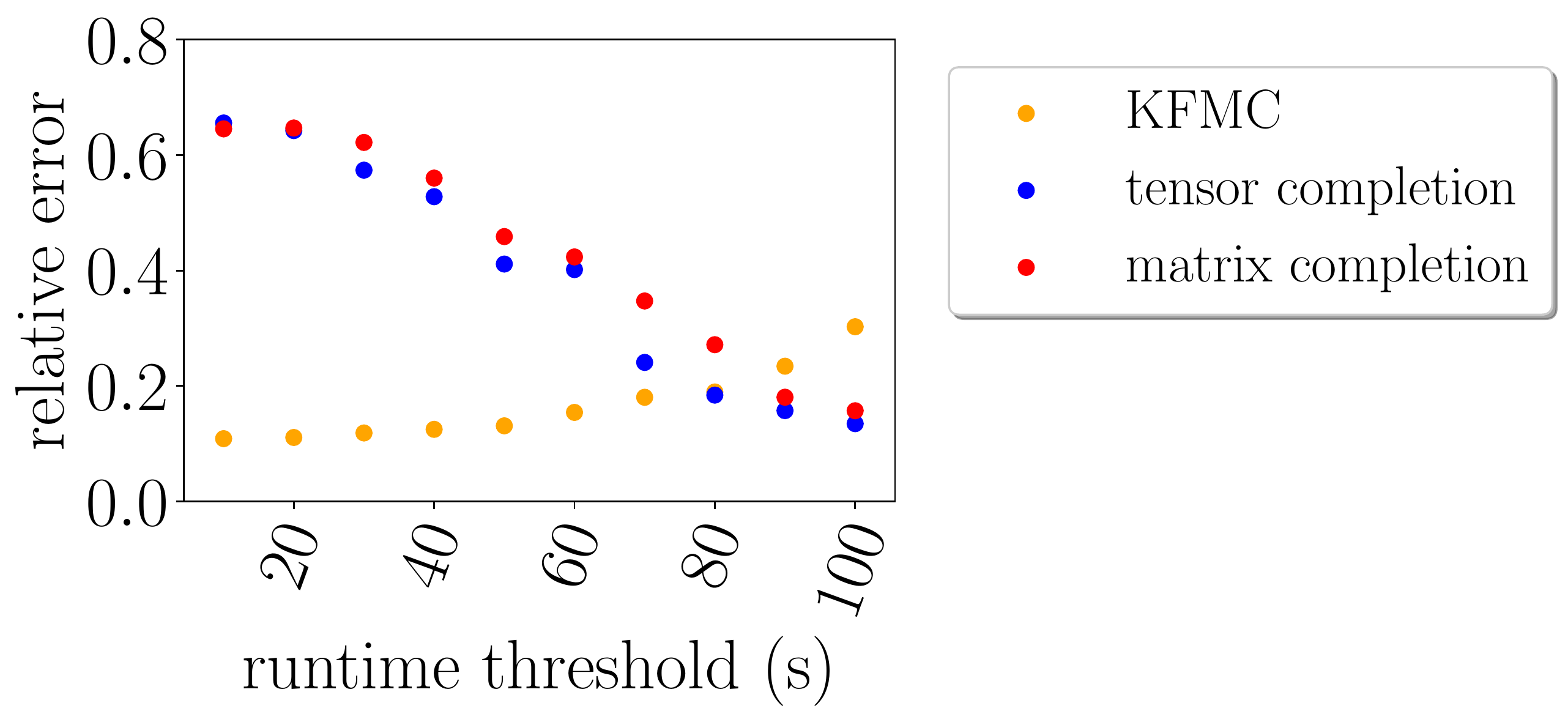}
	\caption{Prediction results on pipeline-dataset combinations that are missing by runtime thresholding: results of combinations that take longer than the threshold are missing.
		Dataset and estimator ranks for tensor and matrix completions are set to be 25. The $r$ in KFMC is set to be 215.
	}
	\label{fig:missing_by_time}
\end{figure}

Figure~\ref{fig:missing_by_time} shows the relative error on entries that are missing by runtime thresholding, which are the entries with runtime between the corresponding threshold and the runtime upper limit in error tensor collect (120 seconds in our experiments).
Similar to Section~\ref{sec:sample_randomly_experiments}, we can see that:
\begin{enumerate}[label=\arabic*, wide, labelwidth=!, labelindent=0pt]
	\item Tensor completion has smaller error than matrix completion in almost all cases.
	\item Relative to the EM approach for tensor and matrix completions, KFMC outperforms at smaller runtime thresholds, which is more interesting in practice. 
\end{enumerate}

\subsection{Pipeline Selection by Greedy Experiment Design}
\label{sec:ED_evaluation}
We compare the performance of different approaches to solve the experiment design problem, so as to choose which pipelines we should sample.

Recall that there are two approaches:
\begin{itemize}
	\item \textbf{Convexification}: Solve the relaxed problem (Equation~\ref{eq:ED_time} with $v_i \in [0, 1]$, $\forall i \in [n]$) with an SLSQP solver, sort the entries in the optimal solution $v^*$, and greedily add the pipeline with large $v_i^*$, until the runtime limit is reached.
	\item \textbf{Greedy}: Solve the original integer programming problem (Equation~\ref{eq:ED_time}) by the greedy algorithm (Algorithm~\ref{alg:ED_time_greedy_without_repetition}), initialized by time-constrained QR (Algorithm~\ref{alg:ED_time_qr_initialization}).
\end{itemize}

\begin{figure}
	\centering
	\begin{subfigure}[t]{.6\linewidth}
		\includegraphics[width=\linewidth]{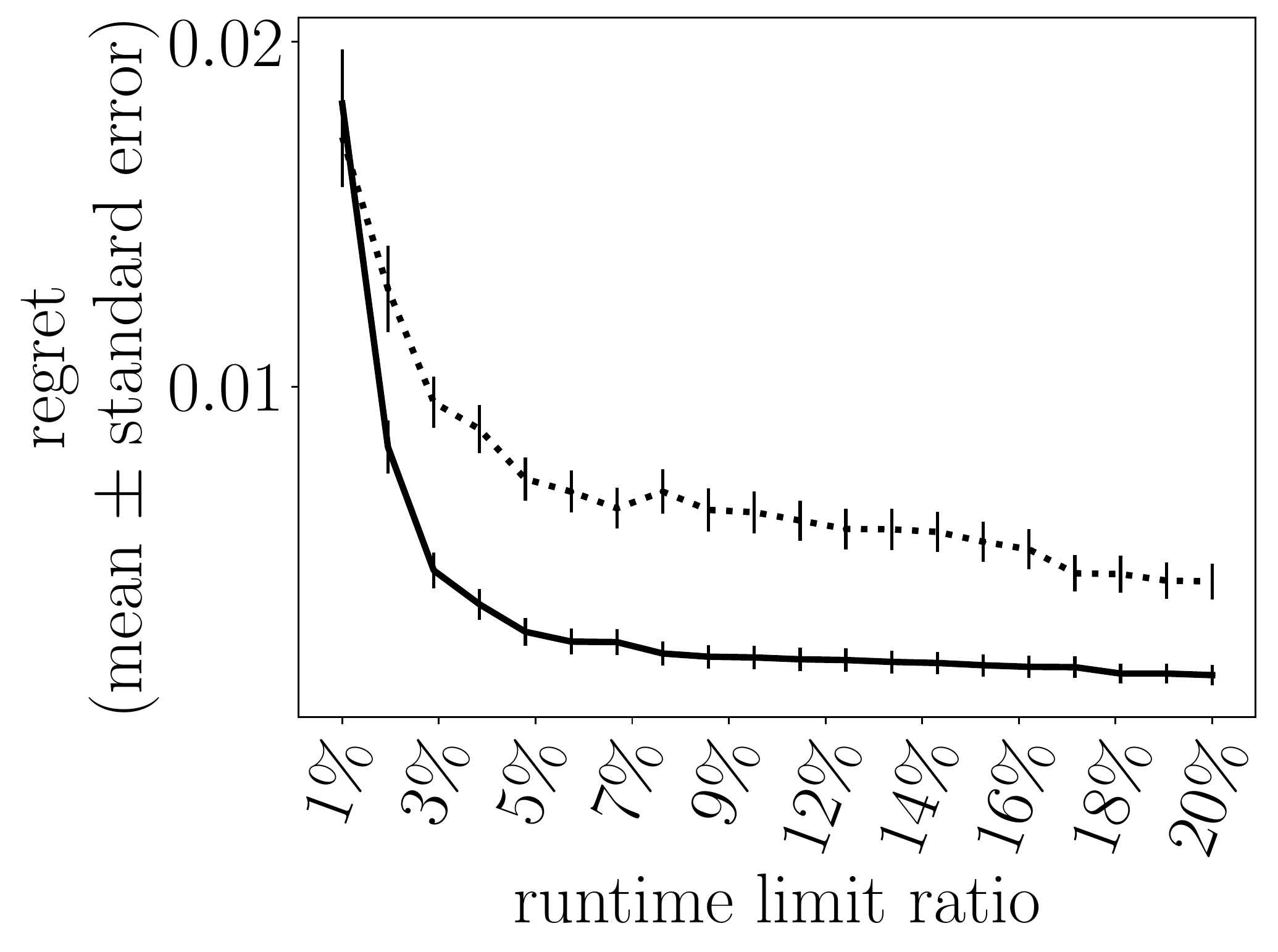}
		\caption{Regret on the subsampled error matrix (215-by-183) for estimator search.}
		\label{fig:ED_error_matrix}
	\end{subfigure}%

	\begin{subfigure}[t]{.8\linewidth}
		\includegraphics[width=\linewidth]{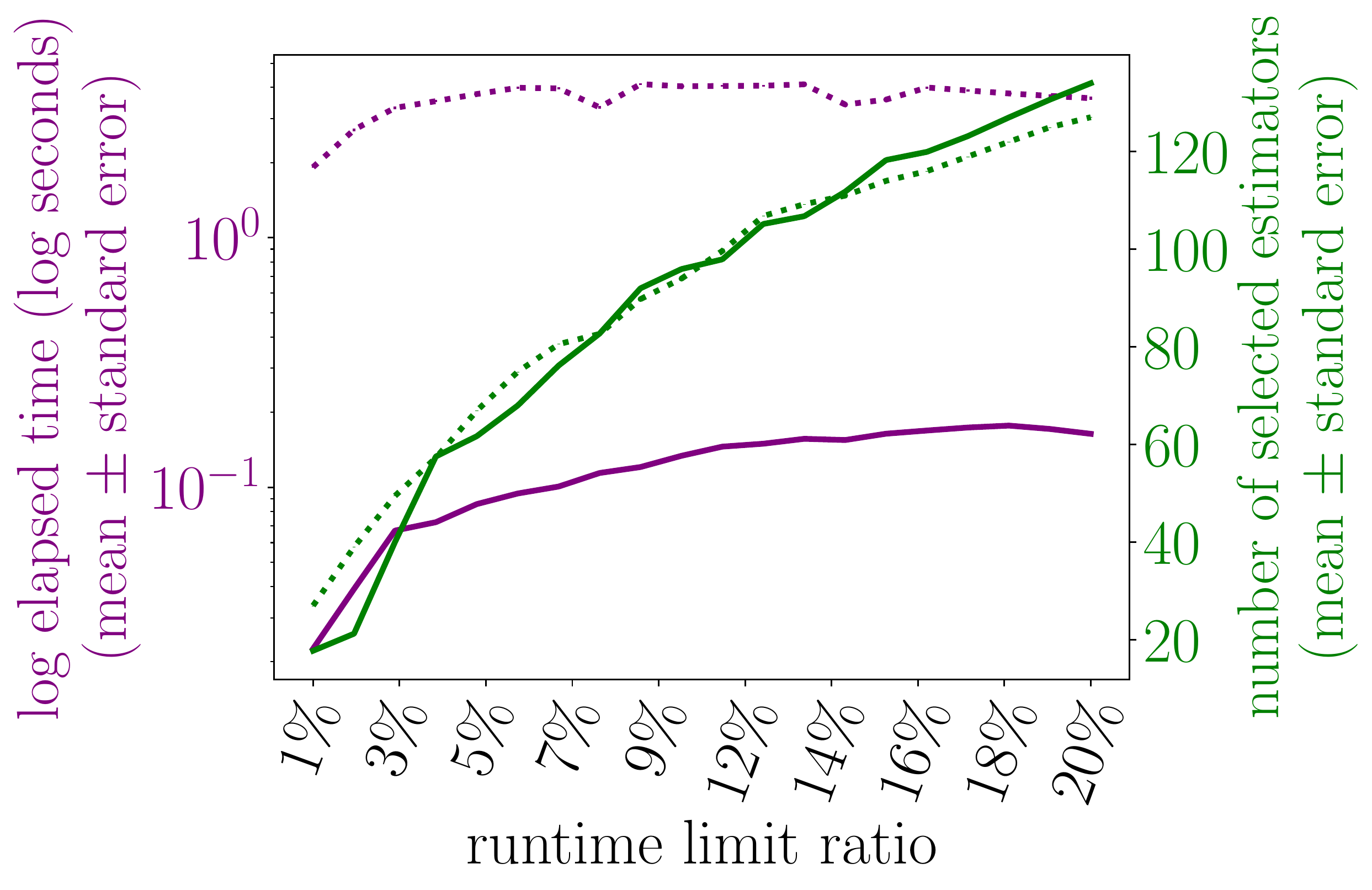}
		\caption{Experiment design running time and number of selected estimators on the subsampled error matrix for estimator search.}
		\label{fig:ED_error_matrix_time}
	\end{subfigure}%
	
	\begin{subfigure}[t]{.6\linewidth}
		\includegraphics[width=\linewidth]{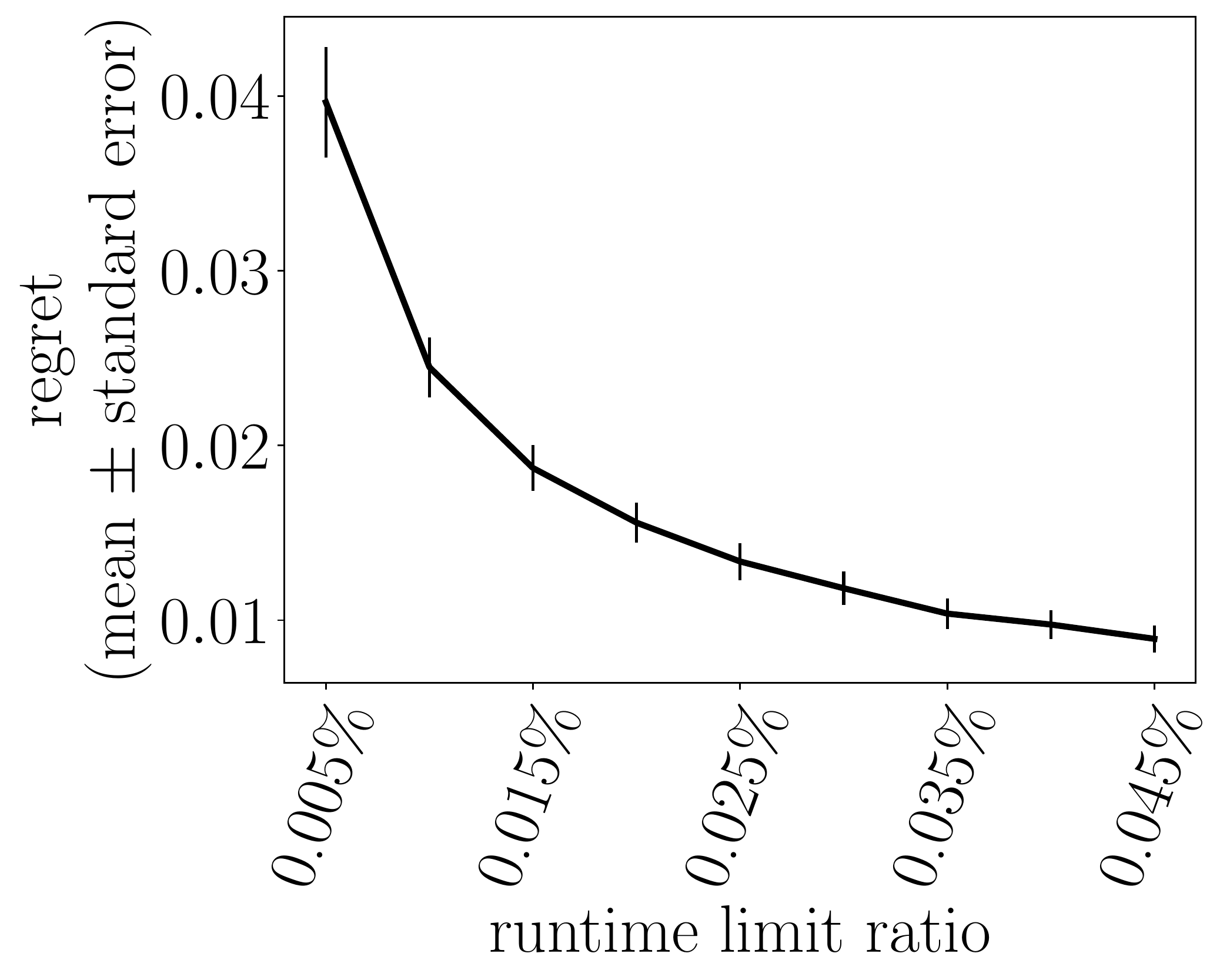}
		\caption{Regret on the full error matrix (215-by-23424) for pipeline search, by the greedy method.}
		\label{fig:ED_error_tensor}
	\end{subfigure}%

	\begin{subfigure}[t]{.8\linewidth}
		\includegraphics[width=\linewidth]{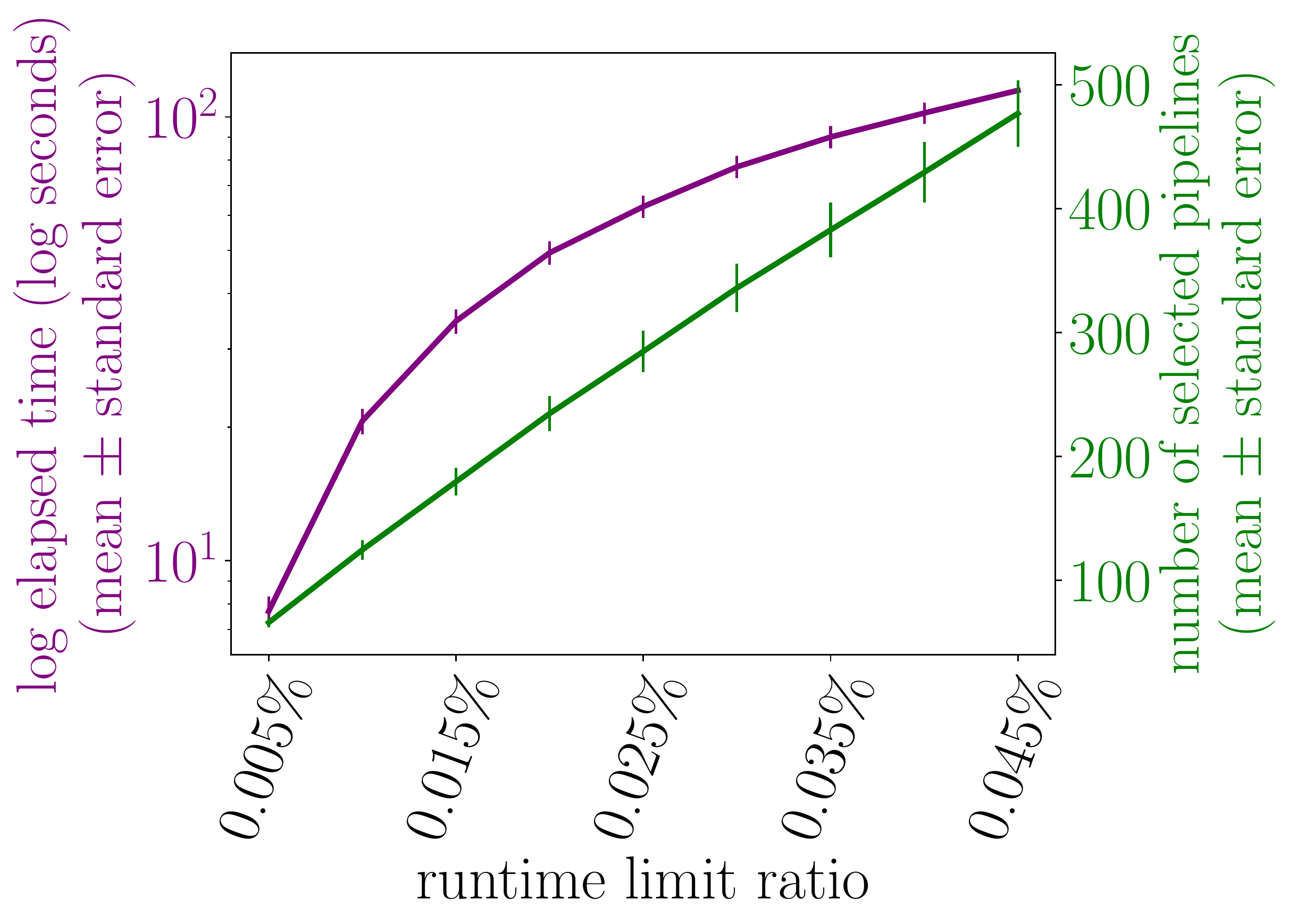}
		\caption{Experiment design running time and number of selected pipelines on the full error matrix for pipeline search, by the greedy method.}
		\label{fig:ED_error_tensor_time}
	\end{subfigure}%
	
	\vspace{.5em}
	\begin{subfigure}[t]{.52\linewidth}
		\includegraphics[width=\linewidth]{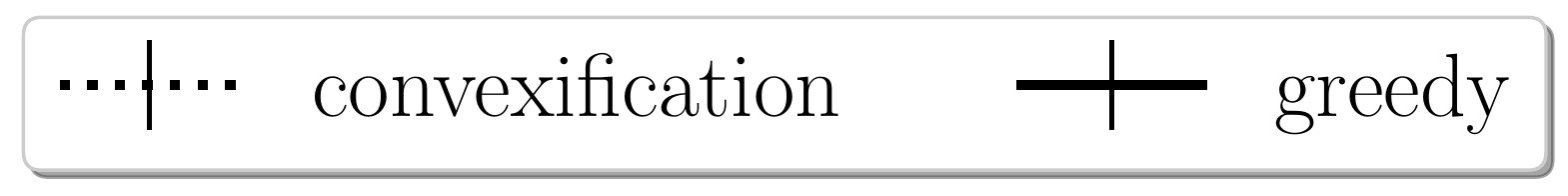}
		\label{fig:legend_ED}
	\end{subfigure}%
	\vspace{-2em}
	\caption{Time-constrained experiment design methods across meta-training datasets.
		The y-axes in \ref{fig:ED_error_matrix} and \ref{fig:ED_error_tensor} are regrets: the difference between minimum pipeline error found by the respective approaches and the actual one.
		The x-axes are runtime limit ratios: ratios of the runtime limit to the total running time of all pipelines on each dataset.
	}
	\label{fig:ED_comparison}
\end{figure}

For our problem, the greedy approach is superior, since the convexification method is prohibitive on our large 215-by-23424 error matrix.
Hence we compare these methods on a subset of pipelines that only differ by estimators, 183 in total.
This is the same as the setting in Oboe \cite{yang2019oboe}.
Shown in Figure~\ref{fig:ED_comparison}, we can see that:
\begin{enumerate}[label=\arabic*, wide, labelwidth=!, labelindent=0pt]
	\item The greedy method performs better for cold-start than convexification (Figure~\ref{fig:ED_error_matrix}): it selects informative designs that better predict the high-percentage pipelines when selecting similar number of designs (Figure~\ref{fig:ED_error_matrix_time}).
	\item The greedy method is more than 30$\times$ faster than convexification, which allows the AutoML system to devote its runtime budget to fitting pipelines instead of searching for the informative pipelines.
	\item Shown in Figure~\ref{fig:ED_error_tensor_time}, the greedy algorithm would still take a fair amount of time if the number of designs we select is large; however, the dataset ranks we choose are less than 50, which means it generally takes less than 10 seconds to choose informative pipelines.
\end{enumerate}

\subsection{Pipeline Runtime Prediction Performance}
\label{sec:runtime_prediction_performance}
The extent of how accurate we can predict running times of pipelines greatly affects performance of our time-constrained pipeline selection system.
Recall that we use order-3 polynomial regression on
$n^\mathcal{D}$ and $p^\mathcal{D}$, the numbers of data points and features in $\mathcal{D}$, and their logarithms.
We shown in Table~\ref{table:runtime_accuracy} that this runtime predictor performs well.
Visualization plots that previously appeared in~\cite{yang2019oboe} is in Appendix~\ref{supp:runtime_prediction_accuracy}, Figure~\ref{fig:runtime_prediction_by_type}. 

\begin{table}
	\captionof{table}{Runtime prediction accuracy on OpenML datasets}
	\label{table:runtime_accuracy}
	\centering
	\begin{tabular}{lrr}
		\hline
		Pipeline estimator type  &  \multicolumn{2}{c}{Runtime prediction accuracy}\\
		& within factor of 2 & within factor of 4\\
		\hline
		Adaboost & 73.6\% & 86.9\%\\
		Decision tree & 62.7\% & 78.9\%\\
		Extra trees	& 71.0\% & 83.8\%\\
		Gradient boosting & 53.4\% & 77.5\%\\
		Gaussian naive Bayes & 67.3\% & 82.3\%\\
		kNN	& 68.7\% & 84.4\%\\
		Logistic regression	& 53.6\% & 76.1\%\\
		Multilayer perceptron & 74.5\% & 88.9\%\\
		Perceptron & 64.5\% & 82.2\%\\
		Random Forest & 69.5\% & 84.9\%\\
		Linear SVM & 56.8\% & 79.5\%\\
		
		\hline
	\end{tabular}
	
\end{table}

\subsection{Learning the Hyperparameter Landscapes}
\label{sec:hyperparameter_landscape}
Hyperparameter landscapes plot pipeline performance with respect to hyperparameter values.
While parameter landscapes have been extensively studied, especially in the deep learning context (for example, \cite{keskar2017large, li2018visualizing, garipov2018loss}), hyperparameter landscapes are less studied.
The previous sections focus on how we can choose among different pipeline component types.
In this section, we show that our tensor surrogate model is able to learn hyperparameter landscapes of different estimator types that exhibit qualitatively different behaviors.

\begin{figure}
	\centering
	\begin{subfigure}[t]{.45\linewidth}
		\includegraphics[width=\linewidth]{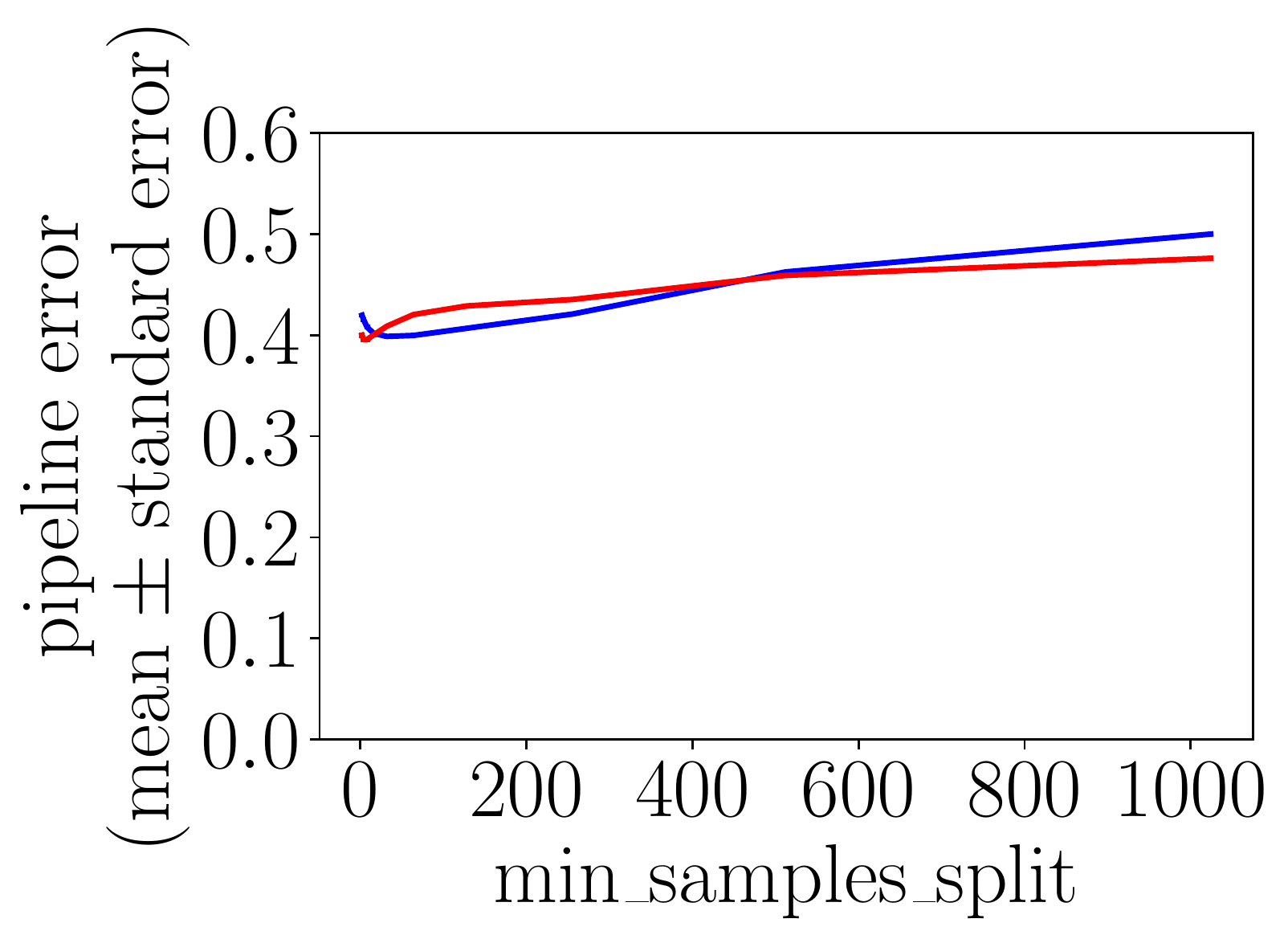}
		\caption{Extra trees on OpenML Dataset 23 (1473 points, 10 features)}
		\label{fig:landscape_1}
	\end{subfigure}%
	\hspace{.09\linewidth}
	\begin{subfigure}[t]{.45\linewidth}
		\includegraphics[width=\linewidth]{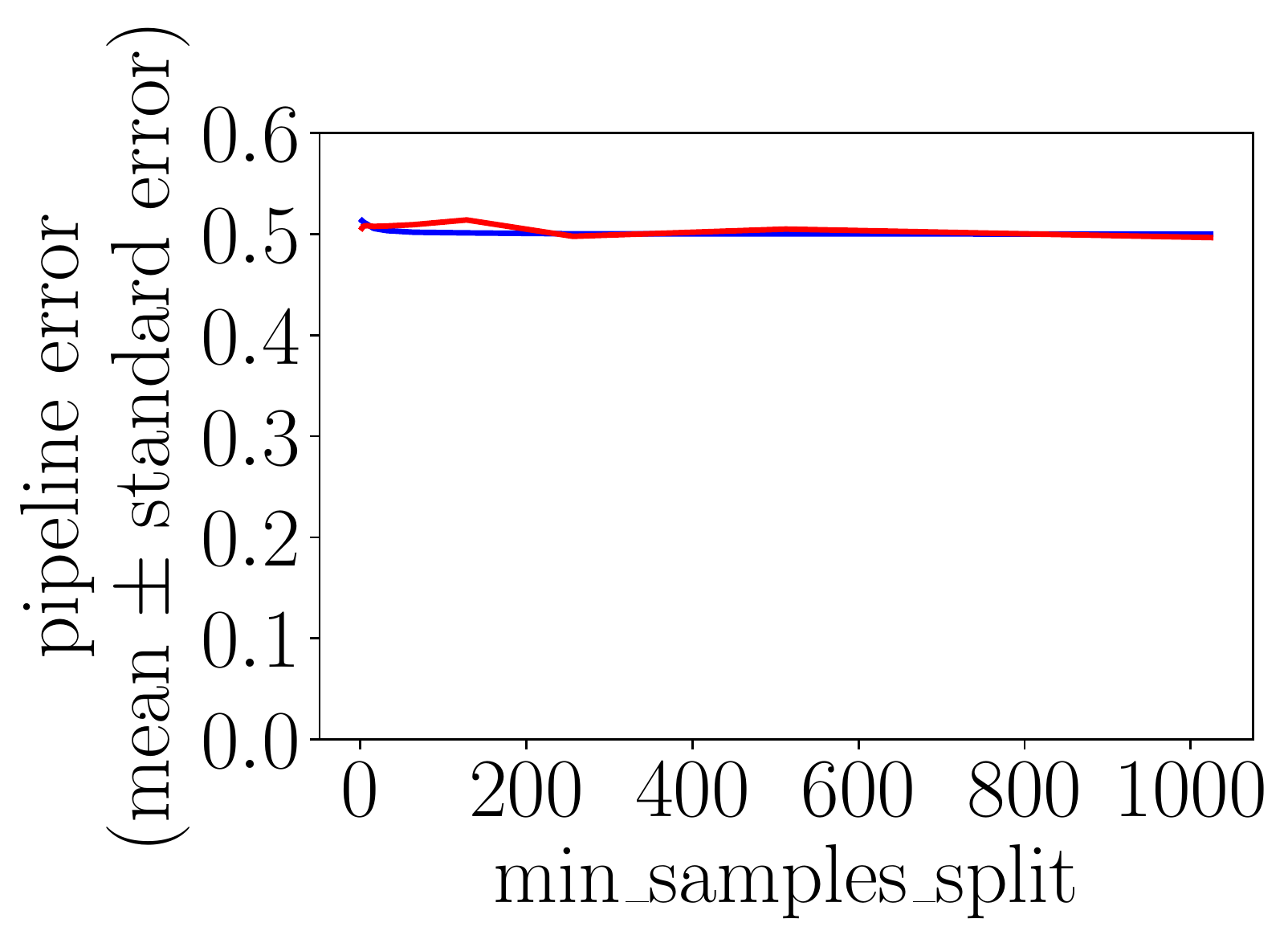}
		\caption{Decision tree on OpenML Dataset 1014 (797 points, 5 features)}
		\label{fig:landscape_2}
	\end{subfigure}%
	
	\begin{subfigure}[t]{.45\linewidth}
		\includegraphics[width=\linewidth]{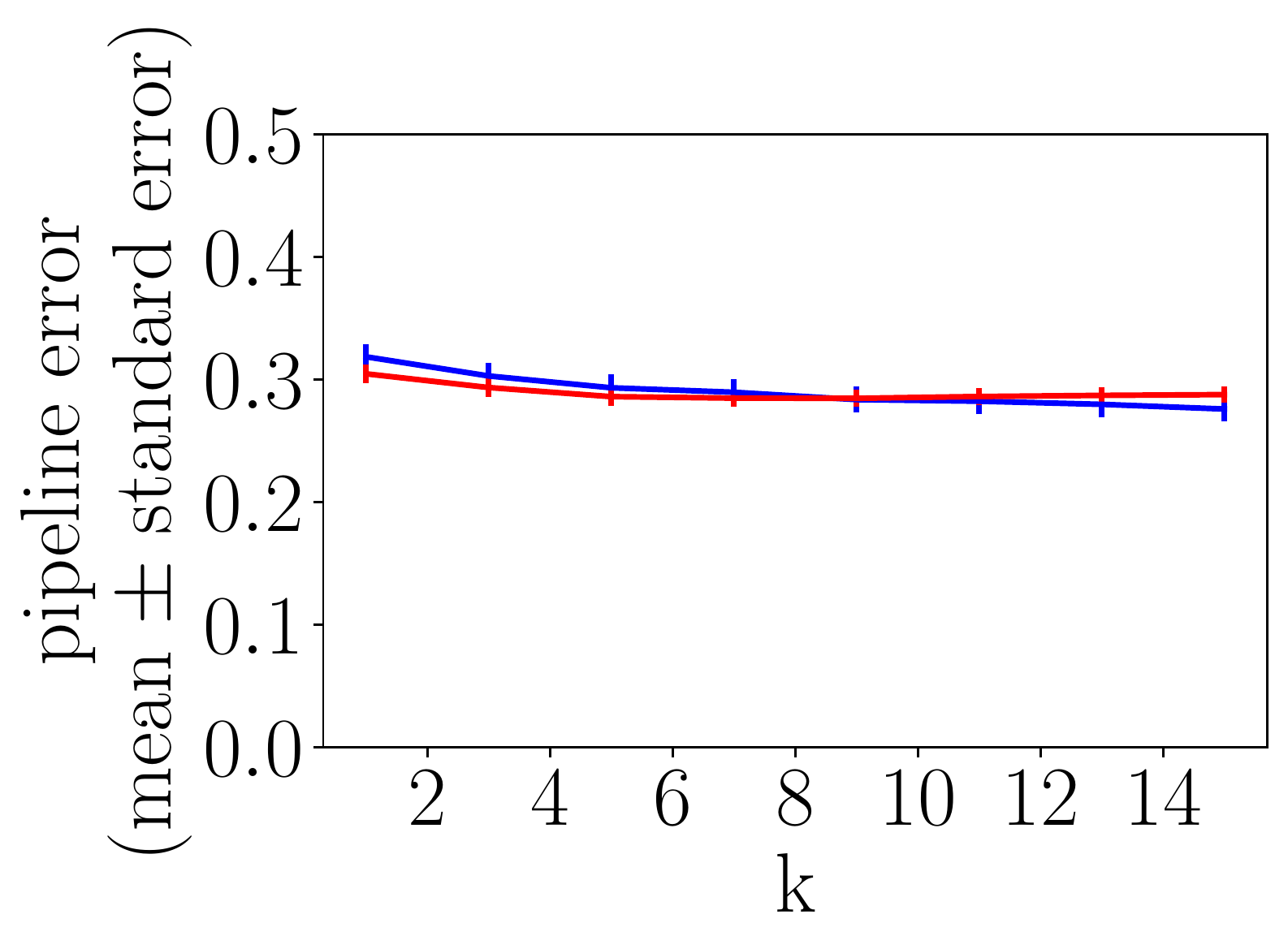}
		\caption{kNN on OpenML Dataset 799 (1000 points, 6 features)}
		\label{fig:landscape_3}
	\end{subfigure}%
	\hspace{.09\linewidth}
	\begin{subfigure}[t]{.45\linewidth}
		\includegraphics[width=\linewidth]{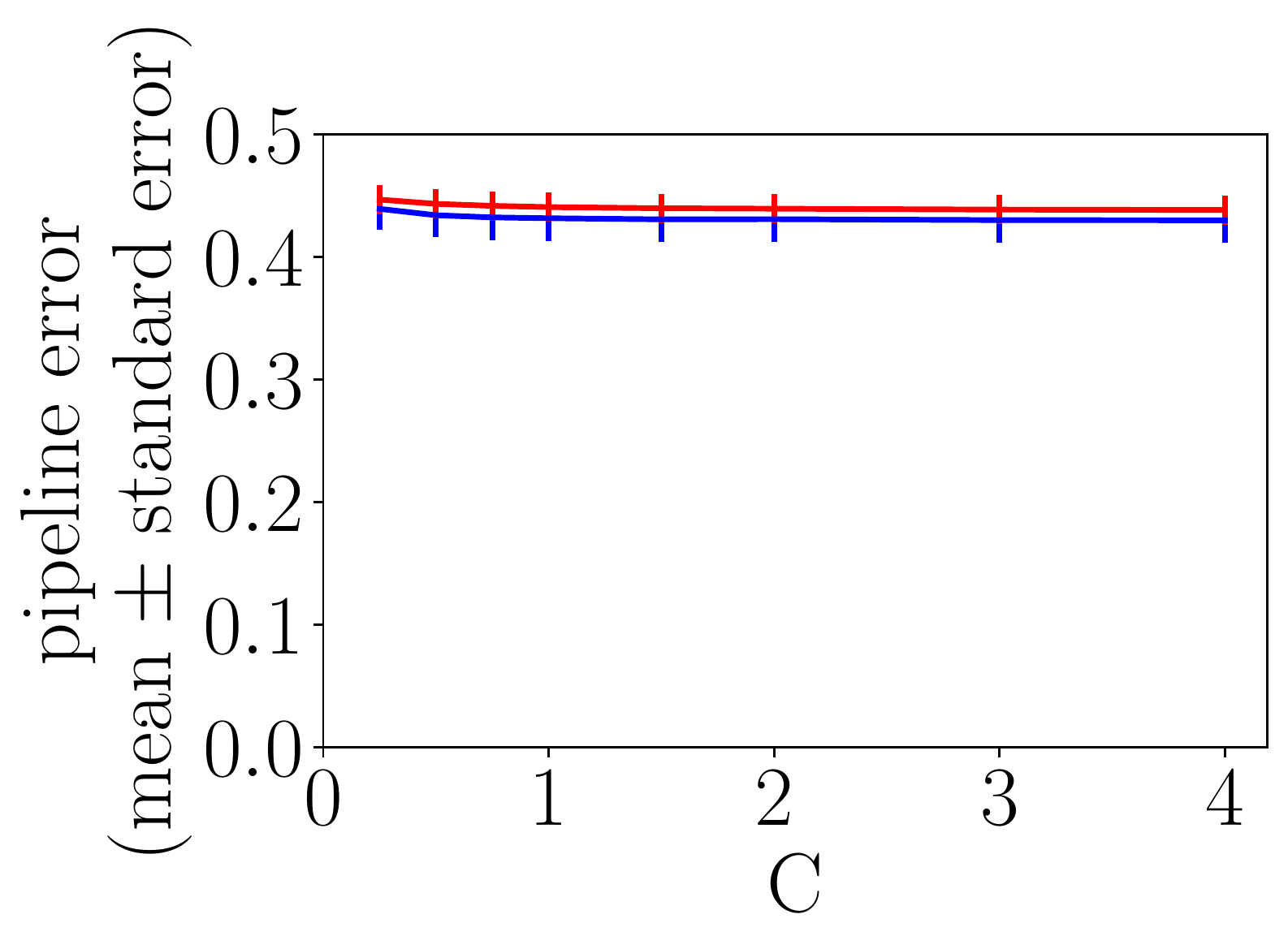}
		\caption{Logistic regression on OpenML Dataset 40971 (1000 data points, 24 features)}
		\label{fig:landscape_4}
	\end{subfigure}%
	
	\vspace{.5em}
	\begin{subfigure}[t]{.52\linewidth}
		\includegraphics[width=\linewidth]{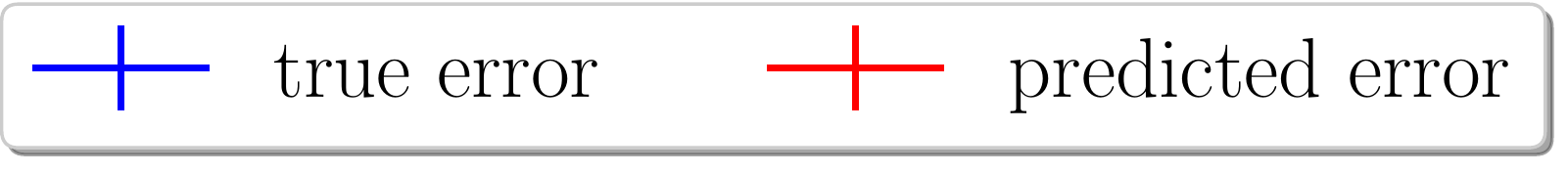}
		\label{fig:legend_hyperparameter_landscape}
	\end{subfigure}%
	\vspace{-2em}
	
	\caption{Hyperparameter landscape prediction examples.}
	\label{fig:landscapes}
\end{figure}

Figure~\ref{fig:landscapes} shows some examples of both real and predicted hyperparameter landscapes after running our system for 135 seconds.
We can see that our predictions match the overall tendencies of the curves.
Their zoomed-in version, shown as Appendix~\ref{supp:zoomed_in_landscapes}, Figure~\ref{fig:zoomed_in_landscapes}, show that our predictions can also capture most of the small variations in these landscapes.

Note our methods do not include a subroutine for hyperparameter optimization: it sets a grid of hyperparameter values for each estimator, instead of optimizing hyperparameters by, for example, Bayesian optimization.
We show by some hyperparameter landscapes that our grid search effectively sample performant hyperparameter settings within the range of hyperparameters, thus a coarse grid suffices.

\section{Conclusion}
This papers develops structured models for AutoML pipeline selection.
The low rank matrix, low Tucker rank tensor and kernelized matrix surrogate model allows us to efficiently learn about new datasets.
Also, we design greedy experiment design methods to select informative pipelines to evaluate.
Empirically, this method improves on the state of the art in AutoML pipeline selection.

There are many avenues for improvement and extensions.
For example, one could enlarge the pipeline search space, 
explore different mechanisms to initialize the greedy method, develop an extension for neural architecture search,
and design task-oriented pipeline selection systems that have better performance on domain-specific datasets.

\ifCLASSOPTIONcompsoc
  \section*{Acknowledgments}
\else
  \section*{Acknowledgment}
\fi

This work was supported in part by DARPA Award FA8750-17-2-0101. 
The authors thank Christophe Giraud-Carrier, Ameet Talwalkar, Iddo Drori, Thorsten Joachims, Raul Astudillo Marban, Matthew Zalesak, Lijun Ding and Davis Wertheimer for helpful discussions, and thank Jack Dunn for a script to parse UCI Machine Learning Repository datasets.

\ifCLASSOPTIONcaptionsoff
  \newpage
\fi

\bibliographystyle{IEEEtran}
\bibliography{cite}

\begin{IEEEbiographynophoto}{Chengrun Yang}
received his BS degree in Physics from Fudan University, Shanghai, China in 2016.
Currently, he is a PhD student at the School of Electrical and Computer Engineering, Cornell University. 
His research interests include the application of low dimensional structures and active learning in resource-constrained learning problems. 
\end{IEEEbiographynophoto}

\begin{IEEEbiographynophoto}{Jicong Fan}
received his BE and ME degrees in Automation and Control Science $\&$
Engineering, from Beijing University of Chemical Technology, Beijing, China, in 2010 and 2013, respectively. 
From 2013 to 2015, he was a research assistant at the University of Hong Kong. 
He received his PhD degree in Electronic Engineering, from City University of Hong Kong, Hong Kong S.A.R. in 2018. 
From 2018.01 to 2018.06, he was a visiting scholar at the
Department of Electrical and Computer Engineering, University of Wisconsin-Madison, USA. 
Currently, he is a postdoc associate at the School of Operations Research and Information Engineering, Cornell University, Ithaca, USA. 
His research interests include computer vision, machine learning, and optimization.
\end{IEEEbiographynophoto}

\begin{IEEEbiographynophoto}{Ziyang Wu}
received the BS degree from Cornell University in 2019. 
He is currently a Master’s student in Computer Science at Cornell University. 
His research interests include computer vision and meta learning, especially in automated machine learning.
\end{IEEEbiographynophoto}

\begin{IEEEbiographynophoto}{Madeleine Udell}
is Assistant Professor of Operations Research and Information Engineering
and Richard and Sybil Smith Sesquicentennial Fellow at Cornell University.
She studies optimization and machine learning for large scale data analysis and control,
with applications in marketing, demographic modeling, medical informatics, engineering system design,
and automated machine learning.
Her work has been recognized by an NSF CAREER award, an Office of Naval Research (ONR) Young Investigator Award,
and an INFORMS Optimization Society Best Student Paper Award (as advisor).
Madeleine completed her PhD at Stanford University in
Computational $\&$ Mathematical Engineering in 2015
under the supervision of Stephen Boyd,
and a one year postdoctoral fellowship at Caltech
in the Center for the Mathematics of Information
hosted by Professor Joel Tropp.
\end{IEEEbiographynophoto}

\appendices
\section{Dataset and Pipeline Configurations}
\label{supp:configurations}
\subsection{Meta-training OpenML Datasets}
\label{supp:dataset_indices}
\begin{footnotesize}
	Indices of the OpenML datasets we use for meta-training:
	2, 3, 5, 7, 9, 11, 12, 13, 14, 15, 16, 18, 20, 22, 23, 24, 25, 27, 28, 29, 30, 31, 35, 36, 37, 38, 39, 40, 41, 42, 44, 46, 48, 50, 53, 54, 59, 60, 181, 182, 183, 187, 285, 307, 313, 316, 329, 336, 337, 338, 375, 377, 389, 446, 450, 458, 463, 469, 475, 694, 715, 717, 718, 720, 721, 723, 725, 728, 730, 732, 733, 735, 737, 740, 742, 743, 744, 745, 746, 747, 748, 749, 750, 751, 753, 763, 769, 773, 776, 778, 779, 788, 792, 794, 796, 797, 799, 803, 805, 806, 807, 813, 818, 819, 820, 824, 825, 826, 830, 832, 837, 838, 847, 853, 855, 863, 866, 869, 870, 871, 873, 877, 880, 884, 888, 896, 900, 903, 904, 906, 907, 908, 909, 910, 911, 912, 913, 915, 917, 920, 923, 925, 926, 933, 934, 935, 936, 937, 941, 943, 952, 953, 954, 955, 958, 962, 970, 971, 973, 976, 978, 979, 980, 983, 987, 991, 994, 995, 996, 997, 1005, 1011, 1012, 1014, 1016, 1020, 1021, 1022, 1025, 1026, 1038, 1039, 1041, 1042, 1048, 1049, 1050, 1054, 1056, 1063, 1065, 1067, 1068, 1069, 1071, 1073, 1100, 1115, 1116, 1121, 4134, 40966, 40971, 40975, 40978, 40979, 40981, 40982, 40983, 40984, 40994, 40997, 41000, 41004, 41005.
\end{footnotesize}

\begin{table*}
	\caption{Pipeline search space}
	\label{table:pipeline_components}
	\centering
	\begin{tabular}{lll}
		\hline
		\textbf{Pipeline component} & \textbf{Algorithm type} &\textbf{Hyperparameter names (values)}\\
		\hline
		Missing value imputer & Simple imputer & \texttt{strategy (mean, median, most\_frequent, constant)}\\
		\hline
		\multirow{2}{*}{Encoder} &
		null &- \\
		& OneHotEncoder &   \texttt{handle\_unknown (ignore), sparse (0)} \\
		\hline
		\multirow{2}{*}{Standardizer} & 
		null &- \\
		&StandardScaler &- \\
		\hline
		\multirow{4}{*}{Dimensionality reducer} 
		& null &- \\
		& PCA &\texttt{n\_components (25\%, 50\%, 75\%)} \\
		& VarianceThreshold &- \\
		& SelectKBest & \texttt{k (25\%, 50\%, 75\%)}\\
		\hline
		\multirow{11}{*}{Estimator} &
		Adaboost & \texttt{n\_estimators (50,100), learning\_rate (1.0,1.5,2.0,2.5,3)} \\
		&	Decision tree & \texttt{min\_samples\_split (2,4,8,16,32,64,128,256,512,1024,0.01,0.001,0.0001,1e-05)} \\
		&	Extra trees & \multicolumn{1}{p{8cm}}{\texttt{min\_samples\_split (2,4,8,16,32,64,128,256,512,1024,0.01,0.001,0.0001,1e-05), criterion (gini,entropy)}} \\
		&	Gradient boosting & \multicolumn{1}{p{11cm}}{\texttt{learning\_rate (0.001,0.01,0.025,0.05,0.1,0.25,0.5), max\_depth (3, 6), max\_features (null,log2)}} \\
		&	Gaussian naive Bayes & - \\
		&	kNN & \texttt{n\_neighbors (1,3,5,7,9,11,13,15), p (1,2)} \\
		&	Logistic regression & \texttt{C (0.25,0.5,0.75,1,1.5,2,3,4), solver (liblinear,saga), penalty (l1,l2)} \\
		&	Multilayer perceptron & \multicolumn{1}{p{11cm}}{\texttt{learning\_rate\_init (0.0001,0.001,0.01),  learning\_rate (adaptive), solver (sgd,adam), alpha (0.0001, 0.01)}}\\
		&	Perceptron & - \\
		&	Random forest & \multicolumn{1}{p{8cm}}{\texttt{min\_samples\_split (2,4,8,16,32,64,128,256,512,1024,0.01,0.001,0.0001,1e-05), criterion (gini,entropy)}} \\
		&	Linear SVM & \texttt{C (0.125,0.25,0.5,0.75,1,2,4,8,16)} \\
		\hline
	\end{tabular}
\end{table*}

\subsection{Meta-test UCI Datasets}
{\footnotesize "acute-inflammations-1", "acute-inflammations-2", "arrhythmia", "balance-scale", "balloons-a", "balloons-b", "balloons-c", "balloons-d", "banknote-authentication", "blood-transfusion-service-center", "breast-cancer-wisconsin-diagnostic", "breast-cancer-wisconsin-original", "breast-cancer-wisconsin-prognostic", "breast-cancer", "car-evaluation", "chess-king-rook-vs-king-pawn", "chess-king-rook-vs-king", "climate-model-simulation-crashes", "cnae-9", "congressional-voting-records", "connectionist-bench-sonar", "connectionist-bench", "contraceptive-method-choice", "credit-approval", "cylinder-bands", "dermatology", "echocardiogram", "ecoli", "fertility", "flags", "glass-identification", "haberman-survival", "hayes-roth", "heart-disease-cleveland", "heart-disease-hungarian", "heart-disease-switzerland", "heart-disease-va", "hepatitis", "hill-valley-noise", "hill-valley", "horse-colic", "image-segmentation", "indian-liver-patient", "ionosphere", "iris", "lenses", "letter-recognition", "libras-movement", "lung-cancer", "magic-gamma-telescope", "mammographic-mass", "monks-problems-1", "monks-problems-2", "monks-problems-3", "mushroom", "nursery", "optical-recognition-handwritten-digits", "ozone-level-detection-eight", "ozone-level-detection-one", "parkinsons", "pen-based-recognition-handwritten-digits", "planning-relax", "poker-hand", "post-operative-patient", "qsar-biodegradation", "seeds", "seismic-bumps", "shuttle-landing-control", "skin-segmentation", "soybean-large", "soybean-small", "spambase", "spect-heart", "spectf-heart", "statlog-project-german-credit", "statlog-project-landsat-satellite", "teaching-assistant-evaluation", "thoracic-surgery", "thyroid-disease-allbp", "thyroid-disease-allhyper", "thyroid-disease-allhypo", "thyroid-disease-allrep", "thyroid-disease-ann-thyroid", "thyroid-disease-dis", "thyroid-disease-new-thyroid", "thyroid-disease-sick-euthyroid", "thyroid-disease-sick", "thyroid-disease-thyroid-0387", "tic-tac-toe-endgame", "trains", "wall-following-robot-navigation-2", "wall-following-robot-navigation-24", "wall-following-robot-navigation-4", "wine", "yeast", "zoo".}

\subsection{Pipeline Search Space}
\label{supp:search_space}
We build pipelines using scikit-learn \cite{scikit-learn} primitives.
The available components are listed in Table~\ref{table:pipeline_components}.
``null'' denotes a pass-through.

\section{Commonly Used Meta-features}
See Table~\ref{table:metafeatures}.

\begin{table*}
	\caption{Dataset Meta-features} 
	\label{table:metafeatures}
	\centering
	\begin{tabular}{lll}
		\hline
		\textbf{Meta-feature name}  &   \textbf{Explanation} \\
		\hline
		number of instances & number of data points in the dataset \\
		log number of instances &  \multicolumn{1}{p{6cm}}{the (natural) logarithm of number of instances}\\
		number of classes &  \\
		number of features &  \\
		log number of features & \multicolumn{1}{p{6cm}}{the (natural) logarithm of number of features} \\
		number of instances with missing values &  \\
		percentage of instances with missing values &  \\
		number of features with missing values & \\
		percentage of features with missing values &  \\
		number of missing values &  \\
		percentage of missing values & \\
		number of numeric features &  \\
		number of categorical features &  \\
		ratio numerical to nominal & the ratio of number of numerical features to the number of categorical features \\
		ratio numerical to nominal &  \\
		dataset ratio &  the ratio of number of features to the number of data points \\
		log dataset ratio &  the natural logarithm of dataset ratio \\
		inverse dataset ratio &  \\
		log inverse dataset ratio &  \\
		class probability (min, max, mean, std) &  the (min, max, mean, std) of ratios of data points in each class\\
		symbols (min, max, mean, std, sum) & the (min, max, mean, std, sum) of the numbers of symbols in all categorical features\\
		kurtosis (min, max, mean, std) &  \\
		skewness (min, max, mean, std) &  \\
		class entropy & the entropy of the distribution of class labels (logarithm base 2) \\
		& \\
		\textbf{landmarking \cite{pfahringer2000meta} meta-features} &  \\
		LDA & \\
		decision tree & decision tree classifier with 10-fold cross validation\\
		decision node learner & \multicolumn{1}{p{12cm}}{ 10-fold cross-validated decision tree classifier with \texttt{criterion=``entropy'', max\_depth=1, min\_samples\_split=2, min\_samples\_leaf=1,  max\_features=None}} \\
		random node learner & \multicolumn{1}{p{12cm}}{10-fold cross-validated decision tree classifier with \texttt{max\_features=1} and the same above for the rest}\\
		1-NN & \\
		PCA fraction of components for 95\% variance & the fraction of components that account for 95\% of variance\\
		PCA kurtosis first PC & kurtosis of the dimensionality-reduced data matrix along the first principal component\\
		PCA skewness first PC & skewness of the dimensionality-reduced data matrix along the first principal component\\
		
		\hline
	\end{tabular}
\end{table*}

\section{Experiment Design for Weighted Least Squares}
\label{supp:WLS}
When factorizing the error matrix by SVD, we approximate performance of different pipelines to different accuracies.
Different accuracies can be characterized by different variances in the linear regression model, thus the weighted least squares (WLS) model that would theoretically give the best linear unbiased estimate to the new dataset embedding may perform better.

\begin{figure}[H]
	\centering
	\includegraphics[width=.7\linewidth]{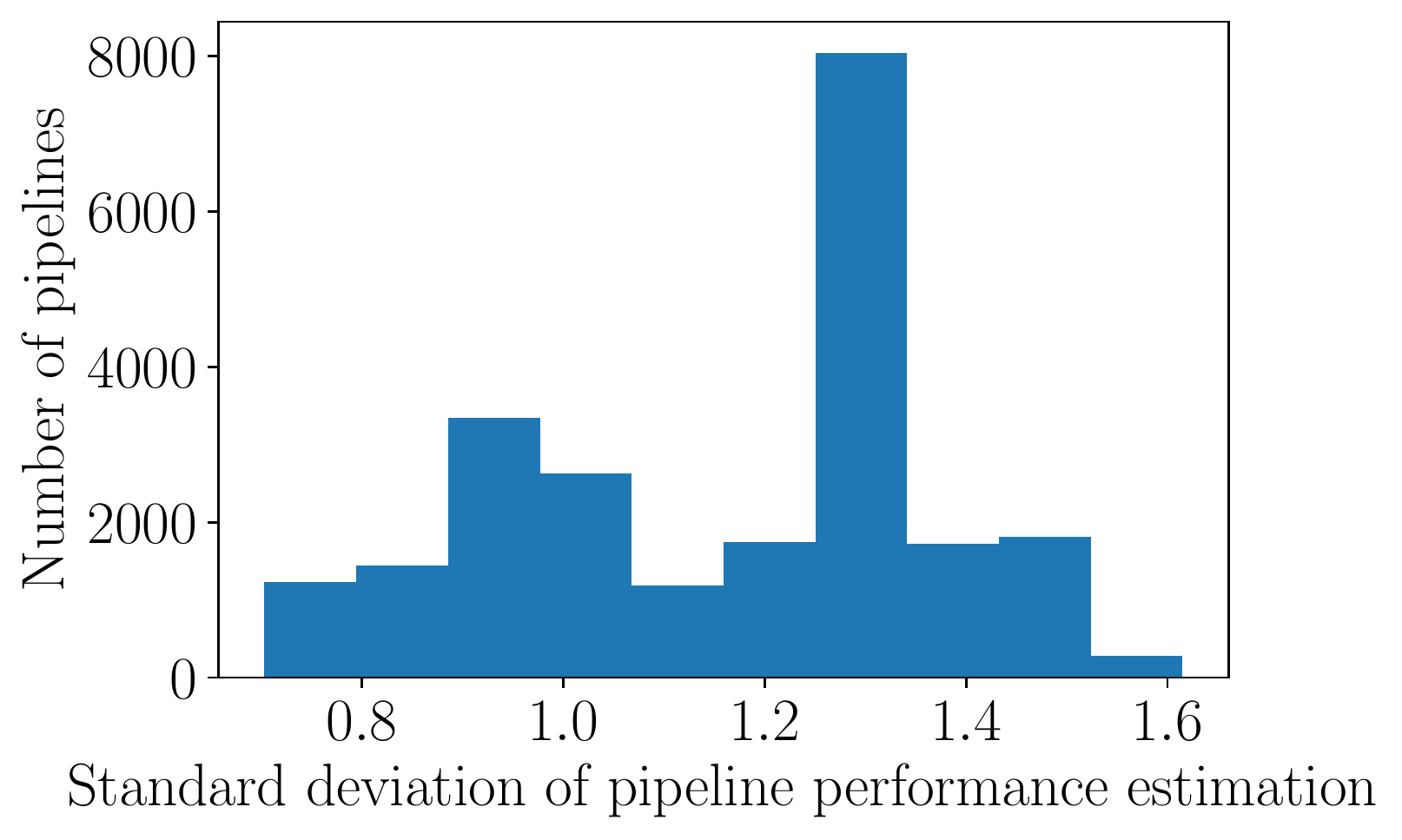}
	\caption{Standard deviation of prediction accuracy of each pipeline, across meta-training datasets.}
	\label{fig:pipeline_std}
\end{figure}

In detail, recall that the constrained $D$-optimal experiment design formulation relies on the assumption that given a low rank matrix multiplication model $X^\top Y = E$, the error term in linear regression $\epsilon \sim \mathcal{N} (0, \sigma^2 I)$, which means each pipeline is predicted to the same accuracy.
In the WLS version of our pipeline performance estimation setting, the pipeline performance vector of the new dataset can be written as $e = Y^\top x + \epsilon$, in which $\epsilon  \sim \mathcal{N} (0, \Sigma)$. $\Sigma = \text{diag}(\sigma_1^2, \sigma_2^2, \dots, \sigma_n^2)$ is a covariance matrix; diagonal in the weighted least squares setting.
For each pipeline $j \in [n]$, we estimate the variance by the sample variance of $E_{:j} - X^\top y_j$, and show a histogram in Figure~\ref{fig:pipeline_std}.
In this case, the time-constrained $D$-experiment design problem to solve becomes
\begin{equation}
\begin{array}{ll}
\text{minimize} & \log \det \Big(\sum_{j=1}^n v_j \frac{y_j y_j^\top }{\sigma_j^2}\Big)^{-1} \\
\text{subject to} & \sum\limits_{j=1}^n v_j \hat{t}_j \leq \tau \\
& v_j \in \{0, 1\}, \forall j \in [n].
\end{array}
\label{eq:ED_time_variance}
\end{equation}

\begin{figure}
	\begin{subfigure}[t]{.45\linewidth}
		\includegraphics[width=\linewidth]{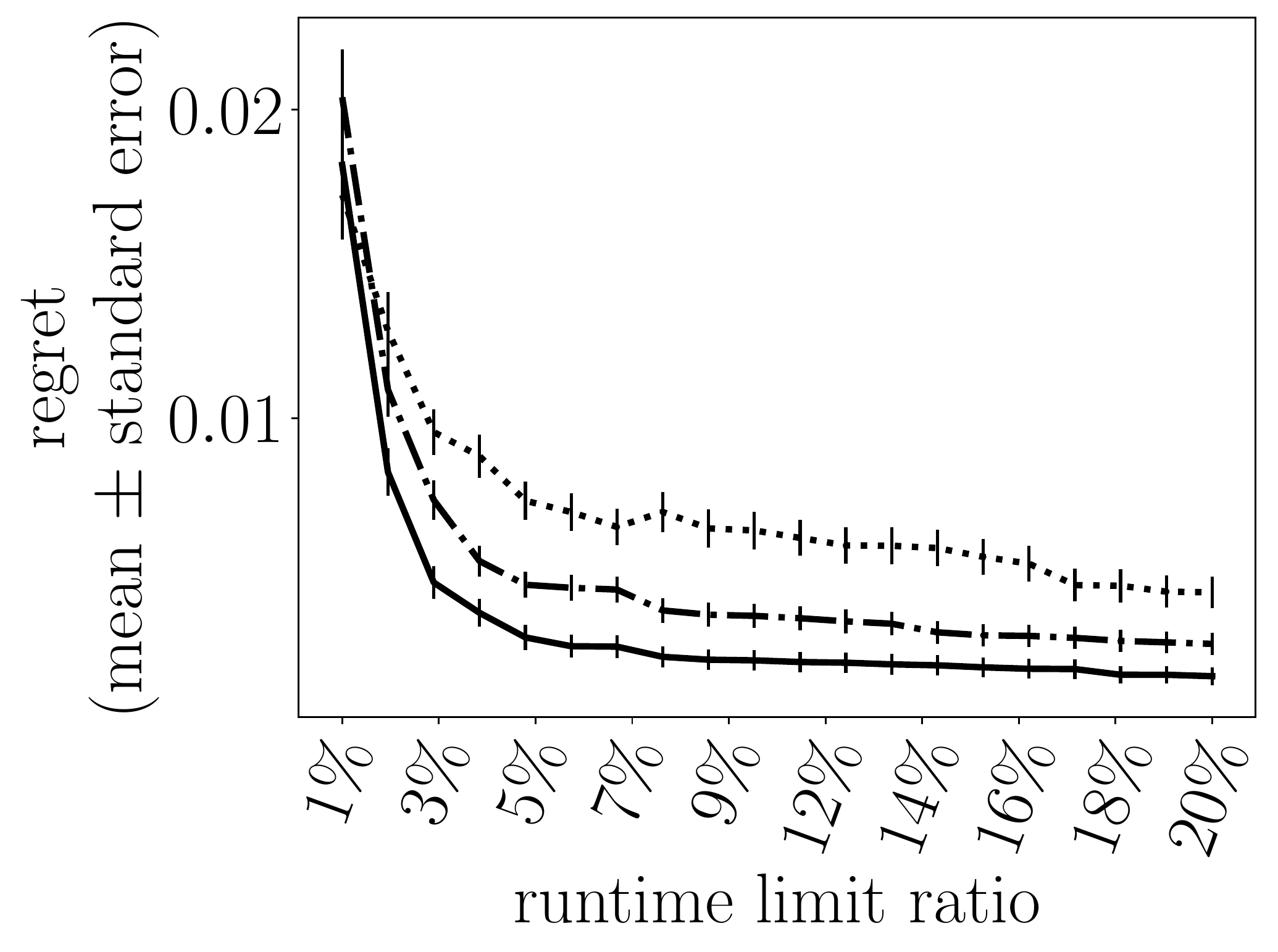}
		\caption{Regret on the subsampled error matrix (215-by-183) for estimator search, including the weighted-greedy method.}
		\label{fig:ED_error_matrix_with_WLS}
	\end{subfigure}%
	\hspace{.09\linewidth}
	\begin{subfigure}[t]{.45\linewidth}
		\includegraphics[width=\linewidth]{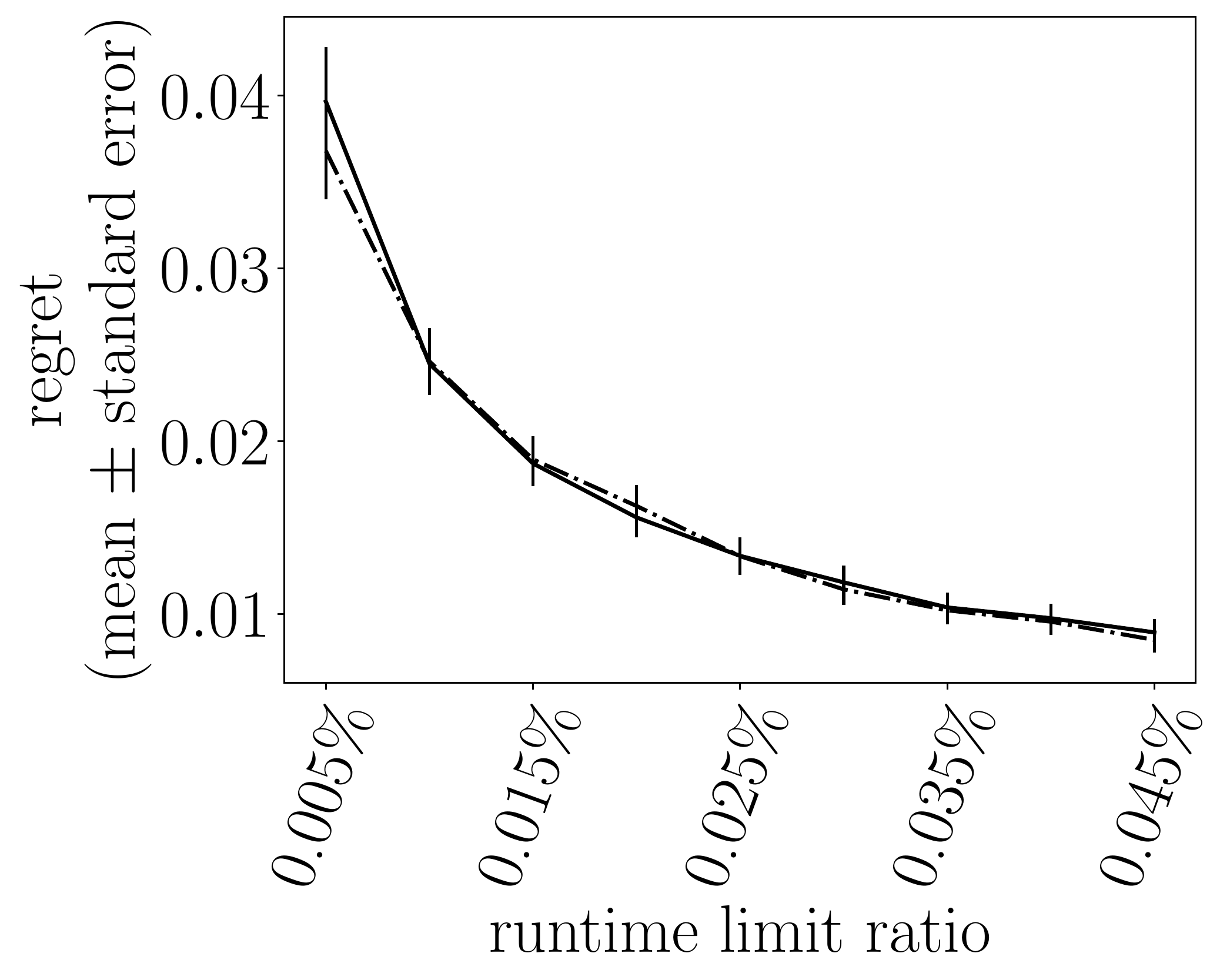}
		\caption{Regret on the full error matrix (215-by-23424) for pipeline search, including the weighted-greedy method.}
		\label{fig:ED_error_tensor_with_WLS}
	\end{subfigure}%
	
	\vspace{.5em}
	\begin{subfigure}[t]{.9\linewidth}
		\includegraphics[width=\linewidth]{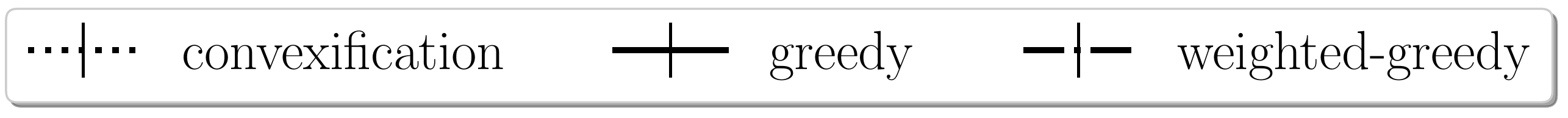}
		\label{fig:legend_ED_with_WLS}
	\end{subfigure}%
	\vspace{-2em}
	
	\caption{Comparison of time-constrained experiment design methods, including the weighted-greedy method.}
	\label{fig:ED_comparison_with_WLS}
\end{figure}

The corresponding greedy approach, which we call \emph{weighted-greedy}, is shown as Algorithm~\ref{alg:ED_time_with_variance_greedy}. 
It differs from the ordinary greedy approach in that each $y_j$ is scaled by $1/\sigma_j$.
Figure~\ref{fig:ED_comparison_with_WLS} shows its performance compared to convexification and greedy.
We can see the weighted-greedy approach performs similarly to the ordinary greedy approach in our experiments.

\begin{algorithm}[H]
	\caption{Greedy algorithm for time-constrained $D$-design in WLS setting, with QR initialization}
	\label{alg:ED_time_with_variance_greedy}
	\begin{algorithmic}[1]
		\Require{design vectors $\{y_j\}_{j=1}^n$, in which $y_j \in \mathbb{R}^k$; pipeline estimation variances $\{\sigma_j^2\}_{j=1}^n$, (predicted) running time of all pipelines $\{\hat{t}_i\}_{i=1}^n$; maximum running time $\tau$}
		\Ensure{The selected set of designs $S\subseteq [n]$}
		\State $y_j \gets y_j / \sigma_j$, $\forall j \in [n]$
		\State $S_0 \gets \texttt{QR\_initialization}(\{y_j\}_{j=1}^n, \{\hat{t}_i\}_{i=1}^n, \tau)$
		\State $S \gets \texttt{Greedy\_without\_repetition}(\{y_j\}_{j=1}^n, \{\hat{t}_i\}_{i=1}^n, \linebreak \tau, S_0)$
	\end{algorithmic}
\end{algorithm}

\section{Zoomed-in Hyperparameter Landscapes}
\label{supp:zoomed_in_landscapes}
\begin{figure}[h]
	\centering
	\begin{subfigure}[t]{.45\linewidth}
		\includegraphics[width=\linewidth]{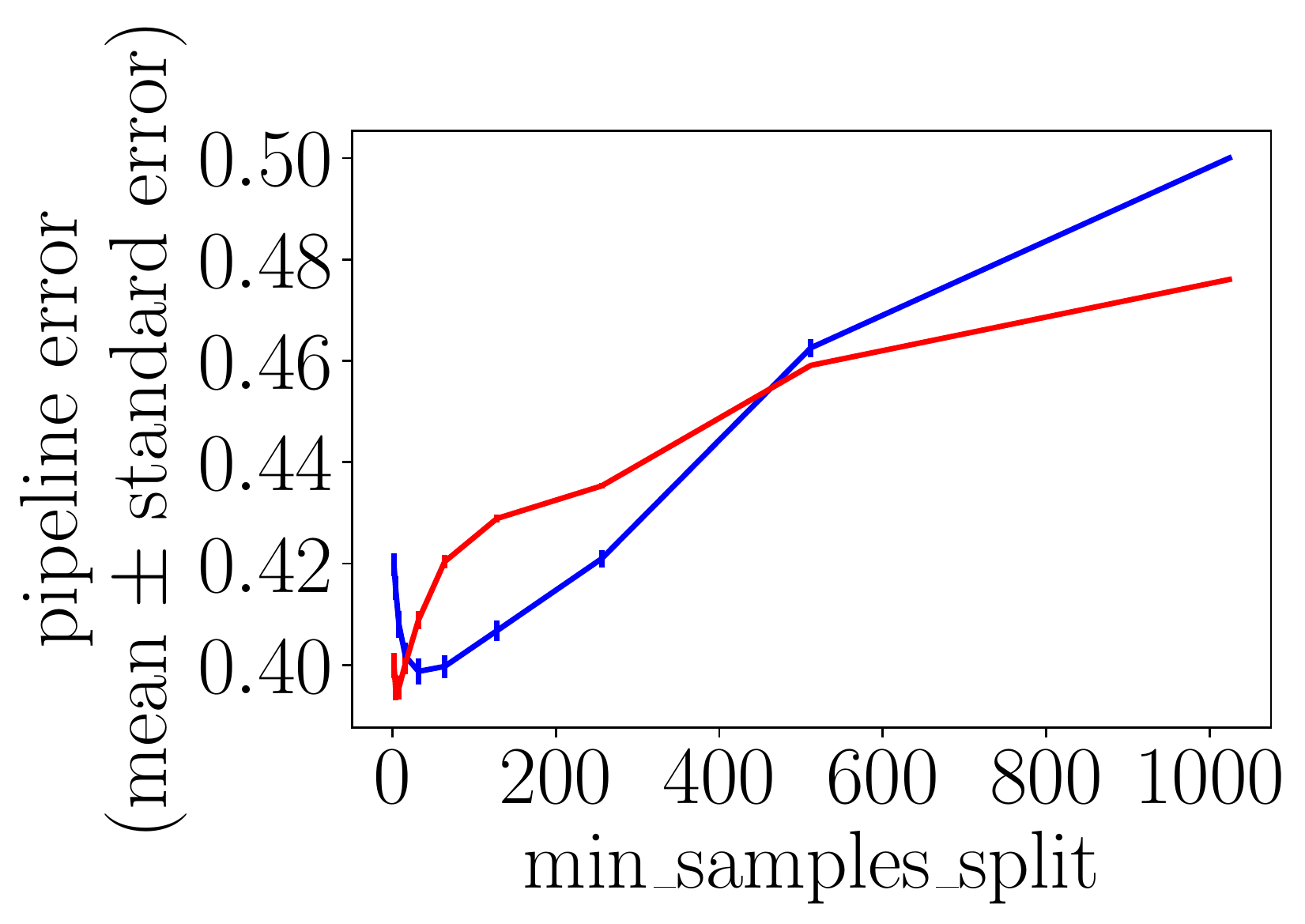}
		\caption{Extra trees on Dataset 23 (1473 points, 10 features)}
		\label{fig:landscape_1_zoomed_in}
	\end{subfigure}%
	\hspace{.09\linewidth}
	\begin{subfigure}[t]{.45\linewidth}
		\includegraphics[width=\linewidth]{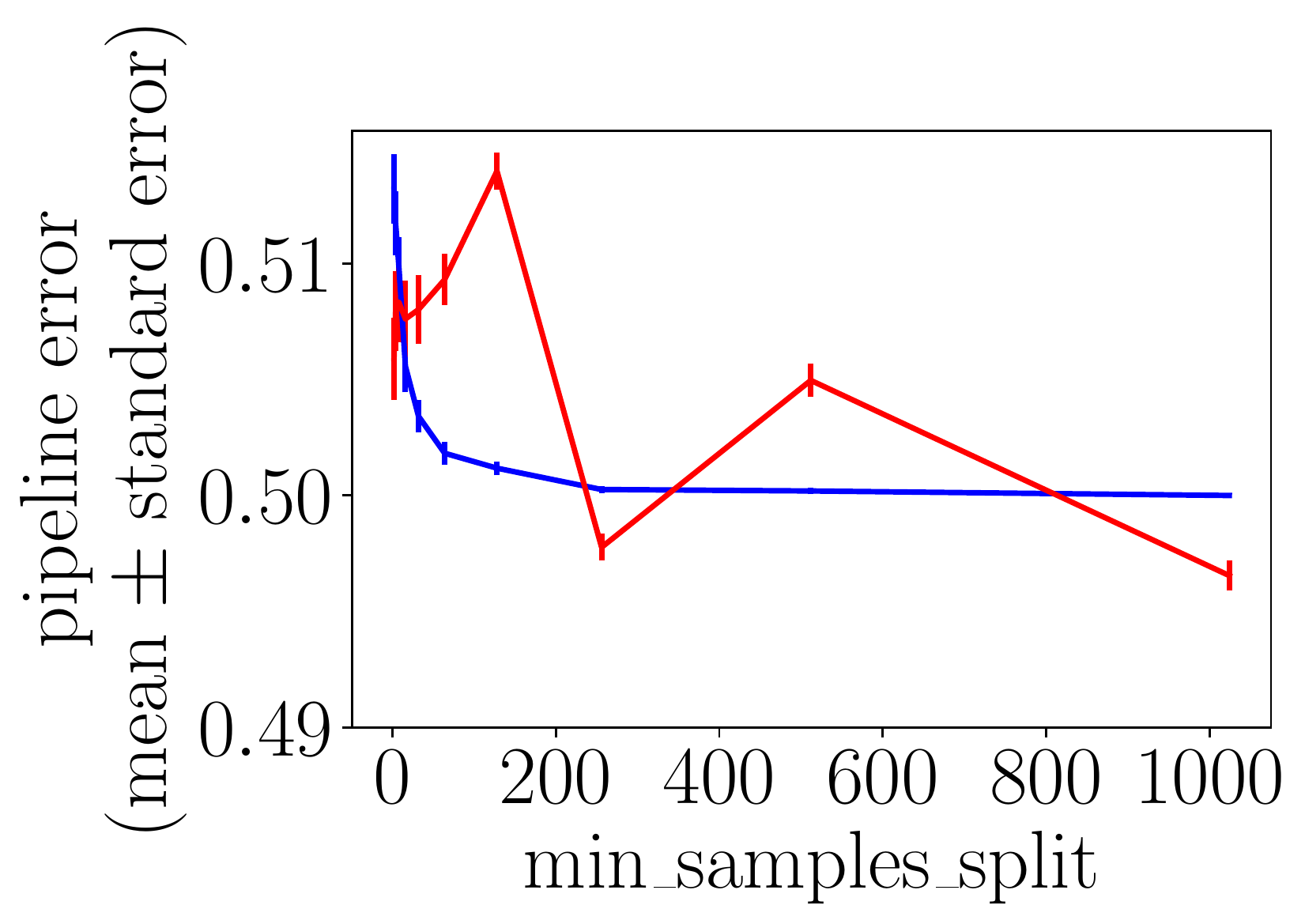}
		\caption{Decision tree on Dataset 1014 (797 points, 5 features)}
		\label{fig:landscape_2_zoomed_in}
	\end{subfigure}%
	
	\begin{subfigure}[t]{.45\linewidth}
		\includegraphics[width=\linewidth]{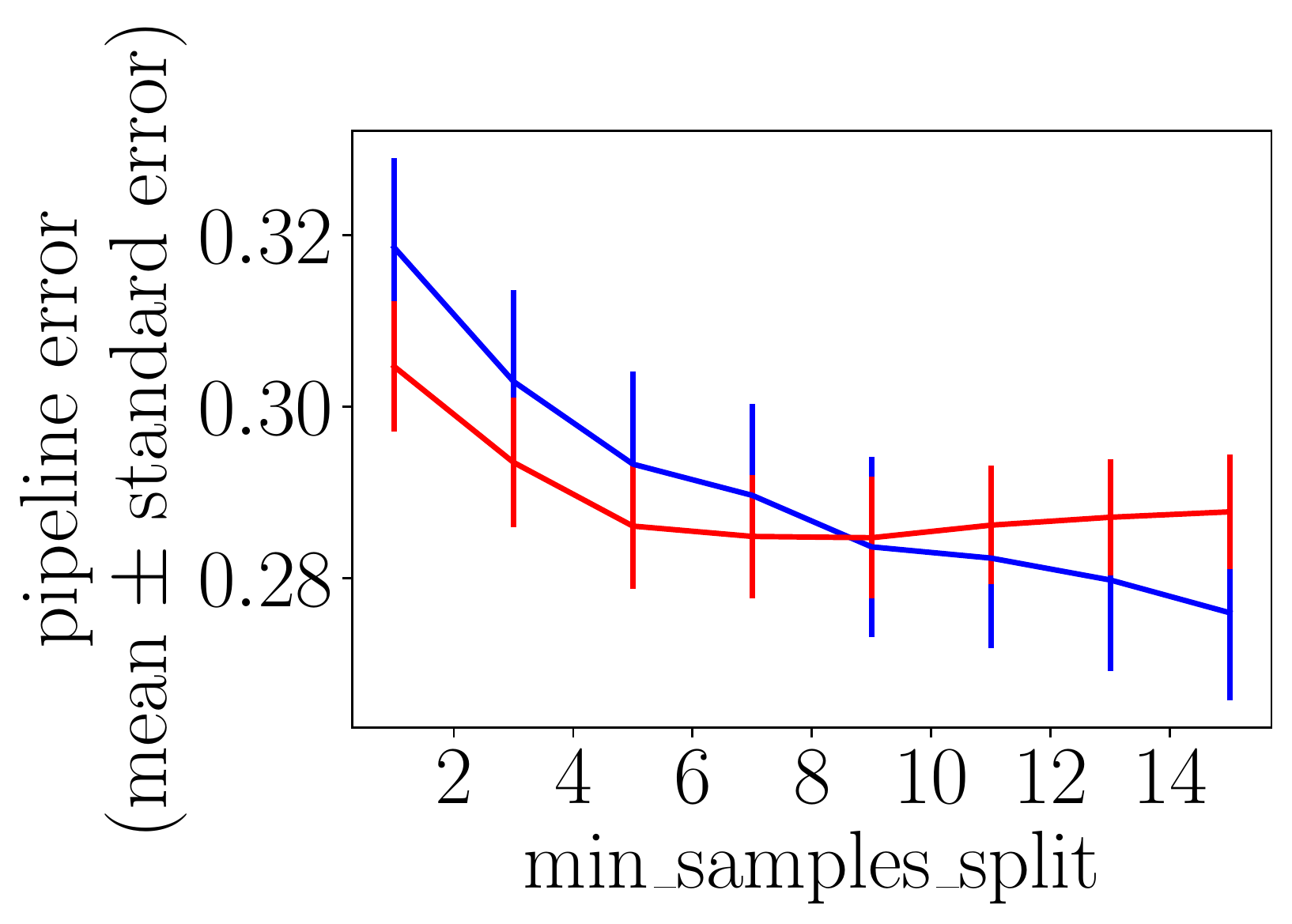}
		\caption{kNN on Dataset 799 (1000 points, 6 features)}
		\label{fig:landscape_3_zoomed_in}
	\end{subfigure}%
	\hspace{.09\linewidth}
	\begin{subfigure}[t]{.45\linewidth}
		\includegraphics[width=\linewidth]{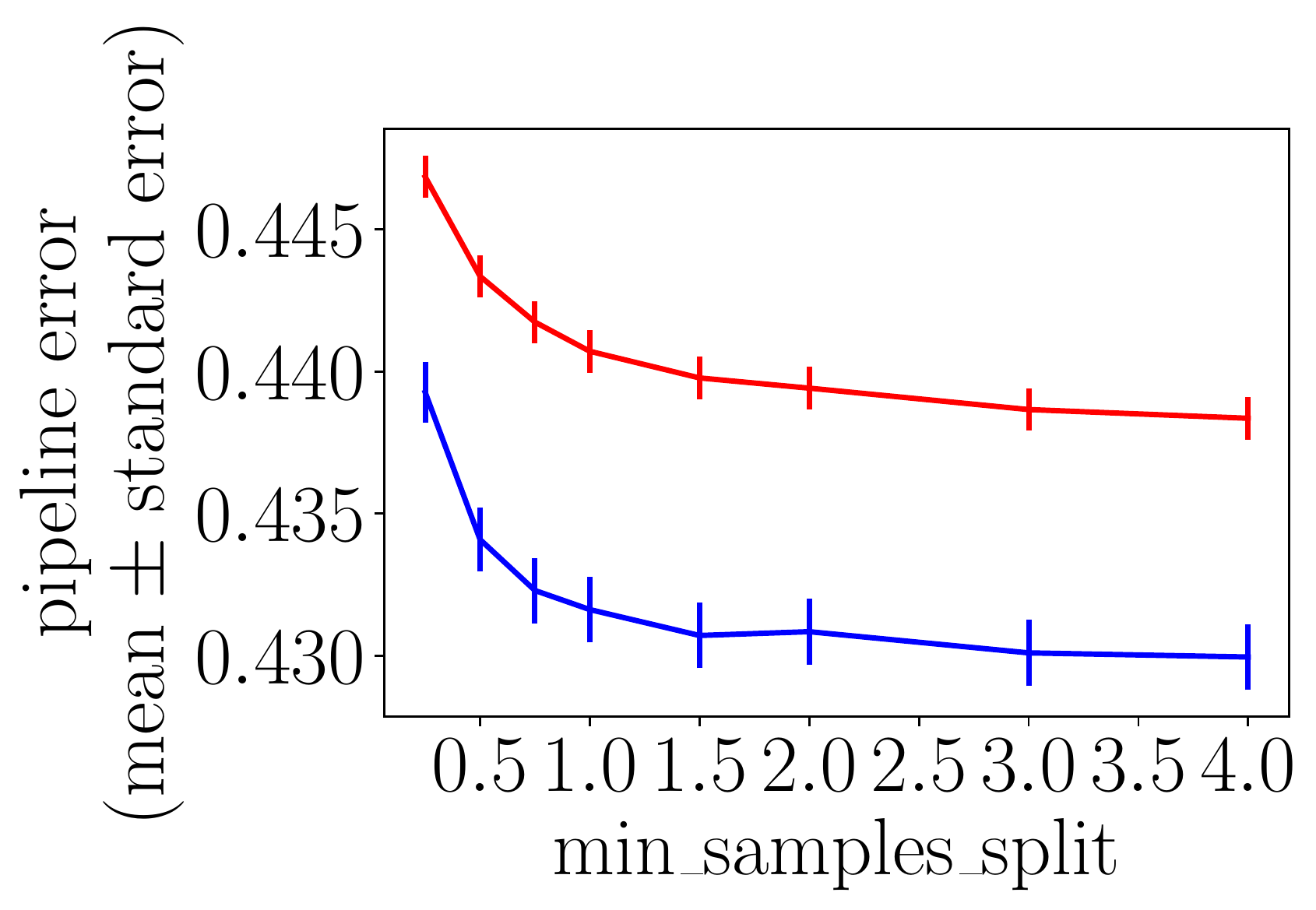}
		\caption{Logistic regression on Dataset 40971 (1000 data points, 24 features)}
		\label{fig:landscape_4_zoomed_in}
	\end{subfigure}%
	
	\begin{subfigure}[t]{.52\linewidth}
		\includegraphics[width=\linewidth]{\figurepath legend_hyperparameter_landscape.pdf}
	\end{subfigure}
	
	\caption{Zoomed-in hyperparameter landscapes in Figure~\ref{fig:landscapes}.
		The y-axes here do not start from 0.}
	\label{fig:zoomed_in_landscapes}
\end{figure}

See Figure \ref{fig:zoomed_in_landscapes}.

\section{Runtime Prediction Accuracy}
\label{supp:runtime_prediction_accuracy}
\begin{figure*}{h}
	\centering
	\begin{subfigure}[b]{0.33\linewidth}
		\includegraphics[width=\linewidth]{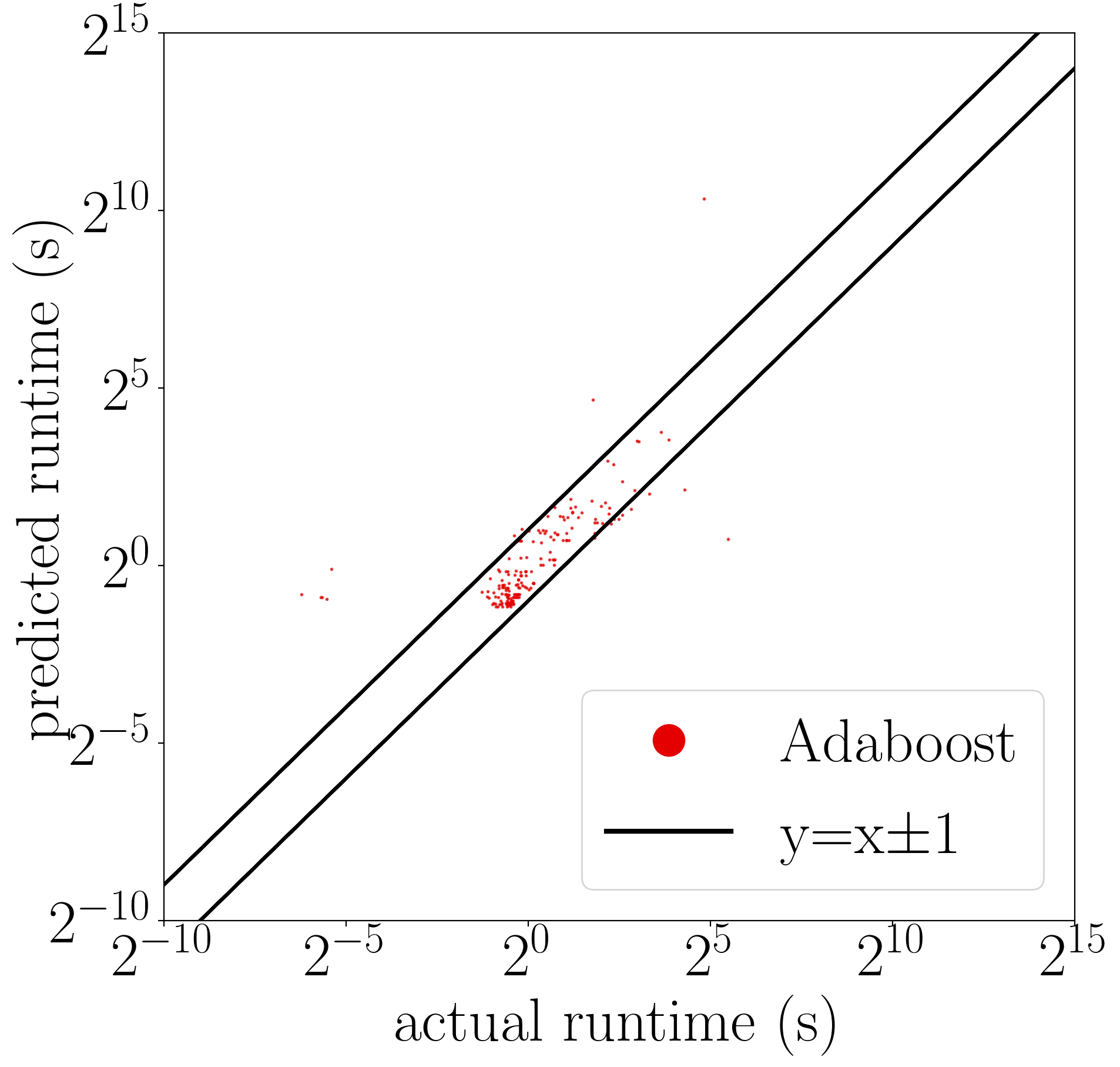}
	\end{subfigure}%
	\begin{subfigure}[b]{0.33\linewidth}
		\includegraphics[width=\linewidth]{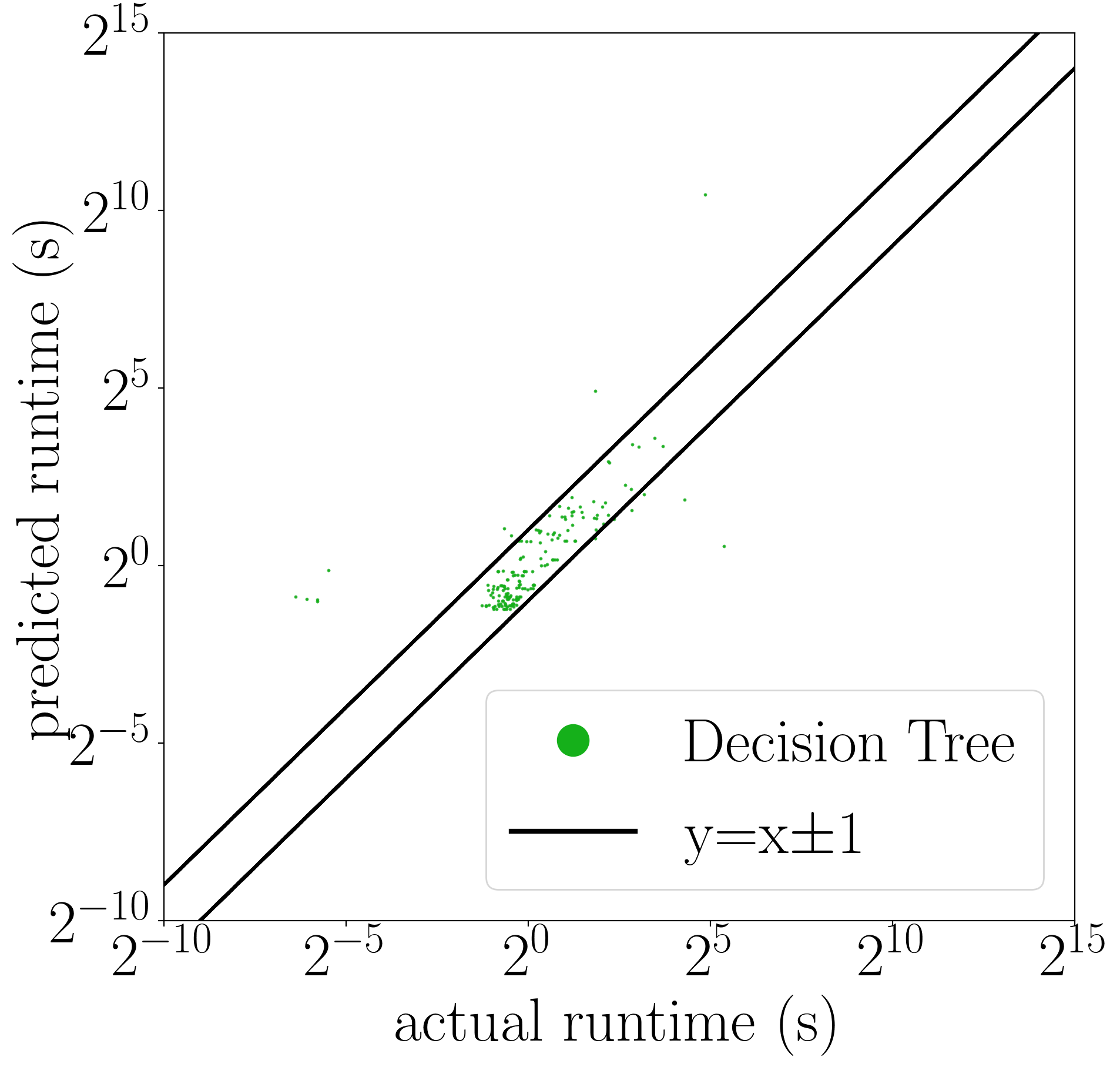}
	\end{subfigure}%
	
	\begin{subfigure}[b]{0.33\linewidth}
		\includegraphics[width=\linewidth]{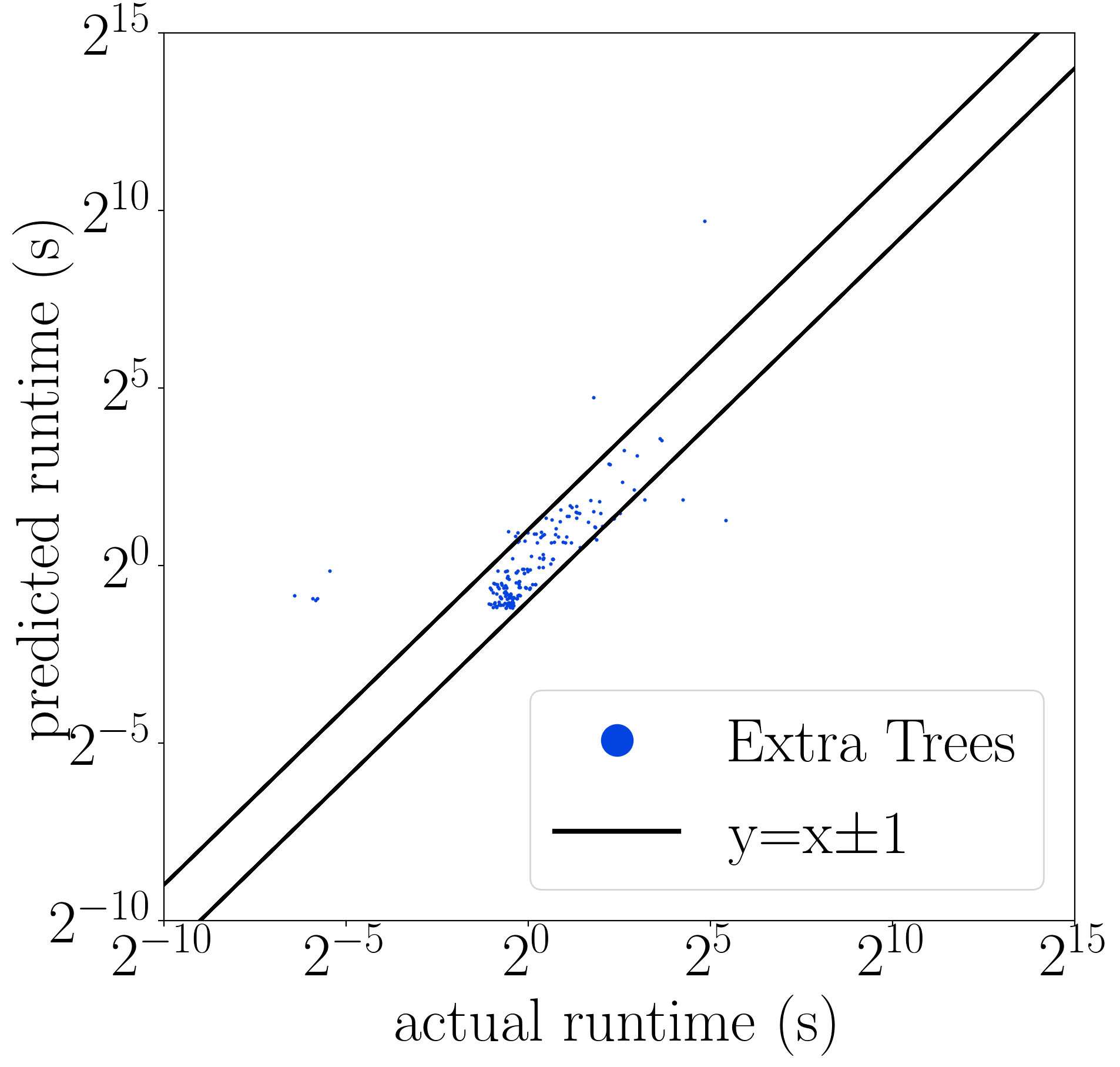}
	\end{subfigure}%	
	\begin{subfigure}[b]{0.33\linewidth}
		\includegraphics[width=\linewidth]{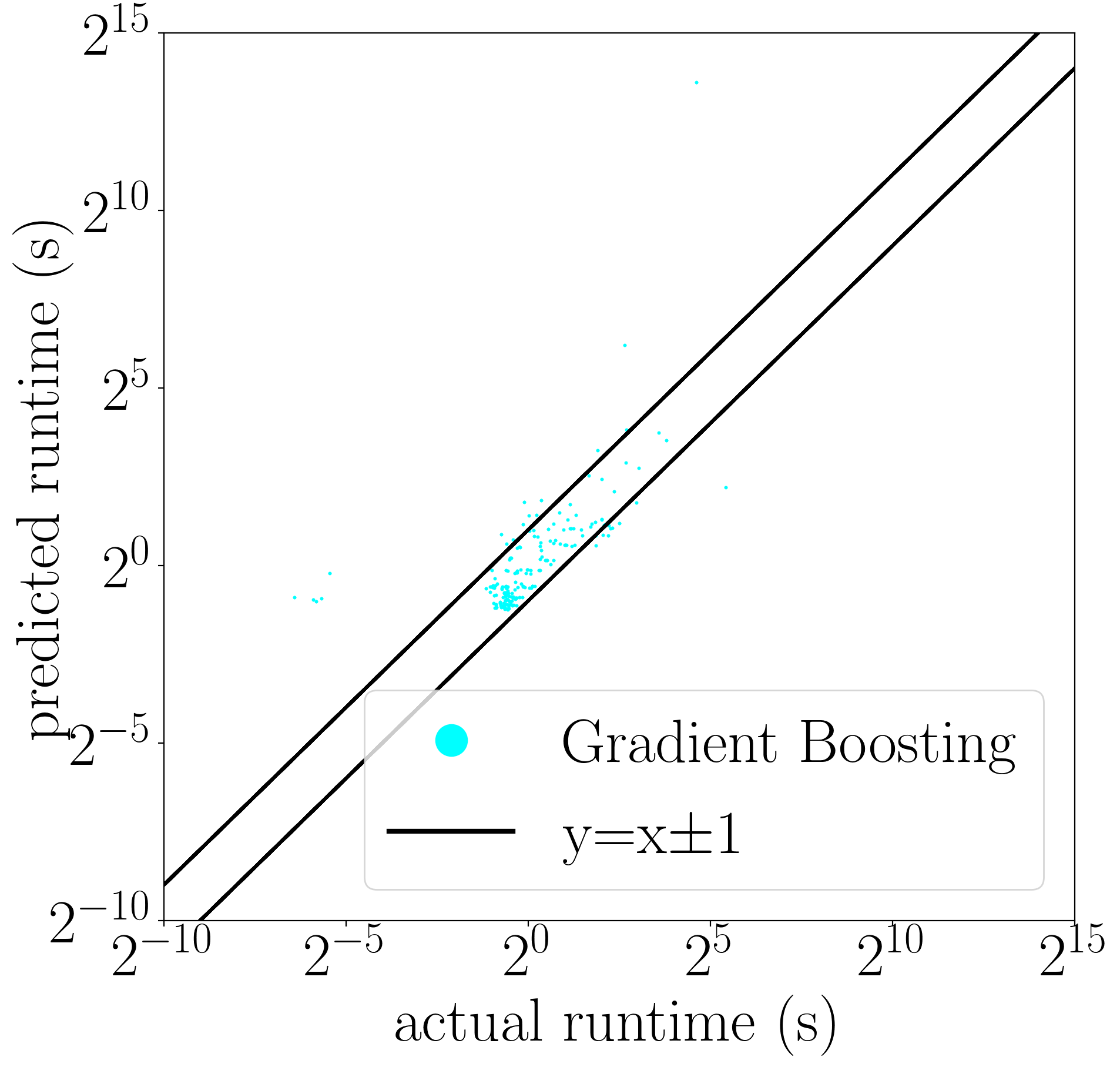}
	\end{subfigure}%
	\begin{subfigure}[b]{0.33\linewidth}
		\includegraphics[width=\linewidth]{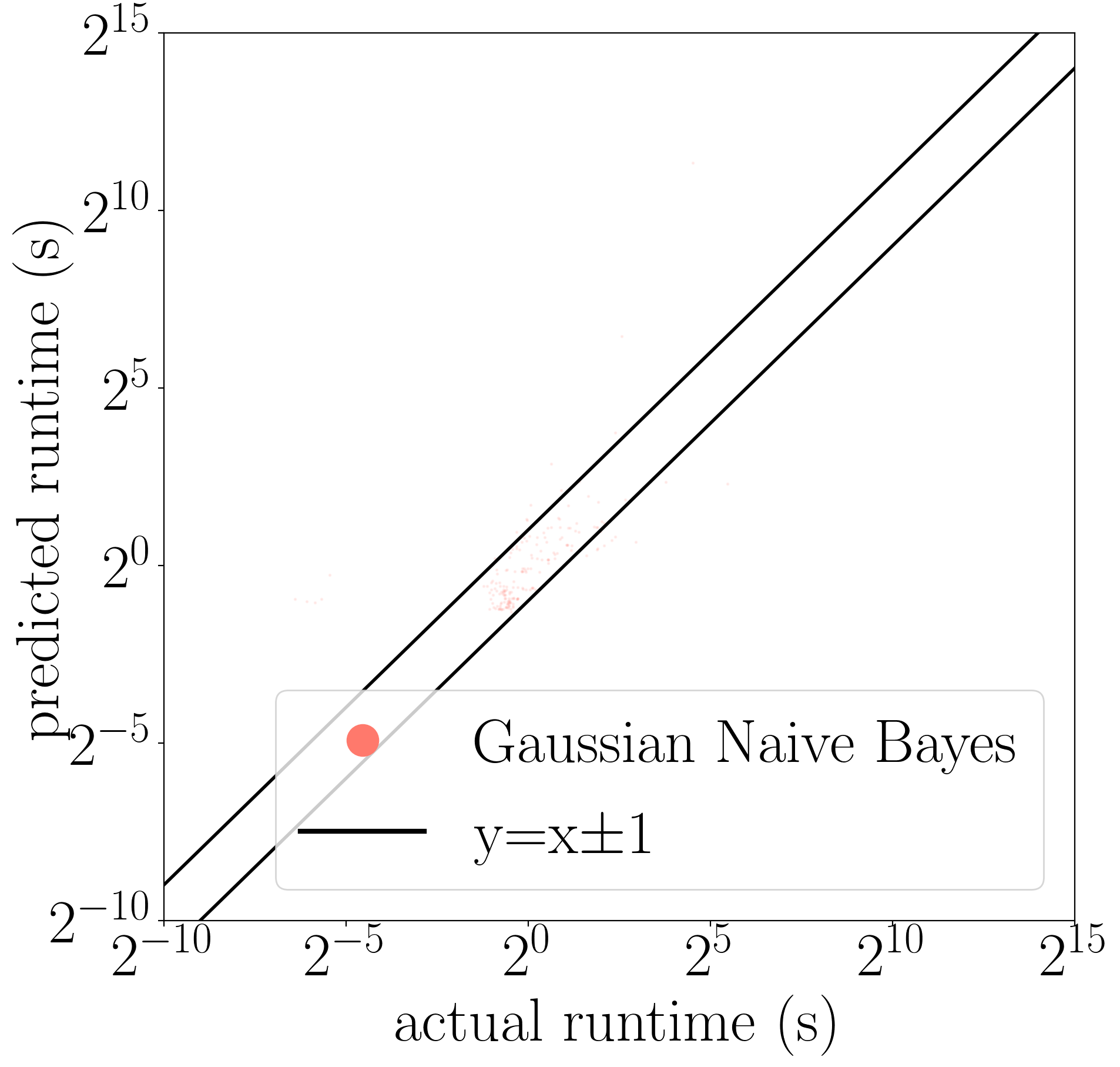}
	\end{subfigure}%
	
	\begin{subfigure}[b]{0.33\linewidth}
		\includegraphics[width=\linewidth]{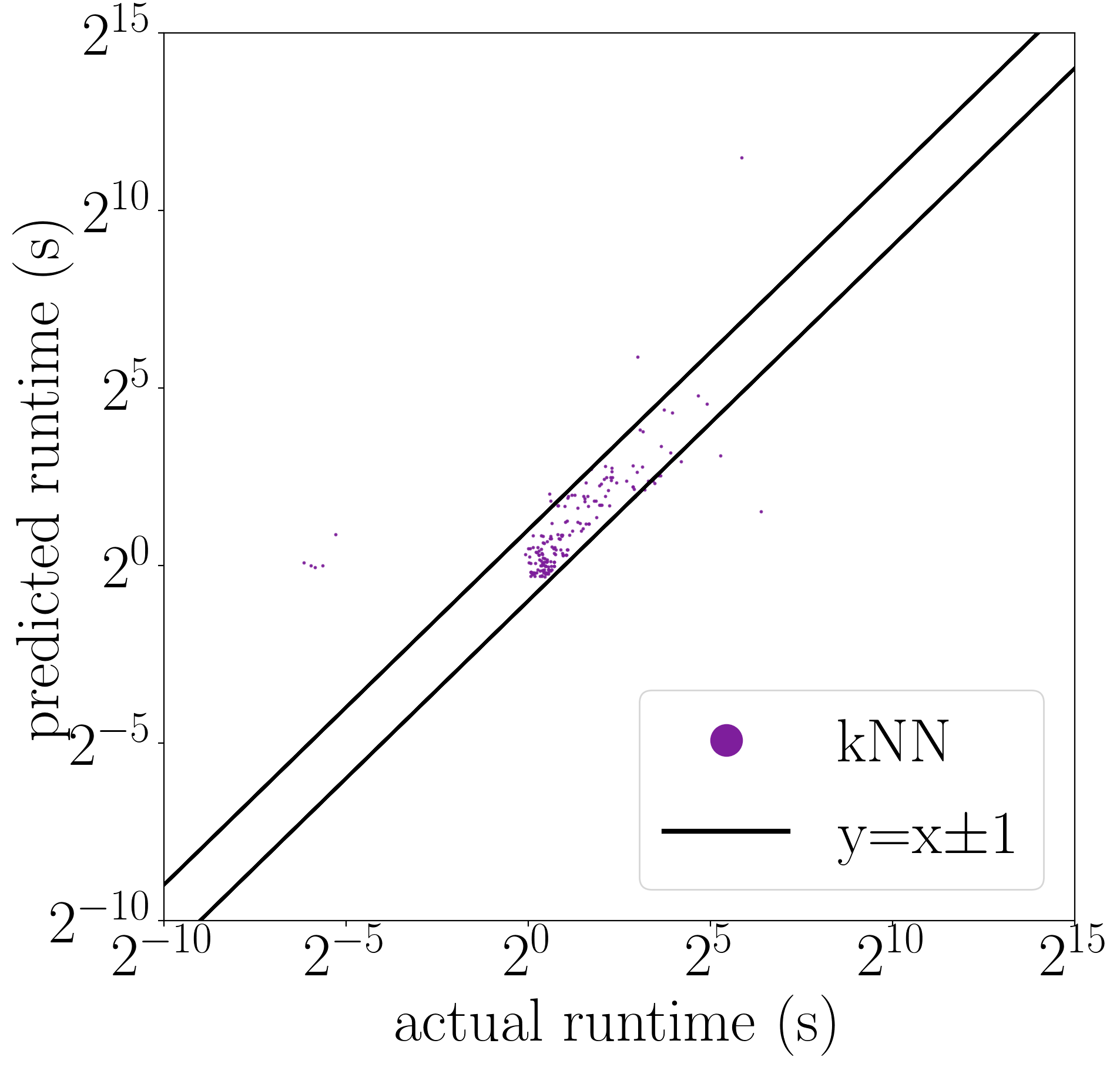}
	\end{subfigure}%	
	\begin{subfigure}[b]{0.33\linewidth}
		\includegraphics[width=\linewidth]{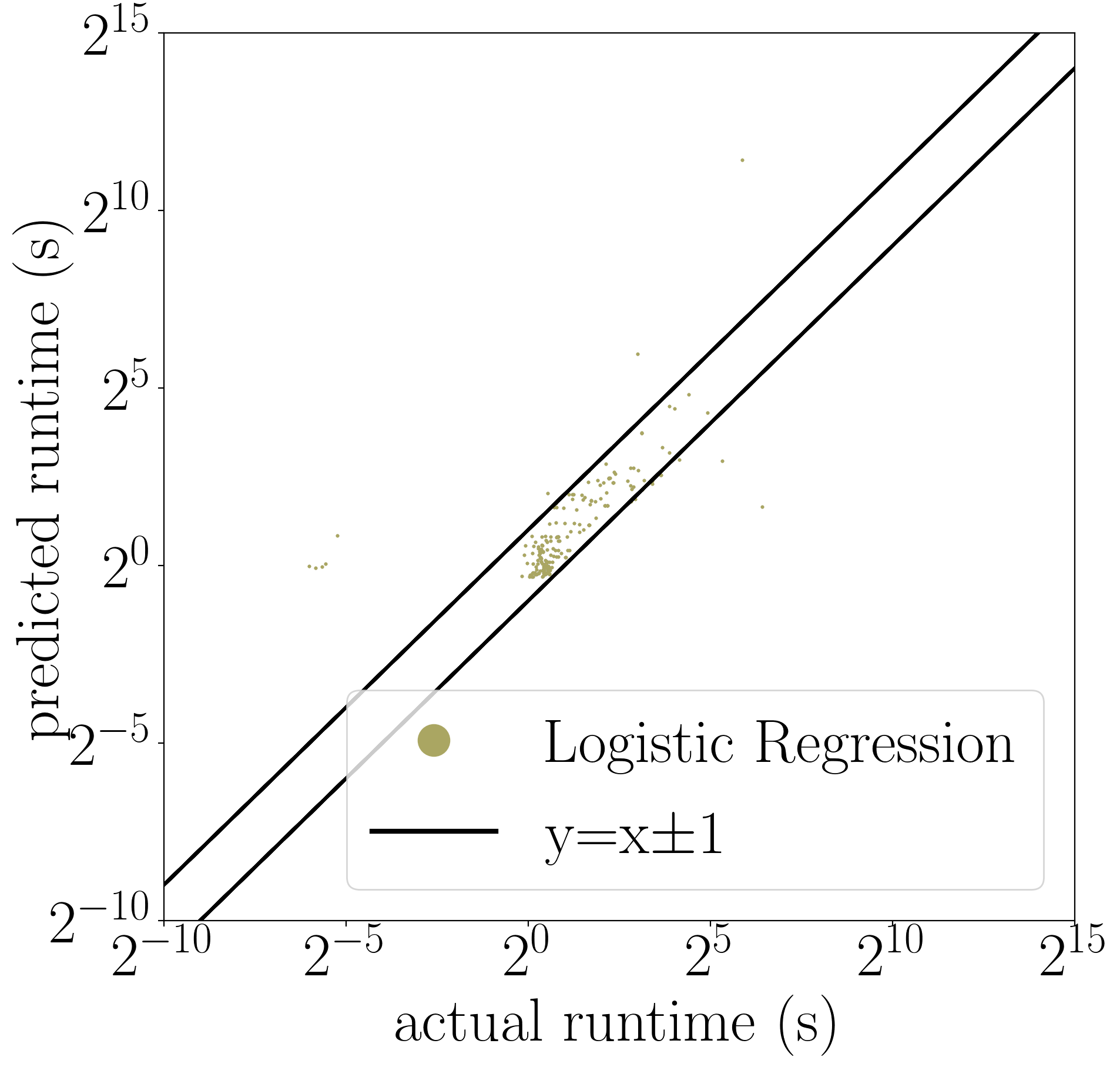}
	\end{subfigure}%
	\begin{subfigure}[b]{0.33\linewidth}
		\includegraphics[width=\linewidth]{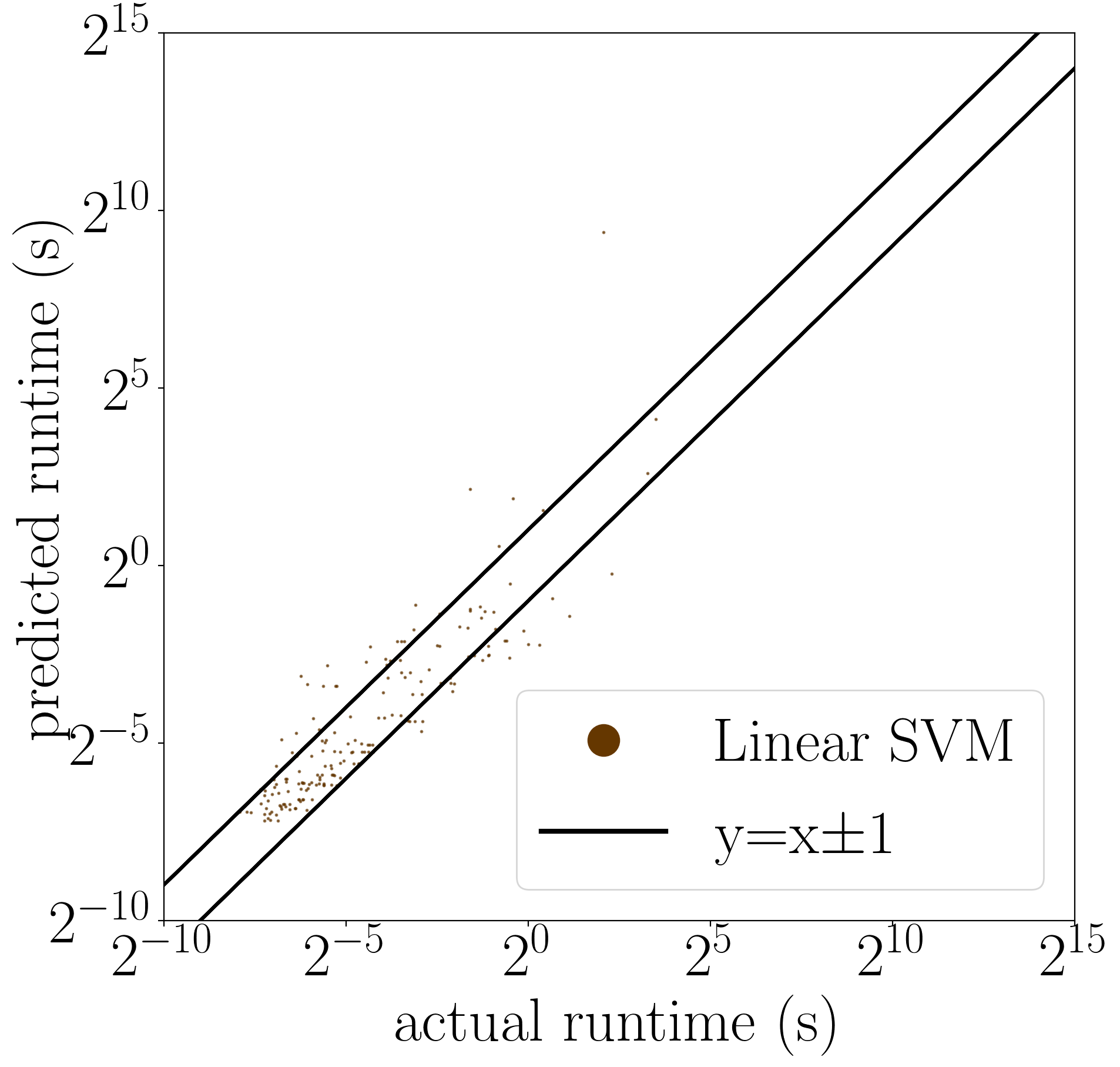}
	\end{subfigure}%
	
	\begin{subfigure}[b]{0.33\linewidth}
		\includegraphics[width=\linewidth]{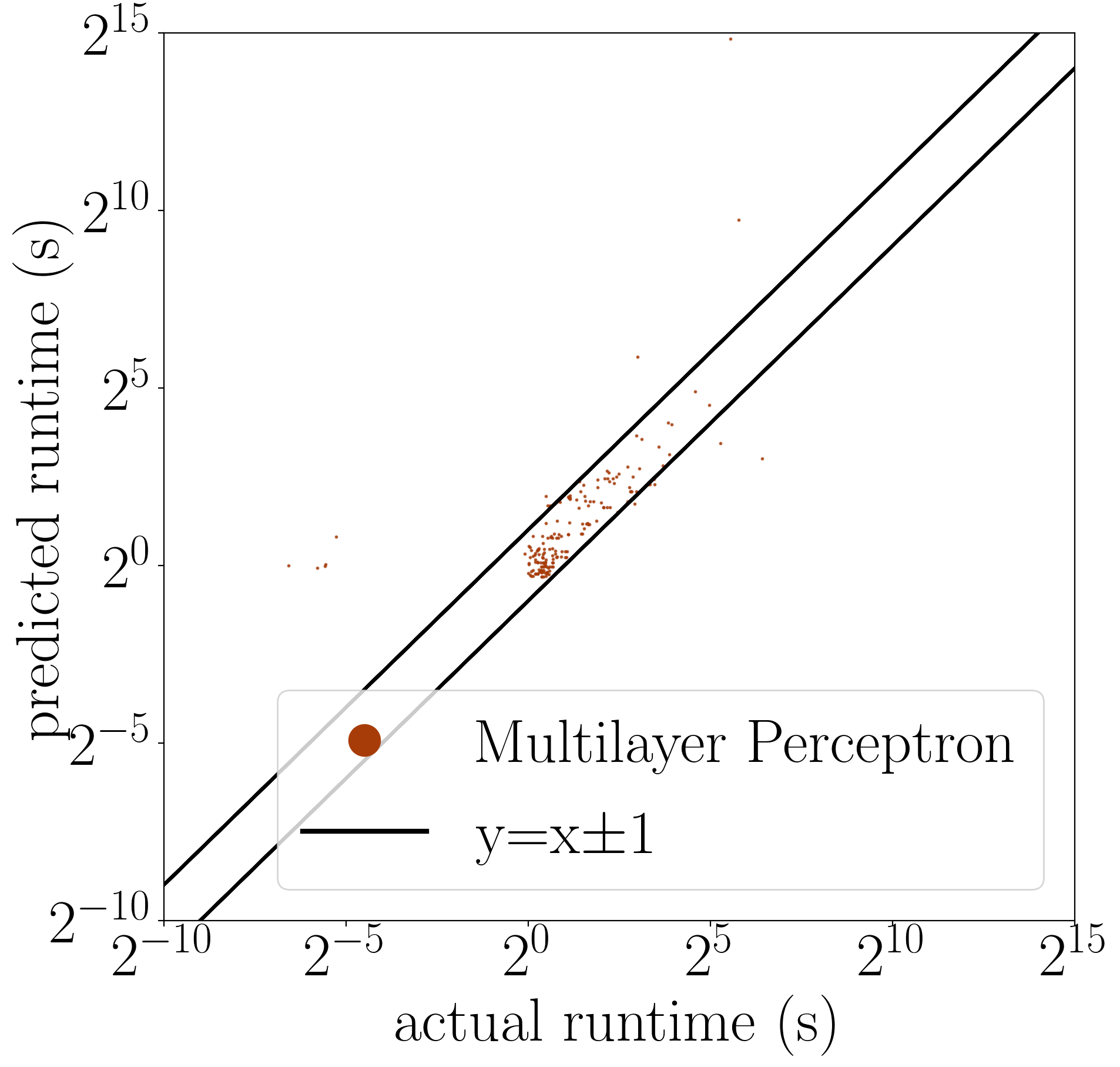}
	\end{subfigure}%
	\begin{subfigure}[b]{0.33\linewidth}
		\includegraphics[width=\linewidth]{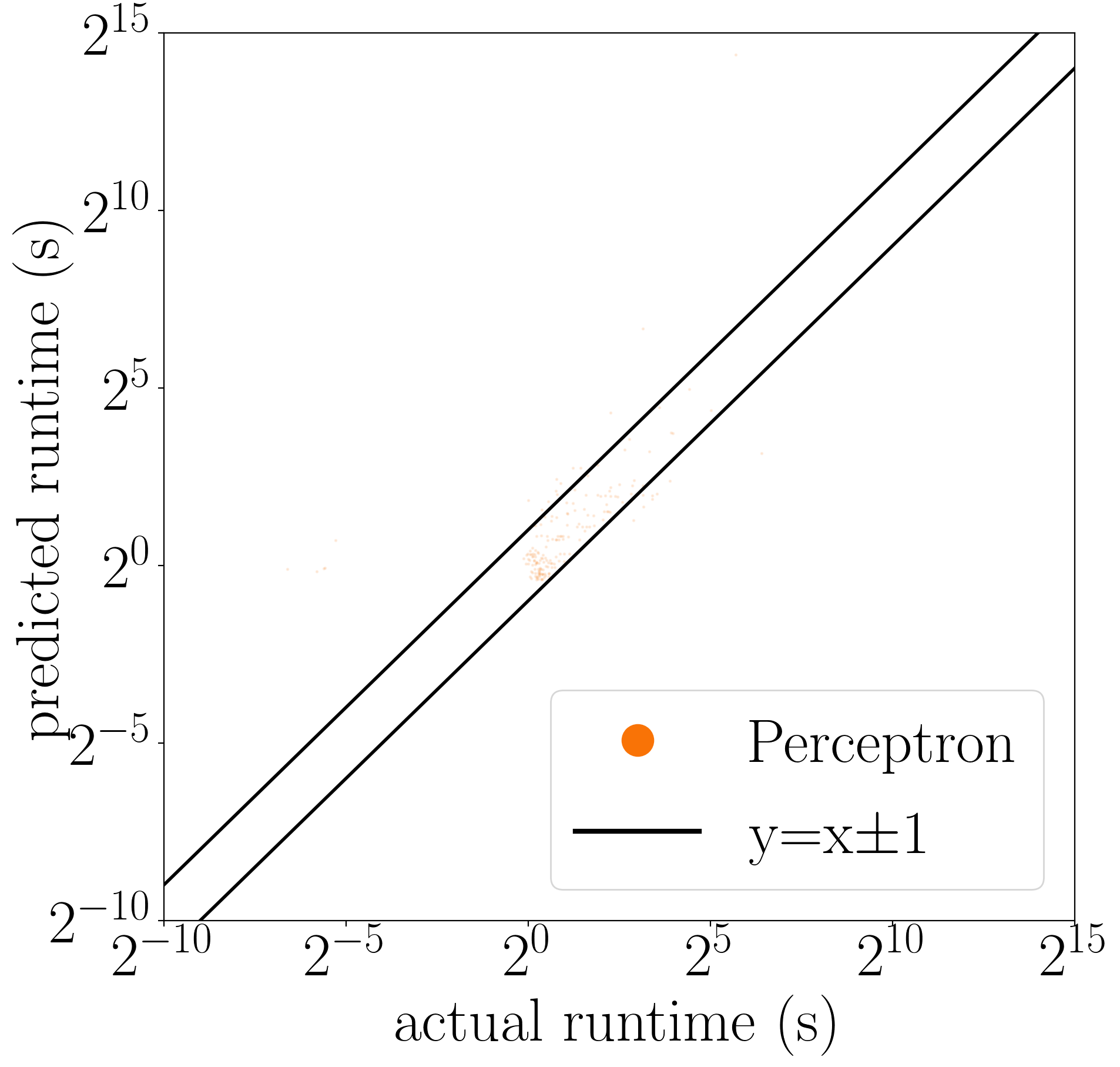}
	\end{subfigure}%
	\begin{subfigure}[b]{0.33\linewidth}
		\includegraphics[width=\linewidth]{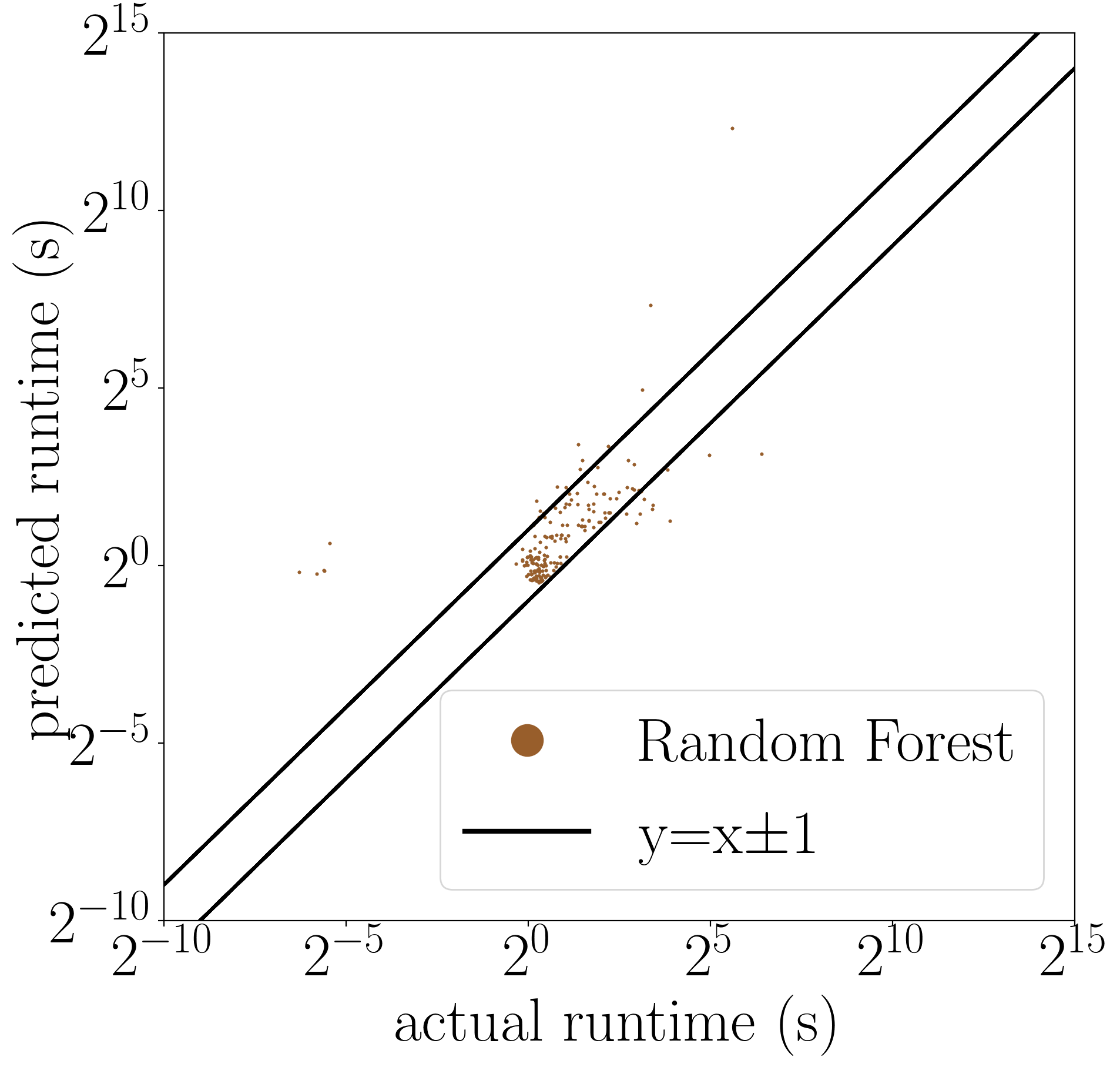}
	\end{subfigure}%
	
	\caption{Runtime prediction performance on different machine learning algorithms, on meta-training OpenML datasets.}
	\label{fig:runtime_prediction_by_type}
\end{figure*}

See Figure \ref{fig:runtime_prediction_by_type}.
Similar plots appeared in our previous KDD version at~\cite{yang2019oboe}. 

\end{document}